\documentclass[10pt,twocolumn,letterpaper]{article}

\usepackage[pagenumbers]{wacv} 

\usepackage{graphicx}
\usepackage{amsmath}
\usepackage{amssymb}
\usepackage{booktabs}
\usepackage{multirow}
\usepackage{array}
\usepackage{xcolor}
\usepackage[accsupp]{axessibility}
\usepackage[pagebackref,breaklinks,colorlinks,citecolor=mydarkblue]{hyperref}

\usepackage[pagebackref,breaklinks,colorlinks]{hyperref}

\usepackage{float}

\usepackage{fontawesome5}

\usepackage{subcaption}

\usepackage[export]{adjustbox}

\usepackage[thinlines]{easytable}
\usepackage{collcell}
\usepackage{tikz}

\newcommand{\increasenoparent}[1]{\textcolor{ForestGreen}{+#1}}

\newcommand{\decreasenoparent}[1]{\textcolor{red}{-#1}}

\newcommand{\na}[0]{\textcolor{gray}{n/a}}

\newcommand{\subsec}[1]{\noindent\textbf{#1}~~}

\makeatletter
\g@addto@macro{\endtabular}{\rowfont{}}
\makeatother
\newcommand{\rowfonttype}{}
\newcommand{\rowfont}[1]{
   \gdef\rowfonttype{#1}#1%
}
\newcolumntype{L}{>{\rowfonttype}l}
\usepackage{listings}

\definecolor{codegreen}{rgb}{0,0.6,0}
\definecolor{codegray}{rgb}{0.5,0.5,0.5}

\definecolor{backcolour}{RGB}{245,248,250}
\definecolor{emph}{RGB}{166,88,53}
\definecolor{nightblue}{RGB}{9,49,105}
\definecolor{keywords}{RGB}{207,33,46}
\definecolor{lightpurple}{RGB}{130,81,223}

\definecolor{MyLightGray}{rgb}{0.95, 0.95, 0.95}
\definecolor{CLIPBlue}{rgb}{0.192, 0.454, 0.643}
\definecolor{ForestGreen}{RGB}{34,139,34}

\lstdefinestyle{mystyle}{
    backgroundcolor=\color{backcolour},   
    commentstyle=\color{codegreen},
    keywordstyle=\color{keywords},
    stringstyle=\color{nightblue},
    basicstyle=\ttfamily\footnotesize,
    breakatwhitespace=false,         
    breaklines=true,                 
    captionpos=b,                    
    keepspaces=true,                 
    showspaces=false,                
    showstringspaces=false,
    showtabs=false,                  
    tabsize=2,
    frame=shadowbox,
    emph={AutoTokenizer,AutoModelForSequenceClassification,Explainer},
    emphstyle={\color{emph}},
    emph={[2]from_pretrained,compute_table},
    emphstyle={[2]\color{lightpurple}}
}

\newcommand{\kubriclogo}{\raisebox{-0.1cm}{\includegraphics[scale=0.025]{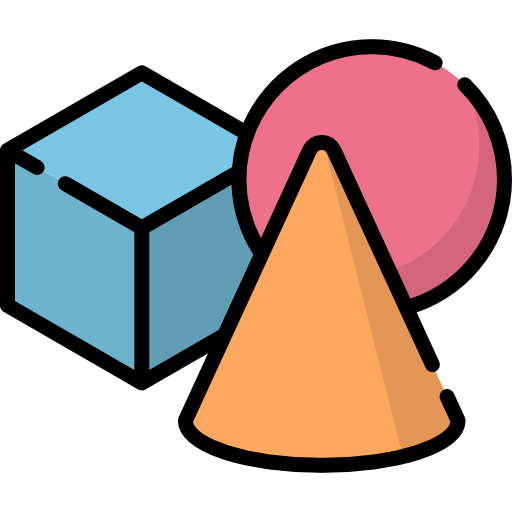}}}
\newcommand{\stdlogo}{\raisebox{-0.09cm}{\includegraphics[scale=0.03]{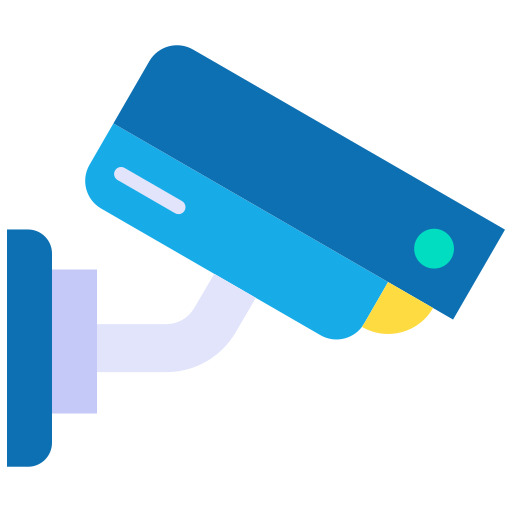}}}
\newcommand{\synthlogo}{\raisebox{-0.070cm}{\includegraphics[scale=0.028]{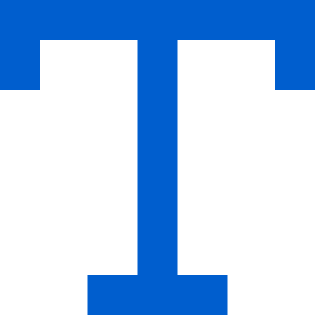}}}
\newcommand{\openlogo}{\raisebox{-0.075cm}{\includegraphics[scale=0.025]{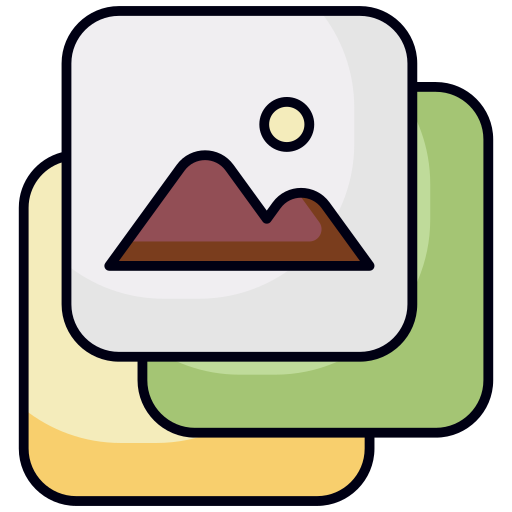}}}
\newcommand{\cocologo}{\raisebox{-0.075cm}{\includegraphics[scale=0.08]{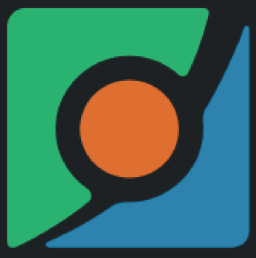}}}

\definecolor{mydarkblue}{rgb}{0,0.53,0.96}

\usepackage[capitalize]{cleveref}
\crefname{section}{Sec.}{Secs.}
\Crefname{section}{Section}{Sections}
\Crefname{table}{Table}{Tables}
\crefname{table}{Tab.}{Tabs.}
\crefname{appendix}{App.}{Apps.}
\definecolor{specialcolor}{rgb}{0.62, 0.32, 0.17}

\usepackage[capitalize]{cleveref}
\crefname{section}{Sec.}{Secs.}
\Crefname{section}{Section}{Sections}
\Crefname{table}{Table}{Tables}
\crefname{table}{Tab.}{Tabs.}

\def\wacvPaperID{989} 
\def\confName{WACV}
\def\confYear{2025}

\title{Improving Zero-Shot Object-Level Change Detection \\by Incorporating Visual Correspondence}

\author{Hung H. Nguyen\\
\small Auburn University\\
{\tt\small hhn0008@auburn.edu}
\and
Pooyan Rahmanzadehgervi\\
\small Auburn University\\
{\tt\small pooyan.rmz@gmail.com}
\and
Long Mai\\
\small Adobe Research\\
{\tt\small mai.t.long88@gmail.com}
\and
Anh Totti Nguyen\\
\small Auburn University\\
{\tt\small anh.ng8@gmail.com}
}

\begin{document}

\twocolumn[{
\renewcommand\twocolumn[1][]{#1}
\maketitle
\begin{center}
\centering
\vspace{-25pt}
\resizebox{\textwidth}{!}{
\fontsize{6pt}{6pt}\selectfont
\begin{tabular}{c@{}c@{}c@{}c@{}c@{}c}
&(a) COCO-Inpainted \cocologo &(b) STD \stdlogo &(c) Kubric-Change \kubriclogo &(d) Synthtext \synthlogo &(e) OpenImages-Inpainted \openlogo \\ 
\rotatebox{90}{\hspace{0.1cm}CYWS \cite{sachdeva2023change}} 
&\includegraphics[width=0.20\linewidth]{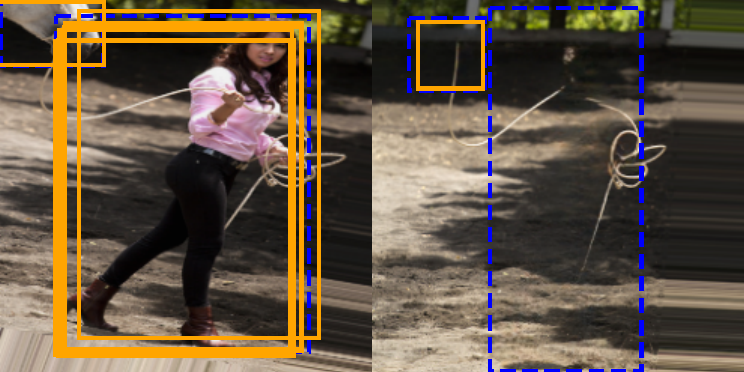} &\includegraphics[width=0.20\linewidth]{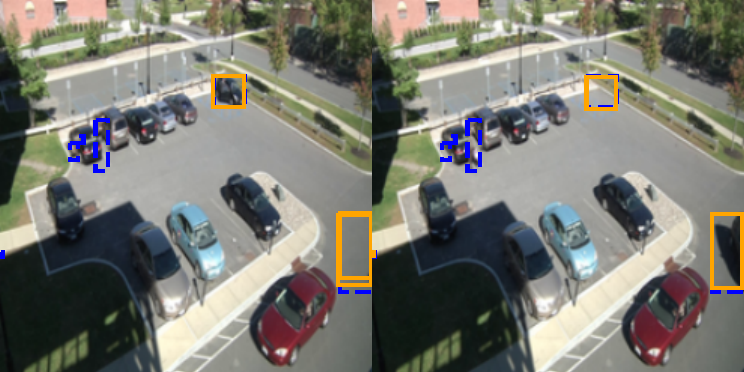}
&\includegraphics[width=0.20\linewidth]{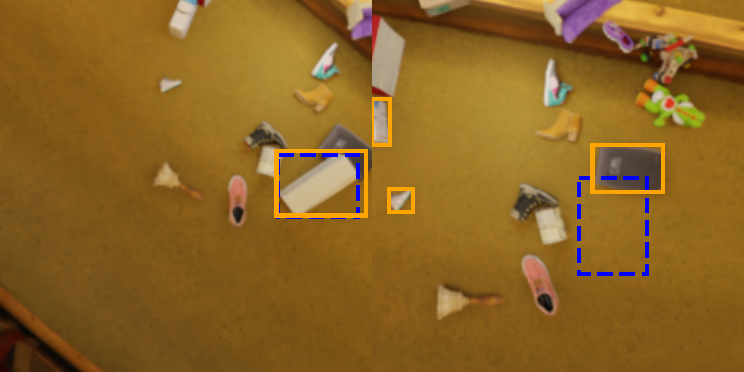}
&\includegraphics[width=0.20\linewidth]{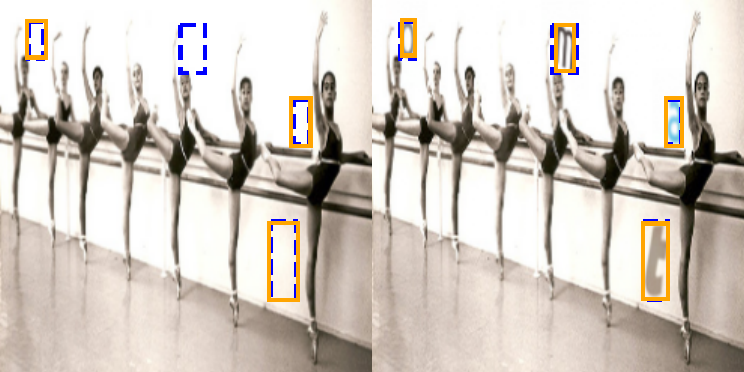}
&\includegraphics[width=0.20\linewidth]{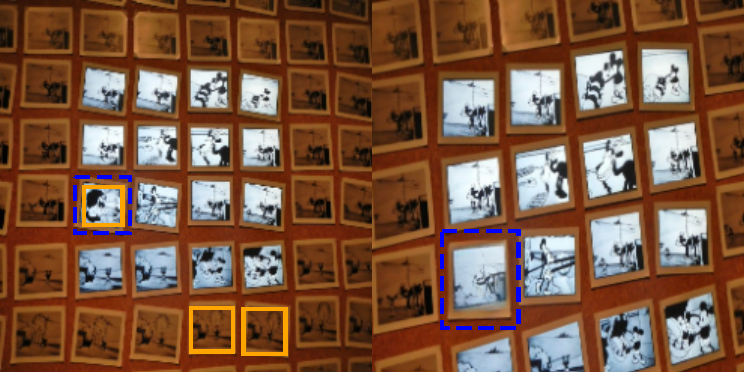}\\
\rotatebox{90}{\hspace{0.4cm}Ours} 
&\includegraphics[width=0.20\linewidth]{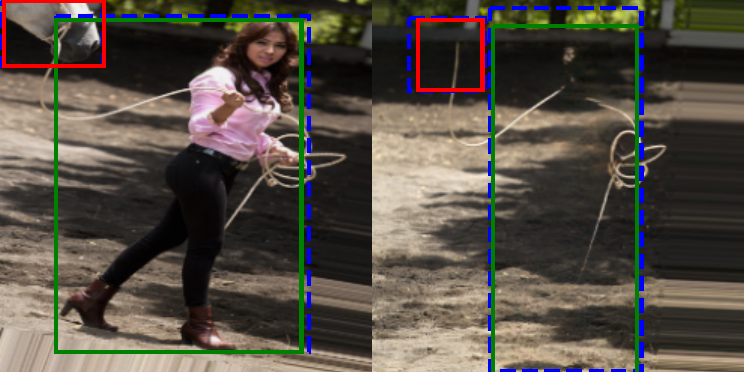} &\includegraphics[width=0.20\linewidth]{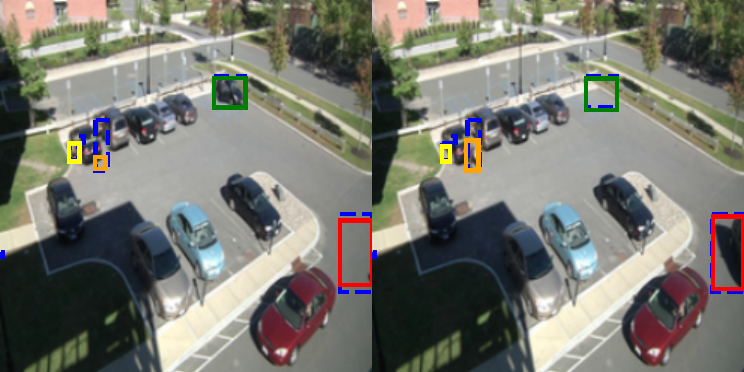} 
&\includegraphics[width=0.20\linewidth]{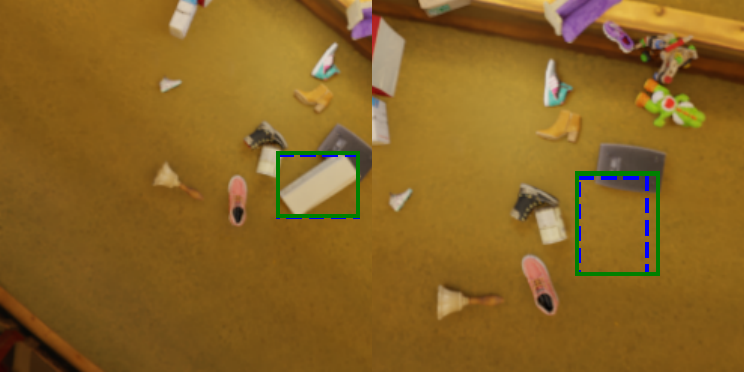}
&\includegraphics[width=0.20\linewidth]{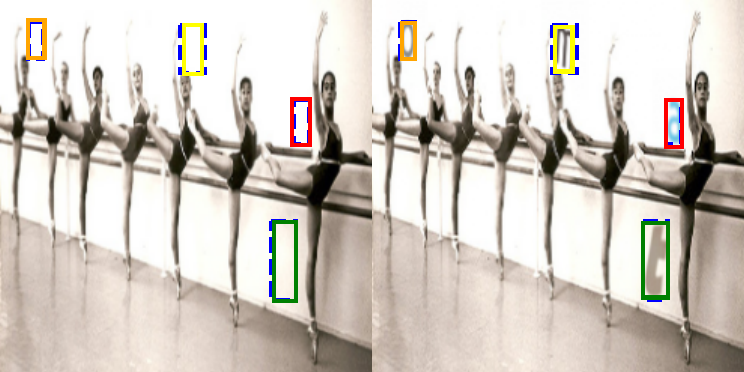}
&\includegraphics[width=0.20\linewidth]{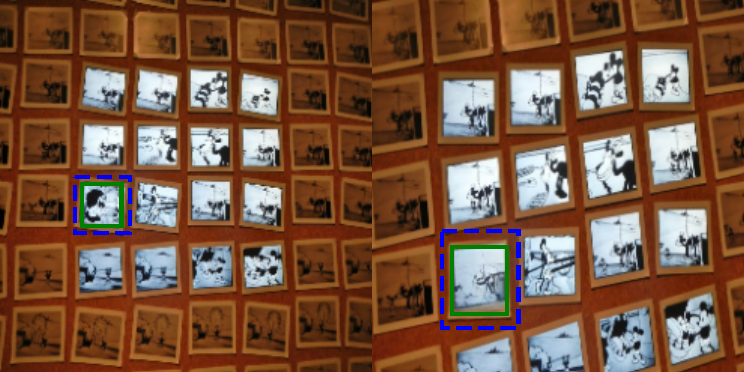}\\
\end{tabular}}
\captionof{figure}{
     At an optimal confidence threshold, CYWS \cite{sachdeva2023change} (top row) sometimes still produces false positives---\textcolor{orange}{$\Box$} in (a) \& (c)---and fails to detect changes (a).
     Dashed \textcolor{blue}{- - -} boxes show groundtruth changes.
     First, we encourage detectors to be more aware of changes via a novel contrastive loss.
     Second, our Hungarian-based post-processing
     reduces false positives (a), improves change-detection accuracy (b), and estimates correspondences (c--d), i.e., paired changes such as (\textcolor{red}{$\Box$}, \textcolor{red}{$\Box$}) and (\textcolor{green!60!black}{$\Box$}, \textcolor{green!60!black}{$\Box$}).
     Our work (bottom row) is \emph{the first} to estimate change correspondences compared to prior works \cite{sachdeva2023change,sachdeva2023change_3d,zheng2024segment} (top row). 
     More qualitative results in (\cref{fig:more qualitative results betwen cyws and our model})
     .} 
\label{fig:teaser}
\vspace{-0.2cm}
\end{center}
}]

\begin{abstract}
Detecting object-level changes between two images across possibly different views (\cref{fig:teaser}) is a core task in many applications that involve visual inspection or camera surveillance.
Existing change-detection approaches suffer from three major limitations: (1) lack of evaluation on image pairs that contain no changes, leading to unreported false positive rates; (2) lack of correspondences (\ie, localizing the regions before and after a change); 
and (3) poor zero-shot generalization across different domains.
To address these issues, we introduce a novel method that leverages change correspondences (a) during training to improve change detection accuracy, and (b) at test time, to minimize false positives. 
That is, we harness the supervision labels of where an object is added or removed to supervise change detectors, improving their accuracy over previous work \cite{sachdeva2023change} by a large margin.
Our work is also the first to predict correspondences between pairs of detected changes using estimated homography and the Hungarian algorithm.
Our model demonstrates superior performance over existing methods, achieving state-of-the-art results in change detection and change correspondence accuracy across both in-distribution and zero-shot benchmarks.
\end{abstract}

\section{Introduction}
\label{sec:intro}

\noindent Identifying key changes between two images is a core task that powers many applications \cite{sachdeva2023change}, \eg, to detect changes across brain scans \cite{dong2021deepatrophy,patriarche2004review}, a missing car in a parking lot (\cref{fig:teaser}) \cite{stent2015detecting, jhamtani2018}, or a defective product in a manufacturing pipeline \cite{wu2018spot}. However, existing work has three major limitations. First, most papers did not test on pairs of images where there are no changes \cite{sachdeva2023change,sachdeva2023change_3d} and therefore do not measure false positives. Many image-difference captioning benchmarks contain only \emph{change} cases \cite{tan2019image-editing-request} or only a small subset of \emph{no-change} examples, e.g., ~10\% of Spot-the-Diff \cite{jhamtani2018}.
Second, prior models are trained to detect only changes; yet, such detected changes are not too usable in the downstream application when there are many changes predicted per image but no correspondence provided (\cref{fig:teaser}). Third, many image-difference prediction works are specialized for a single domain (e.g., remote sensing \cite{xu2021change}) and do not measure zero-shot generalization to unseen datasets \cite{xu2021change,zheng2024segment}.

In this paper, we address these three problems by leveraging change correspondences during (a) training to improve change detection precision, and (b) test time to reduce false positives.
Compared to prior works \cite{sachdeva2023change,sachdeva2023change_3d,zheng2024segment}, \textbf{our work is the first to predict correspondences in addition to the changes} (\cref{fig:teaser}).
That is, we propose a {post-processing} algorithm based on an estimated homography and the Hungarian algorithm \cite{kuhn1955hungarian,carion2020end} to reduce false positives (\cref{fig:teaser}) of a state-of-the-art change detector \cite{sachdeva2023change}.
Intuitively, first, we run a pre-trained change detector \cite{sachdeva2023change} on a pair of images to collect a set of predicted changes on each image.
Then, we project the predicted boxes in image 1 onto image 2 and filter out those that do not substantially overlap with any predicted changes in image 2, arriving at higher-precision change predictions (\cref{fig:alignment}).
A similar procedure is used to filter out the predicted changes in image 2.
We then harness the Hungarian algorithm to predict correspondences in addition to the predicted changes (\cref{fig:teaser}).
Our main findings are:\footnote{Code and data are available on \href{https://github.com/anguyen8/image-diff}{github}.}

\begin{enumerate}
    \item Leveraging correspondence labels in finetuning detectors leads to state-of-the-art change detectors, outperforming CYWS\cite{sachdeva2023change} by a large margin (from \increasenoparent{1.05} to \increasenoparent{9.04} in mAP) on all five benchmarks (\cref{subsec:no change}).
    
    \item Our proposed contrastive matching loss function for finetuning change detectors also improves the accuracy in predicting correspondences, from \increasenoparent{1.31} to \increasenoparent{6.56}), on all five benchmarks (\cref{sec:corr pred}).
    \item Moreover, we have established a new metric for evaluating matching scores among different models, facilitating a consistent and comparative assessment of change detection performance (\cref{sec:metrics}).
    
    \item We present OpenImages-Inpainted, \ie, a novel change detection dataset with $\sim$1.3M image pairs, where image pairs consist of exactly 1 change derived from realistic scenes of the OpenImages \cite{openimages} dataset. Our OpenImages-Inpainted has no view-transformation artifacts (compared to COCO-Inpainted \cite{sachdeva2023change}) and minimal inpainting artifacts (compared to Img-Diff \cite{jiao2024img}).
    
\end{enumerate}

\section{Problem formulation}

\subsection{Definition of Changes}

\begin{figure*}[t]
     \centering
     \begin{tabular}{c@{}c@{}c@{}c}
     & {\scriptsize (a) viewpoint, colors } & {\scriptsize (b) viewpoint} &{ \scriptsize (c) move = \textcolor{green!40!black}{remove} + \textcolor{red}{add} }\\ 
     &\includegraphics[width=0.33\linewidth]{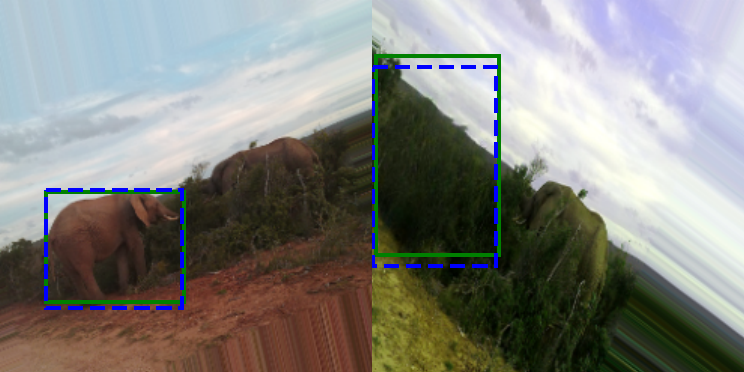} &\includegraphics[width=0.33\linewidth]{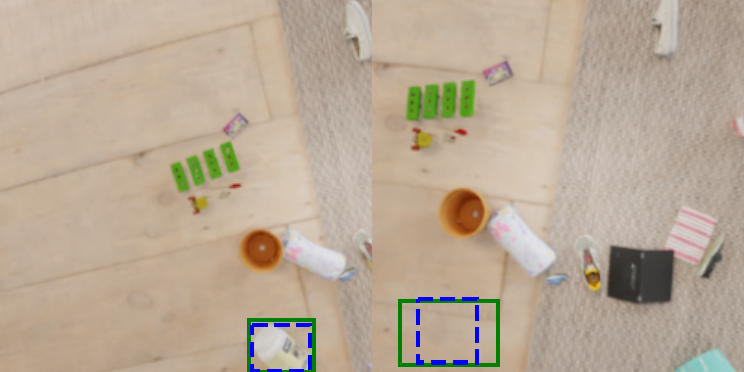}
     &\includegraphics[width=0.33\linewidth]{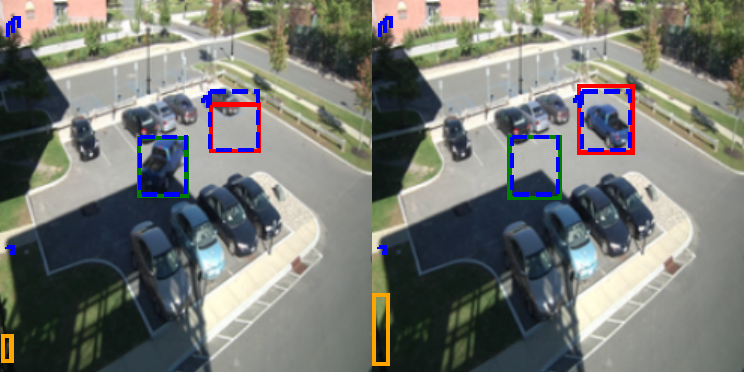} \\
     \end{tabular}
     \vspace{-0.45cm}
     \caption{
     \textbf{Example predicted changes}.
     In \cocologo, our method detects a removal of an \textcolor{green!40!black}{elephant} (a) despite viewpoint and color differences between two images.
     A tiny, white object (b) removal is detected despite viewpoint differences in \kubriclogo.
     In \stdlogo, a blue truck moving from one location to another is correctly detected as two changes: a \textcolor{green!40!black}{removal} and an \textcolor{red}{addition}.
     Colored, solid-bordered boxes, \eg, (\textcolor{red}{$\Box$}, \textcolor{red}{$\Box$}), (\textcolor{green!60!black}{$\Box$}, \textcolor{green!60!black}{$\Box$}), (\textcolor{orange}{$\Box$}, \textcolor{orange}{$\Box$}), show predicted \emph{paired} changes.
     }
     \label{fig:prob def}
\end{figure*}

\noindent We define a change to be an addition, absence, or modification of an \textbf{object} in one image compared to the other (see \cref{fig:prob def}a--b).
A major challenge is to detect such object-level changes in the presence of changes in camera viewpoint (COCO-I, KC), colors (COCO-I), or lighting (STD), which we do not aim to detect.
The objects that change include humans, animals (COCO-I), man-made objects (COCO-I, STD, KC, OI), and letters (SC).

In the case where the same object \emph{moves} 
from one location to another (\cref{fig:prob def}c) across two images, we expect two changes to be detected: (1) An object is removed from the first location in image 1, and (2) an object is added to the second location in image 2.
That is, two pairs of corresponding changes are to be predicted.


\subsection{Five benchmarks}
\label{sec:datasets}
\noindent Following \cite{sachdeva2023change}, we train and test both our model and CYWS \cite{sachdeva2023change} on COCO-Inpainted. 
Additionally, we test these models \textbf{zero-shot} on four \emph{unseen} change-detection benchmarks: STD \cite{jhamtani2018}, Kubric-Change \cite{sachdeva2023change}, and Synthtext-Change \cite{sachdeva2023change} and our proposed OpenImages-Inpainted.

\noindent\textbf{COCO-Inpainted} (COCO-I) \cocologo \cite{sachdeva2023change} contains 57K, 3K, and 4.5K image pairs in the train, validation, and test sets, respectively.
In each pair, one image is originally from COCO and the other is a clone with $N$ objects removed ($1 \leq N \leq 24$) from the image.
The test set is divided into three groups based on the size of removed objects: small (38\%), medium (39\%), and large (23\%) (see \cref{fig:prob def}a). 
Images are subjected to random affine transformations and color jittering. 
Combined with cropping, these modifications yield image pairs, where all objects may not appear in both images.

\noindent\textbf{VIRAT-STD} (STD) \stdlogo \cite{jhamtani2018}~~~A random 1,000 pairs of images (see \cref{fig:prob def}b) is selected from the Spot-the-Difference dataset \cite{jhamtani2018}, a dataset of camera surveillance images of street views.
Two images in each pair have almost identical views but are taken at different times.
Objects being changed are typically humans and cars.

\noindent\textbf{Kubric-Change} (KC) \kubriclogo \cite{sachdeva2023change} comprises 1,605 test cases (see \cref{fig:prob def}c). The scenes comprise a randomly chosen assortment of 3D objects on a ground plane with a random texture. After applying the change to the scene, the camera's position in the 3D space slightly moves, yielding two different views of the scene.

\noindent\textbf{Synthtext-Change (SC) \synthlogo} \cite{sachdeva2023change} consists of 5K pairs of real images with \emph{N} changes, where $1 \leq N \leq 6$. Each change includes an arbitrary letter synthetically placed on one image at random locations. 

\noindent\textbf{OpenImages-Inpainted (OI) \openlogo} 
To address the view-transformation and inpainting artifacts in \cocologo~and Img-Diff \cite{jiao2024img},  we create $\sim$1.3M pairs of images containing exactly 1 change. We adopt the original images from OpenImages dataset \cite{openimages}, and remove a single object using LaMa \cite{suvorov2021resolution} inpainter, similar to COCO-I \cite{sachdeva2023change}. We filter the object sizes based on their relative bounding box area to the image size and keep the objects that fall within the range of 0.01 to 0.04. This ensures that objects are neither tiny nor overly large. We rotate a random image in the pair within the range of [-10, 10] degrees, and then apply random croppings to generate viewpoint differences. We use a 5K subset for testing and the remaining images for training.

\subsection{Evaluation metrics}
\label{sec:metrics}

\subsec{Change detection evaluation}
We use object localization metrics to evaluate the accuracy of change detection. Specifically, we follow CYWS and use mAP for the top $k=100$ predicted boxes with the highest confidence scores in PASCAL VOC \cite{padillaCITE2020} style (see the \href{https://github.com/ragavsachdeva/The-Change-You-Want-to-See/blob/main/models/centernet_with_coam.py#L50}{code}). We threshold bounding boxes by confidence scores to filter out low-confidence predictions, ensuring better precision by reducing the false positives (FP) for mAP.

\subsec{Correspondence evaluation}
Common metrics for change detection focus on detecting changed objects, ignoring the correspondence information \cite{sachdeva2023change_3d,sachdeva2023change}. 
We are the first to adopt the F1 score ($\frac{\mathrm{2\times Precision\times Recall}}{\mathrm{Precision+Recall}}$) from the classification tasks to evaluate the correctness of the correspondence predictions in the change detection task. That is, we compare each predicted pair of correspondence to the ground truth corresponding boxes to see if they match.

\noindent We define True Positive (TP), False Positive (FP), and False Negative (FN):
\begin{itemize}
    \item \textbf{TP:} For each predicted box in the correspondence pair, we calculate their IoU with ground truth. If the IoU value for each predicted box in the pair is $\geq0.5$, the pair is labeled TP. If more than 1 prediction matches the same ground truth, the pair with the highest IoU value, for each box in the pair, is assigned TP.

    \item \textbf{FP:} Any predicted pair in which either one or both of its boxes does not meet the $IoU \geq 0.5$ criteria is labeled FP. If the boxes in the pair meet the criteria but do not have the highest IoU, we label them FP.
    \item \textbf{FN:} For each ground truth box, if there is no predicted pair with a nonzero IoU, we label the prediction FN.
\end{itemize}


\section{Methods}
\noindent CYWS \cite{sachdeva2023change}, a SOTA change detection model, is a U-Net coupled with CenterNet head \cite{duan2019centernet} to detect changes in two images. That is, they predict $100$ boxes per image, assuming the images always contain changes.
Here, we summarize the current problems with this method that limit its real-world applications.
\begin{enumerate} 
    \item They assume that each image pair always contains $\geq 1$ changes. However, there are many cases in the real world where no changes exist.
    \item Given that they only detect boxes and the correspondence information is not predicted, it is not trivial to understand and relate its predictions with each other across two images when several changes are present (see \cref{fig:teaser}).
\end{enumerate}
In this work, we aim to solve these problems, \ie, we address both \textbf{change} and \textbf{no-change} scenarios and predict a correspondence between the changed objects.

\begin{figure*}
    \centering
    \includegraphics[width=\textwidth]{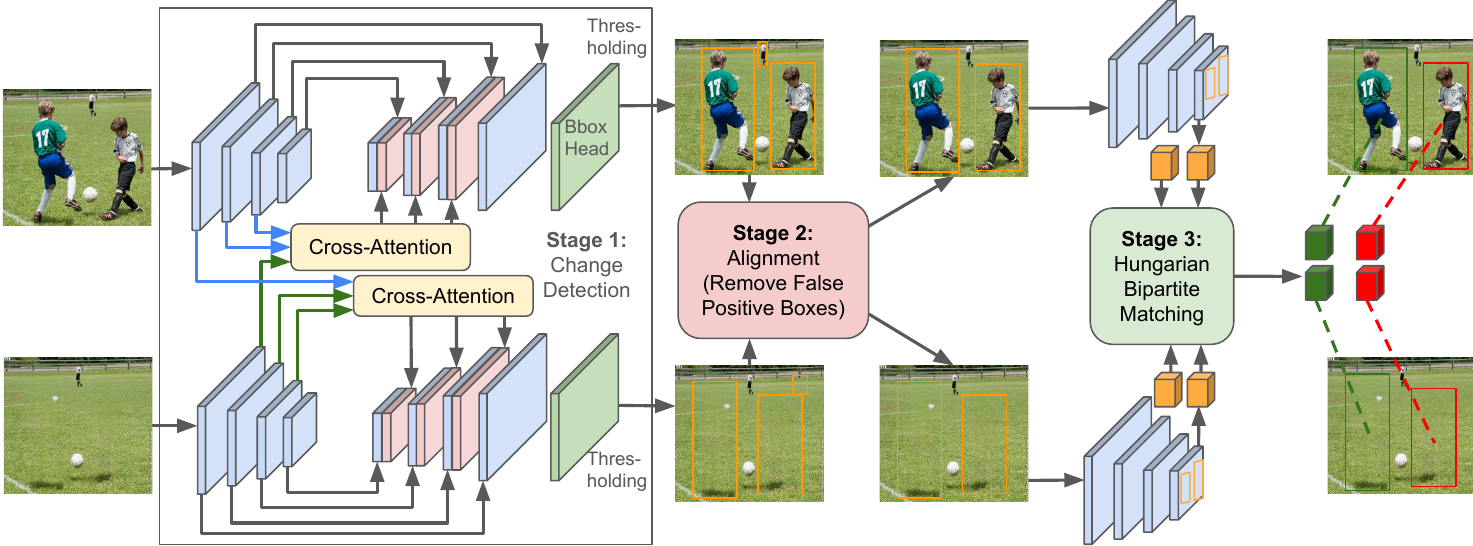}
    \caption{\textbf{Our approach of detecting changes and predicting their correspondence.} Our approach comprises three major stages. The first is the change detector, which we employ from the CYWS paper to identify changes between two images. The second is the alignment step, where an ideal detection threshold is established before forwarding anticipated boxes to the alignment process, aiding in the removal of false positive predicted boxes. The third is the matching algorithm, which takes the output from the alignment step to determine the correspondence between each pair of changes between the two images.}
    \label{fig:sol}
\end{figure*}

\noindent Our change detection pipeline consists of 3 stages: (1) change detection backbone, (2) alignment, and (3) correspondence prediction. Given a pair of images in Stage 1 (\cref{fig:sol}), a change detector \eg, CYWS \cite{sachdeva2023change}, detects boxes over the changed objects. Then, in Stage 2, we aim to reduce the false positive predictions and remove the boxes that are poor candidates for correspondence prediction via an alignment stage. Finally, we use the Hungarian algorithm with a contrastive matching loss to predict the correspondences in Stage 3. 


\subsection{Stage 1: Change detection}
\noindent The change detection backbone identifies change locations between left and right images using bounding boxes. After applying an optimal detection threshold to filter out predicted boxes with low confidence scores, the remaining bounding boxes are used as input for Stage Two and Stage Three (\cref{fig:sol}).

\noindent \textbf{U-Net encoder} The ResNet-50 architecture is employed as the encoder backbone. The input image has a shape of $3\times256\times256$. The output of the last layer (Layer 4.2) has a shape of $8\times8\times2048$.

\noindent \textbf{Cross-attention} The cross-attention module facilitates information exchange between left and right images, enabling accurate computation of changes between the two. This process generates three feature maps with shapes $8\times8\times4096$, $16\times16\times2048$, and $32\times32\times1024$, respectively. 

\noindent \textbf{U-Net decoder} The decoder utilizes the three feature maps produced by the cross-attention module as input and upsamples them to generate feature maps with a shape of $64\times256\times256$. Skip connections from the encoder and scSE \cite{roy2018concurrent} blocks are incorporated into the upsampling process. The decoder output passes to the Bbox head.

\noindent \textbf{Bbox head}: The bbox head employs CenterNet \cite{duan2019centernet} to predict bounding boxes for the detected change regions in the two images. CenterNet produces three output maps: center map ($1\times256\times256$), offset map ($1\times256\times256$), and and height-width map ($2\times256\times256$).

\subsection{Stage 2: Alignment}
\label{sec:align}
\noindent Thresholding the box predictions with the confidence score significantly reduces the number of false positives (\cref{tab:add align results}). However, we take an additional step to eliminate boxes that are poor candidates for alignment. The alignment stage is based on the premise that if a predicted change box appears in the left image, there should be a corresponding box in the right image. This helps us refine candidate boxes for the subsequent matching stage in our proposed solution.

\noindent To identify the alignment box of a candidate box in the left image, we first determine the transformation matrix—an affine transformation between the two images. We use SuperGlue \cite{sarlin2020superglue} to establish point correspondences between the images, and apply RANSAC to eliminate outliers. 
We use SuperGlue because of its lightweight and high accuracy point-matching performance \cite{sarlin2020superglue}.

\begin{figure}[hb]
    \centering
    \begin{subfigure}{0.46\textwidth}
        \includegraphics[width=1.0\columnwidth]{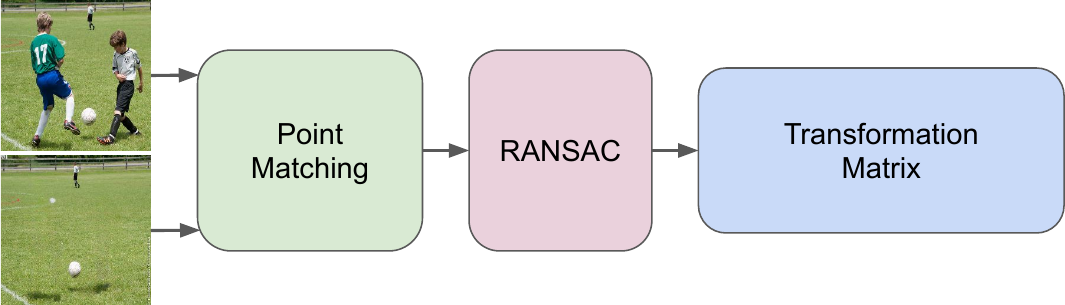}
        \caption{Estimate transformation matrix}
        \label{fig:estimate}
    \end{subfigure}
    \begin{subfigure}{0.46\textwidth}
        \includegraphics[width=1.0\columnwidth]{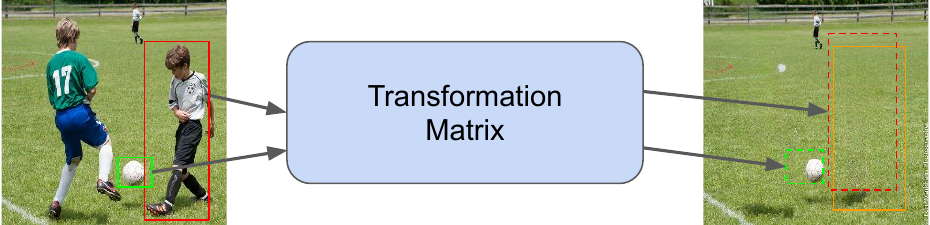}
        \caption{Estimate bounding boxes locations in left image to right image}
        \label{fig:align box}
    \end{subfigure}
    \caption{\textbf{Alignment Overview.} (\cref{fig:estimate}) illustrates the process for estimating the homography matrix. In (\cref{fig:align box}), dashed \textcolor{red}{$\Box$} represent the corresponding box in the right image of  \textcolor{red}{$\Box$} in the left image. Similarly for \textcolor{green}{$\Box$} box.}
    \label{fig:alignment}
\end{figure}

\noindent A candidate box is valid if its alignment overlaps with any box in the other image (\textsc{IoU} $> 0$). Otherwise, it is invalid and excluded. Two candidate boxes (red and green) in the left image are aligned with corresponding dashed boxes in the right image. The red box's alignment overlaps with the orange box, making it valid, while the green box's alignment does not overlap and is discarded (\cref{fig:align box}).

\subsection{Stage 3: Correspondence prediction}
\label{sec:corr}
\noindent The alignment stage combined with the confidence thresholding substantially reduces the false positives, yielding improved mAP on five benchmarks (see \cref{tab:ablation study loss}). Yet, the lack of correspondence information remains unsolved (see the outputs of Stage 2 in \cref{fig:sol}).

\noindent Here, we aim to predict the correspondence between the predicted boxes for each image pair given the embeddings of each box. That is, we first extract embeddings from the feature maps in the backbone (Stage 1) for each aligned box of Stage 2. Then, we use the Hungarian bipartite matching algorithm jointly with a contrastive matching loss to predict the final correspondence. 

\subsec{Box embedding extraction} 
Since each predicted box intersects with $\geq1$ image patches in the feature maps of Stage 1, we use 2 different methods to extract the box embeddings, and choose the best one based on the mAP score in \cref{sec:crop vs avg}:
\begin{enumerate}
    \item \textbf{Mean pooling Method}: We hypothesize that the mean of patches associated with a predicted box enriches the correspondence embedding vector of the box with contextual information surrounding the object. We input each image (of size $256\times256$) into the image encoder to obtain a feature volume of $8\times8\times2048$. From the $8\times8 = 64$ patch embeddings, we select all N embeddings corresponding the patches that overlap with a given bounding box in the input image space. Then, we take the mean of the N embeddings to obtain final embedding of size $2048$ (\href{https://github.com/anguyen8/image-diff/blob/main/models/centernet_with_coam.py#L264C52-L264C70}{code}).
    \item \textbf{Region Cropping Method}: This method evaluates whether excessive contextual information surrounding an object negatively impacts the quality of the embedding vector. To address this, only the information within the predicted bounding box is utilized. We crop the input image to bounding-box region to create a cropped image (\href{https://github.com/anguyen8/image-diff/blob/main/models/centernet_with_coam.py#L1860}{code}). We feed the cropped image into a ResNet-50 image encoder and average the $8\times8\times2048$ feature output from layer 4.2 to obtain a $2048$ dimensional embedding. 
\end{enumerate}

\noindent A key challenge in implementing change detection in real-world scenarios is identifying the correspondence between changes detected in two images. 
We use the Hungarian algorithm to match the predicted bounding boxes between the two images. Given, $e_{i}$, and $e_{j}$, the embeddings of two bounding boxes from (\cref{sec:corr}) we calculate a cost matrix using the ground distance (\cref{eq:cost}) similar to \cite{phan2022deepface}.

\begin{equation}
    d_{ij} = 1-\frac{\langle e_{i} \cdot e_j\rangle}{\lVert e_{i}\rVert \lVert e_{j}\rVert}
    \label{eq:cost}
\end{equation}

\noindent where $i, j$ are indices of matrix elements. Using the cost matrix the Hungarian algorithm assigns the correspondence between boxes from the first and second image such that the total cost is minimum.

\begin{figure}
    \centering
    \includegraphics[width=1.0\columnwidth]{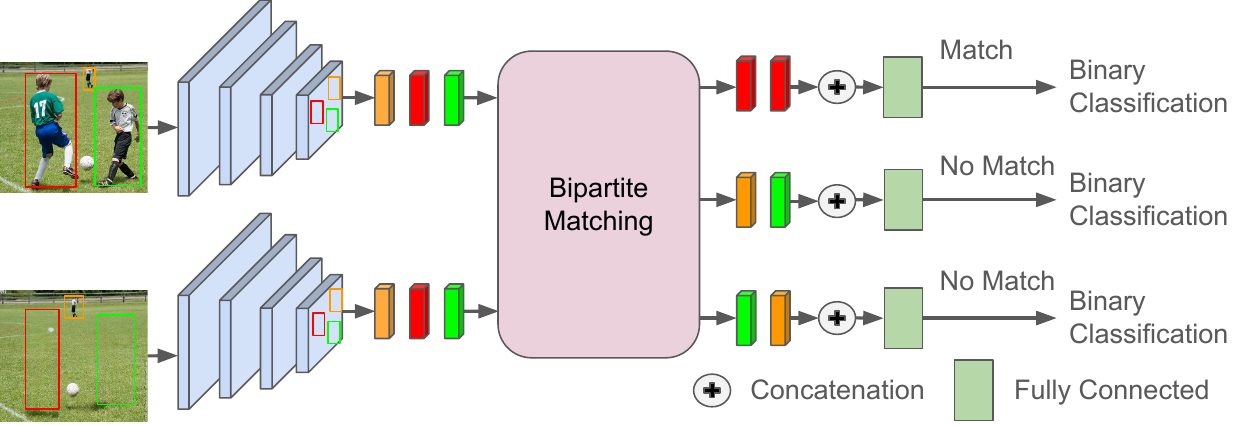}
    \caption{\textbf{Contrastive matching loss.}
    Contrastive matching loss trains the model to distinguish between positive and negative pairs of image patches. Each ground-truth bounding box is assigned an embedding from the last layer of encoder, and the Hungarian algorithm is employed to establish matches between ground-truth boxes across two images. A matched pair is labeled as positive if it aligns with the correspondence ground truth; otherwise, it is designated as negative. Embeddings of positive and negative pairs are concatenated and passed through a fully connected layer to compute BCELoss.
    }
    \label{fig:contrastive matching loss}
\end{figure}

\subsec{Contrastive matching loss} We use the contrastive matching loss to train the model to classify pairs of matched boxes obtained from the Hungarian algorithm. 
The Hungarian algorithm is not perfectly accurate, \ie, it achieves an F1 score of 91.68\% on the \cocologo ~dataset (see \cref{sec:Upper bound accuracy of correspondence algorithm app}) when using the ground truth boxes, and it generates both negative and positive matchings. Specifically, we compare the matched boxes with the correspondence information in the ground truth, classifying them as either: (1) positives, \ie, they match the ground truth, or (2) negatives, \ie, they do not. We leverage this fact and train the model using our contrastive matching loss. First, the embeddings of matched pairs are concatenated and processed through a fully connected layer. Then, we use a binary classification loss (\href{https://pytorch.org/docs/stable/generated/torch.nn.BCELoss.html}{BCELoss}), treating the matched pairs as predictions and the ground-truth correspondence as targets.

\noindent The final training objective in (\cref{eq:final}) consists of two main components: (1) object detection loss and (2) contrastive matching loss. The detection loss integrates center-based loss components, ensuring precise localization and classification.
\begin{equation}
\label{eq:final}
    L_\text{total} = L_\text{CenterNet} + \alpha L_\text{DETR} + \beta L_\text{contrastive}
\end{equation}
where $L_\text{CenterNet}$ is the CenterNet detection loss \cite{duan2019centernet}, and $L_\text{DETR}$\cite{carion2020end} is the combination of $L_{1}$ loss and \textsc{GIoU} loss, $L_\text{contrastive} $ is our contrastive matching loss. The comprehensive analysis of each loss component shows in (\cref{sec:loss ablation study}).

\subsection{Training hyperparameters}
\label{sec:training params}
\noindent This section specifies the training hyperparameters. We fine-tuned the CYWS change detector \cite{sachdeva2023change} using contrastive matching loss and DETR loss \cite{carion2020end}, leveraging the pre-trained CYWS model. Transformation estimation followed the method in \cite{sachdeva2023change_3d}. The fine-tuning process ran for 200 epochs on four A100 GPUs with a batch size of 16, optimized using the Adam algorithm \cite{kingma2014adam} with a learning rate of $0.0001$ and weight decay of $0.0005$. The final loss (\cref{eq:final}) used $\alpha = 3$ and $\beta = 2$. A detailed analysis of hyperparameter selection is provided in (\cref{sec:training_hyper_app}).

\vspace{-0.2cm}
\section{Hyperparameters Tuning}
\subsection{Training contrastive matching loss with only ground-truth achieves the highest mAP}
\noindent We analyze the impact of assigning embeddings from ground-truth boxes or predicted boxes as inputs to the Hungarian algorithm on the matching process (\cref{fig:sol}).

\subsec{Experiments} 
We evaluate three embedding assignment methods for training the contrastive matching loss. The first method assigns embeddings exclusively to predicted bounding boxes. The second method assigns embeddings only to ground-truth bounding boxes. The third method, a hybrid approach, utilizes both ground-truth and predicted bounding boxes, where predicted embeddings are passed to the Hungarian algorithm, and correctly assigned matches replace predicted embeddings with their corresponding ground-truth embeddings, while incorrect assignments retain the original predicted embeddings. In all methods, the output from the Hungarian algorithm is used to compute the contrastive matching loss (\cref{fig:contrastive matching loss}).

\subsec{Results} The method relying solely on only ground-truth boxes achieves the highest mean average precision (mAP) across the \cocologo, \kubriclogo, and \openlogo ~datasets. In contrast, combining ground-truth and anticipated boxes results in improved mAP for both the \stdlogo ~and \synthlogo ~datasets. However, due to variations in viewpoint across the \cocologo, \kubriclogo, and \openlogo ~datasets, leveraging predicted boxes for feature embeddings proves insufficient for accurately capturing differences between paired images. Consequently, the mAP is reduced when using predicted boxes alone, compared to using solely ground-truth boxes (\cref{tab:ground-truth vs predicted}). 
\begin{table}[h]
\centering
\resizebox{\columnwidth}{!}{
\begin{tabular}{lccccccc}
\toprule
Model & Predicted & Ground-Truth & \cocologo & \stdlogo & \kubriclogo & \synthlogo & \openlogo \\
\midrule
CYWS\cite{sachdeva2023change}        &       &      & 62.73          & 53.98            & 76.13           & 88.97     & 69.70\\
\text{Our}        & \checkmark     &       & 69.60          & 56.08            & 79.95           & 89.60       & 75.43\\
\text{Our}        &      &\checkmark        & \textbf{71.77}          & 57.00            & \textbf{81.07}           & 90.02       &\textbf{78.84}\\
\text{Our}       & \checkmark          & \checkmark            & 70.86          & \textbf{57.26}            & 79.28           & \textbf{90.44}       & 76.57\\
\bottomrule
\end{tabular}}
\caption{\textbf{Ground-truth} bounding boxes are crucial for training contrastive matching loss effectively. Relying solely on predicted bounding boxes or combining them with ground-truth bounding boxes significantly degrades the model's \textbf{mAP} accuracy.}
\label{tab:ground-truth vs predicted}
\end{table}

\subsection{Encoder feature maps yield better localization than decoder feature maps}
\noindent To compute the cost matrix for the Hungarian algorithm in Stage Three, features are extracted from the encoder or decoder in Stage One to generate embeddings for each predicted bounding box (\cref{fig:sol}). We hypothesize that using features from different decoder layers allows the extraction of multi-scale information, resulting in embeddings with richer representations compared to those generated solely from the encoder's output. Specifically, we evaluate features obtained from the output of the encoder's final layer and the outputs of the first three initial layers of the decoder.

\subsec{Experiments} In the first experiment, the output from Layer 4.2 of the encoder (ResNet-50) is used, resulting in an embedding of size $2048$ being assigned to each predicted bounding box. In the second experiment, the feature volumes from the first three initial layers of the decoder, with dimensions $8\times8\times4096$, $16\times16\times2048$, and $32\times32\times1024$, respectively, are used. We concatenate embedding extracted from three decoder layers to form the final embedding of size $7168$ (\href{https://github.com/anguyen8/image-diff/blob/main/models/centernet_with_coam.py#L1871}{code}). 

\subsec{Results} Using the embeddings from the decoder layer does not lead to a better \textbf{mAP} score (see \cref{tab:encoder vs decoder}) across all datasets. The feature map obtained from the encoder has a higher value of $\increasenoparent{0.97}$ in the \cocologo ~dataset, $\increasenoparent{2.19}$ in the \kubriclogo ~dataset, and $\increasenoparent{1.01}$ in the \synthlogo ~dataset. However, on the \stdlogo ~dataset, it yields a marginal improvement of only $\increasenoparent{0.03}$. Therefore, we use the \textbf{Encoder} feature map for our finetuned model.

\begin{table}[h]
\resizebox{\columnwidth}{!}{
\centering
\begin{tabular}{lccccccc}
\toprule
Model & Encoder & Decoder & \cocologo & \stdlogo & \kubriclogo & \synthlogo & \openlogo \\
\midrule
\text{Our}        &         &\checkmark          & 70.80          & \textbf{57.03}            & 78.88           & 89.01      
 &78.58\\
\text{Our}     &\checkmark       &            & \textbf{71.77}          & 57.00            & \textbf{81.07}           & \textbf{90.02}       &\textbf{78.84}\\
\bottomrule
\end{tabular}}
\caption{Feature embeddings derived from \textbf{encoder} feature maps outperform those derived from decoder feature maps on mAP.}
\label{tab:encoder vs decoder}
\end{table}


\vspace{-0.2cm}
\subsection{Ablation study of loss function}
\label{sec:loss ablation study}
\noindent Our fine-tuning loss (\cref{eq:final}) has three components: CenterNet loss, DETR loss, and our novel contrastive matching loss. Here, we run an ablation study to show that all three losses contribute to the final result.

\subsec{Experiments} We conduct fine-tuning experiments on the CYWS model under various configurations. In the first setup, we used only the CenterNet loss and DETR loss for training. In our ablation study, we fine-tune the model under different configurations to evaluate the impact of the DETR loss and the Contrastive matching loss on change detection performance. Specifically, we experiment with our model fine-tuned with and without the DETR loss, as well as with and without the Contrastive matching loss. For all these experiments, the models were initialized with weights derived from the pre-trained CYWS model.


\subsec{Results} We find that DETR loss contributes improvements of  \increasenoparent{7.02}, \increasenoparent{2.45}, \increasenoparent{2.95}, \increasenoparent{0.79},
\increasenoparent{6.73} in the \cocologo, \stdlogo, \kubriclogo, \synthlogo, \openlogo ~datasets, respectively, compared to using CenterNet alone. Similarly, the Contrastive matching loss leads to enhancements of $\increasenoparent{6.7}$, $\increasenoparent{3.02}$, $\increasenoparent{4.94}$, $\increasenoparent{1.11}$, and $\increasenoparent{3.49}$ across the same datasets, respectively, compared to CenterNet alone. Adding all three losses results in the highest mAP across 4 out of 5 datasets.

\begin{table}[ht]
\centering
\resizebox{\columnwidth}{!}{
\begin{tabular}{lcccccccc}
\toprule
Model &CenterNet &DETR &Contrastive & \cocologo & \stdlogo & \kubriclogo & \synthlogo & \openlogo  \\
\midrule
CYWS \cite{sachdeva2023change}      &\checkmark     &   &   & 62.73      & 53.98         & 76.13     & 88.97    & 69.70 \\
\text{Our} &\checkmark    &\checkmark   &        & 69.75          & 56.43            & 79.08          & 89.76      & 76.43\\
\text{Our} &\checkmark     &     &\checkmark        & 69.43          & 56.07            & 79.11           & \textbf{90.08}       & 73.19  \\
\text{Our} &\checkmark     &\checkmark     &\checkmark          &
\textbf{71.77}          & \textbf{57.00}            & \textbf{81.07}           & 90.02          & \textbf{78.84}\\
\bottomrule
\end{tabular}}
\caption{\textbf{Loss ablation study.} contrastive matching loss significantly improves mAP score across all datasets, achieving substantial improvements and remaining effective when combined with other losses, such as DETR loss.}
\label{tab:ablation study loss}
\end{table}


\section{Results}
\subsection{Given the same performance on no-change cases, our finetuned detector outperforms state-of-the-art CYWS}
\label{subsec:no change}
\noindent Real-world applications require models to perform well in both the change and no-change cases. 
We test our hypothesis that visual correspondence (\ie, the binary supervision labels of whether two image patches contain a change or not) improves change detection accuracy. We evaluate change detection performance by applying a detection threshold to ensure that the average number of predicted boxes per image in no-change cases remains below 0.01. This is a critical consideration for practical deployment, which has been overlooked in prior work \cite{sachdeva2023change}.

\subsec{Experiments} We initialize our model with the pre-trained CYWS weights and fine-tune it (\cref{sec:training params}) using the $L_{total}$ loss function (\cref{eq:final}). We evaluate mean Average Precision (mAP) on five datasets (\cref{sec:datasets}) using both CYWS \cite{sachdeva2023change} and our models. 
We choose the optimal detection threshold at $0.25$ for both models, ensuring the average number of predicted boxes per image in the no-change case remains below \textbf{0.01}.

\subsec{Results} Our fine-tuned model outperforms CYWS \cite{sachdeva2023change} across all THREE post-processing STAGES (see \cref{tab:mAP}). Since we keep the fine-tuning strategy fixed and repeat the experiment with various post-processing techniques, we contribute the positive delta in mAP score to our contrastive matching loss (\cref{sec:corr}). That is, \textbf{our contrastive loss improves change detection performance across both change and no-change pairs compared to CYWS} \cite{sachdeva2023change}. This performance gap further increases across all five datasets when an optimal threshold is applied. For instance, in the \cocologo ~dataset, the margin increases from \increasenoparent{9.04} to \increasenoparent{10.97}.

\begin{table}[H]
\centering
\resizebox{\columnwidth}{!}{%
\begin{tabular}{lcccccccccc}
\toprule
\text{}  &Model &Det-thres &Align &Hung &Thres & \cocologo & \stdlogo & \kubriclogo & \synthlogo & \openlogo \\
\midrule
\text{}  &\textsc{CYWS}  &    &    &    &\na            & 62.73          & 53.98            & 76.13           & 88.97    & 69.70\\
\text{(a)}  &\text{Our} &    &    &    &\na            & \textbf{71.77}         &\textbf{57.00}            &\textbf{81.07}          &\textbf{90.02}      &\textbf{78.84}\\
\text{}   &    &    &    &    &       &$\increasenoparent{9.04}$         &$\increasenoparent{3.02}$            &$\increasenoparent{4.94}$           &$\increasenoparent{1.05}$
&$\increasenoparent{9.14}$\\
\midrule
\text{}  &\textsc{CYWS}  &\checkmark   &   &    & 0.25            & 51.25          & 43.40            & 65.46           & 86.38        & 61.68\\
\text{(b)}   &\text{Our}  &\checkmark  &   &        & 0.25            & \textbf{62.22}          & \textbf{47.80}            & \textbf{71.25}           & \textbf{88.08}        &\textbf{72.69}\\
\text{}  &   &   &   &        &             &$\increasenoparent{10.97}$          &$\increasenoparent{4.40}$            &$\increasenoparent{5.79}$           &$\increasenoparent{1.70}$
&$\increasenoparent{11.01}$\\
\midrule
\text{}   &\textsc{CYWS} &\checkmark  &\checkmark   &     & 0.25            & 47.83          & 40.44            & 64.98           & 84.76      & 44.39\\
\text{(c)}   &\text{Our} &\checkmark  &\checkmark   &       & 0.25            & \textbf{59.23}          & \textbf{44.62}            & \textbf{69.23}           & \textbf{86.59}        & \textbf{58.83}\\
\text{}   &   &   &   &       &            &$\increasenoparent{11.40}$          &$\increasenoparent{4.18}$            &$\increasenoparent{4.25}$           &$\increasenoparent{1.83}$
&$\increasenoparent{14.44}$\\
\midrule
\text{}   &\textsc{CYWS} &\checkmark  &\checkmark   &\checkmark        & 0.25            & 44.48          & 39.93            & 54.16           & 83.78      & 44.41\\
\text{(d)}   &\text{Our} &\checkmark  &\checkmark   &\checkmark       & 0.25            & \textbf{56.59}          & \textbf{44.67}            & \textbf{60.85}           & \textbf{85.86}        & \textbf{58.88}\\
\text{} &   &   &   &       &            &$\increasenoparent{12.02}$          &$\increasenoparent{5.34}$            &$\increasenoparent{6.69}$           &$\increasenoparent{2.08}$
&$\increasenoparent{14.47}$\\
\bottomrule
\end{tabular}
}
\caption{\textbf{Change detection (mAP)}. 
The mAP value is calculated using several settings. In this context, we have selected an optimal detection threshold that ensures the average number of predicted boxes per image is below \textbf{0.01} on the No-change cases.}
\label{tab:mAP}
\end{table}

\noindent Each additional stage, including detection threshold, alignment, and Hungarian matching, contributes to a monotonic reduction in false positives across all five benchmark datasets for both change cases (\cref{tab:corr f1}) and no-change cases.
On the \cocologo ~dataset, false positives are reduced by \decreasenoparent{1.045}. Similar reductions are observed across other datasets: a reduction of \decreasenoparent{1.474} on the \stdlogo ~dataset, \decreasenoparent{0.216} on \kubriclogo, \decreasenoparent{1.428} on \synthlogo, and \decreasenoparent{0.059} on \openlogo ~(\cref{tab:no change}).

\begin{table}[h]
\centering
\resizebox{\columnwidth}{!}{
\begin{tabular}{lccccccccc}
\toprule
Det-thres &Align &Hung &Thres & \cocologo & \stdlogo & \kubriclogo & \synthlogo & \openlogo\\
\midrule
   &    &      &\na            & 100          & 100            & 100           & 100          & 100\\
 \checkmark   &    &        & 0.1            & 1.640          & 1.737            & 1.372           & 1.731   
& 0.119 \\
 \checkmark   &\checkmark    &        & 0.1            & 1.253          & 1.055            & 1.219           & 1.058    &0.054\\
 \checkmark   &    &\checkmark        & 0.1            & 0.982          & 0.945            & 1.309           & 0.976     &0.060\\
 \checkmark   &\checkmark    &\checkmark        & 0.1            & \textbf{0.595} (\decreasenoparent{1.045})         & \textbf{0.263} (\decreasenoparent{1.474})          & \textbf{1.156} (\decreasenoparent{0.216})           & \textbf{0.303} (\decreasenoparent{1.428})       &\textbf{0.060} (\decreasenoparent{0.059})\\
\bottomrule
\end{tabular}}
\caption{\textbf{Average number of predicted Boxes per image for no change cases ($\downarrow$).} The effectiveness of applying detection threshold, alignment, and Hungarian in removing false positive predicted box in no change case of the CYWS model. See (\cref{fig:false positive predicted boxes reduction}) for qualitative examples.}
\label{tab:no change}
\end{table}

\subsection{The alignment stage plays a crucial role in the success of the matching algorithm}
\label{sec:corr pred}
\noindent Given two sets of boxes of predicted changes \cite{sachdeva2023change}, our Hungarian-based matching algorithm’s goal is to pair up corresponding changes. The alignment stage (\cref{fig:sol}) identifies pairs of corresponding boxes between two images and eliminates boxes that do not have a match. We aim to test the matching accuracy with and without the Alignment stage to understand its importance.  

\subsec{Experiment} We repeat our correspondence prediction algorithm (\cref{sec:corr}) on all five benchmarks with and without the Alignment stage.

\subsec{Results} 
We find that the Alignment stage plays a crucial role, responsible for \increasenoparent{34.57} in the \stdlogo ~dataset, \increasenoparent{29.27} in the \synthlogo ~dataset of CYWS model in matching accuracy. Similarly, in \stdlogo ~and \synthlogo ~datasets, our model's improvement is \increasenoparent{38.18} and \increasenoparent{30.04}, respectively (see \cref{tab:add align results}). See (\cref{fig:w/ and w/o alignment}) for qualitative results.

\begin{table}[h]
\centering
\resizebox{\columnwidth}{!}{
\begin{tabular}{lccccccccc}
\toprule
Model &Det-thres &Align &Hung &Thres & \cocologo & \stdlogo & \kubriclogo & \synthlogo & \openlogo \\
\midrule
\textsc{CYWS} &\checkmark   &   &\checkmark      & 0.25      & 39.72          & 18.78            & 56.70           & 54.73       & 55.91\\
\textsc{CYWS} &\checkmark   &\checkmark   &      & 0.25      & 38.40          & 53.22          & 63.93           & 83.83       & 57.59\\
\textsc{CYWS}  &\checkmark   &\checkmark   &\checkmark        & 0.25            & \textbf{41.79}          & \textbf{53.35} (\increasenoparent{34.57})              & \textbf{63.96}           & \textbf{84.00} (\increasenoparent{29.27})      & \textbf{57.67}\\
\midrule
\text{Our} &\checkmark   &   &\checkmark   & 0.25            & 45.77          &18.81            & 63.84           & 55.27      & 68.15\\
\text{Our} &\checkmark   & \checkmark   &   & 0.25            & 46.00          &56.06            & 68.98           & 85.22      & 69.45\\
\text{Our} &\checkmark   &\checkmark   &\checkmark        & 0.25         & \textbf{48.35}          & \textbf{56.99} (\increasenoparent{38.18})           & \textbf{69.72}           & \textbf{85.31} (\increasenoparent{30.04})      &\textbf{69.53}\\
\bottomrule
\end{tabular}}
\caption{\textbf{Alignment} stage contributes to the success of the matching algorithm based on the F1 score. See ({\cref{fig:w/ and w/o alignment}}) for qualitative examples.}
\label{tab:add align results}
\end{table}

\subsection{Contrastive matching loss improves change matching accuracy}
\label{sec:matching score}
\noindent The contrastive matching loss directs the model to focus on regions exhibiting changes in both images, filtering out false positives. This approach improves change detection accuracy and boosts the correspondence score relative to the CYWS model.

\subsec{Experiment} We evaluate our model and the CYWS model under three configurations: using a detection threshold, incorporating an alignment stage, and applying the Hungarian algorithm to detect changes across five datasets (\cref{sec:datasets}). The matching score was computed with and without alignment on these datasets.
\begin{table}[h]
\centering
\resizebox{\columnwidth}{!}{
\begin{tabular}{lccccccccc}
\toprule
Model &Det-thres &Align &Hung &Thres & \cocologo & \stdlogo & \kubriclogo & \synthlogo & \openlogo \\
\midrule
\textsc{CYWS} &\checkmark   &   &\checkmark      & 0.25      & 39.72          & 18.78            & 56.70           & 54.73    & 55.91\\
\text{Our} &\checkmark   &   &\checkmark   & 0.25            & \textbf{45.77}          & \textbf{18.81}            & \textbf{63.84}           & \textbf{55.27}     &\textbf{68.15}\\
\text{} &   &   &   &            &\increasenoparent{6.05}          &\increasenoparent{0.03}            &\increasenoparent{7.14}           &\increasenoparent{0.54}
&\increasenoparent{12.24}\\
\midrule
\textsc{CYWS} &\checkmark   &\checkmark   &      & 0.25      & 38.40          & 53.22            & 63.93           & 83.83       & 57.59\\
\text{Our} &\checkmark   & \checkmark   &   & 0.25            & \textbf{46.00}          &\textbf{56.06}            &\textbf{68.98}           &\textbf{85.22}      &\textbf{69.45}\\
\text{} &   &   &        &            &\increasenoparent{7.60}          &\increasenoparent{2.84}            &\increasenoparent{5.05}           &\increasenoparent{1.39}
&\increasenoparent{11.86}\\
\midrule
\textsc{CYWS}  &\checkmark   &\checkmark   &\checkmark        & 0.25            & 41.79          & 53.35            & 63.96           & 84.00      & 57.67\\
\text{Our} &\checkmark   &\checkmark   &\checkmark        & 0.25         & \textbf{48.35}          & \textbf{56.99}            & \textbf{69.72}           & \textbf{85.31}     & \textbf{69.53}\\
\text{} &   &   &        &            &\increasenoparent{6.56}          &\increasenoparent{3.64}            &\increasenoparent{5.76}           &\increasenoparent{1.31}
&\increasenoparent{11.86}\\
\bottomrule
\end{tabular}}
\caption{\textbf{Change correspondence F1 Score}. We examine the matching score in two scenarios—one with alignment and the other without—between our model and CYWS model. Our model performs better than CYWS model in two scenarios. For qualitative examples, see (\cref{fig:correspondence our model and CYWS}).}
\label{tab:corr f1}
\end{table}


\subsec{Results} Our model, trained with contrastive matching loss, surpasses CYWS across five datasets under both scenarios—with and without alignment. Without alignment, our model yields improvements of \increasenoparent{6.05} on \cocologo, \increasenoparent{7.14} on \kubriclogo, and \increasenoparent{12.24} on \openlogo. With alignment, it maintains a strong advantage with \increasenoparent{6.56} on \cocologo, \increasenoparent{3.64} on \stdlogo, \increasenoparent{5.76} on \kubriclogo, and \increasenoparent{1.31} on \synthlogo ~(\cref{tab:corr f1}).

\section{Related Work}
\label{sec:literature} 
\noindent \textbf{Change Detection}
The state-of-the-art model CYWS\cite{sachdeva2023change} targets change detection for 2D objects in surveillance images, demonstrating broad applicability without retraining. To extend this capability to 3D objects, CYWS-3D\cite{sachdeva2023change_3d} was proposed. However, neither approach identifies corresponding changes between image pairs.

\noindent Methods like Changemamba\cite{chen2024changemamba}, SCanNet\cite{ding2024joint}, and STADE-CDNet\cite{li2024stade} are specifically designed for remote sensing applications. In this domain, images generally exhibit a single change between two images, simplifying the correspondence problem. In contrast, our approach addresses a more complex correspondence problem, involving multiple changes between two surveillance images (see \cref{fig:more qualitative results betwen cyws and our model}b).

\subsec{Change Segmentation} Prior research, including \cite{alcantarilla2018street, Sakurada2015ChangeDF}, has focused on detecting changes in street views, while studies such as \cite{8608001,gong2017superpixel,wang2021adaptive, xu2021change, yang2021deep} concentrate on satellite imagery. \cite{zheng2024segment} presents a novel zero-shot change segmentation approach specifically for satellite images. Similarly, our model demonstrates strong performance across four zero-shot benchmarks.

\subsec{Change Captioning} 
The Spot-the-Diff (\stdlogo) change captioning dataset, introduced by \cite{jhamtani2018}, contains 13,000 image pairs captured from surveillance cameras. Research in this domain also explores remote sensing image datasets \cite{chang2023changes} and addresses challenges in datasets such as CLEVR-Change, CLEVR-DC, and Bird-to-Words \cite{park2019robust, guo2022clip4idc, kim2021agnostic, shi2020finding, hosseinzadeh2021image, yao2022image, forbes2019neural, huang2021image}, which either simulate or capture real-world changes. These works lack effective change localization, and change captioning becomes more complex when multiple changes occur between two images. Our approach addresses these issues by providing change localization with correspondence, simplifying interpretation. \cite{sun2024stvchrono} presents STVchrono, a benchmark dataset of 71,900 Google Street View images from 18 years across 50 cities to study long-term changes in outdoor scenes. However, its creation is labor-intensive and time-consuming, limiting scalability. In contrast, our \openlogo ~dataset can be efficiently scaled with a simple process.

\section{Discussion and Conclusions}
\noindent \textbf{Limitations} We observe that the accuracy of Point estimation (Stage 2 in our pipeline) plays a critical role in our pipeline. Specifically, images with significant distortions or detailed textures (see \cref{fig:more qualitative results betwen cyws and our model}a) poses a challenge to Point Estimate to align two images and leads to estimation accuracy declines, impacting alignment stage effectiveness. 


\subsec{Conclusions} This study proposes a novel contrastive matching loss function that improves detector accuracy and matching accuracy, surpassing the CYWS method. The post-processing algorithm ensures accurate pairing of changes, and a new metric is introduced for evaluating matching scores across models.

\section*{Acknowledgement}
\noindent We thank Thang Pham, Ali Yildirim, Giang Nguyen, and Tin Nguyen at Auburn University for feedback and discussions of the earlier results.
AN was supported by the NSF Grant No. 1850117 \& 2145767, and donations from NaphCare Foundation \& Adobe Research.

\clearpage
{\small
\bibliographystyle{ieee_fullname}

\begin{thebibliography}{10}\itemsep=-1pt

\bibitem{alcantarilla2018street}
Pablo~F Alcantarilla, Simon Stent, German Ros, Roberto Arroyo, and Riccardo
  Gherardi.
\newblock Street-view change detection with deconvolutional networks.
\newblock {\em Autonomous Robots}, 42:1301--1322, 2018.

\bibitem{carion2020end}
Nicolas Carion, Francisco Massa, Gabriel Synnaeve, Nicolas Usunier, Alexander
  Kirillov, and Sergey Zagoruyko.
\newblock End-to-end object detection with transformers.
\newblock In {\em European conference on computer vision}, pages 213--229.
  Springer, 2020.

\bibitem{chang2023changes}
Shizhen Chang and Pedram Ghamisi.
\newblock Changes to captions: An attentive network for remote sensing change
  captioning.
\newblock {\em arXiv preprint arXiv:2304.01091}, 2023.

\bibitem{chen2024changemamba}
Hongruixuan Chen, Jian Song, Chengxi Han, Junshi Xia, and Naoto Yokoya.
\newblock Changemamba: Remote sensing change detection with spatio-temporal
  state space model.
\newblock {\em arXiv preprint arXiv:2404.03425}, 2024.

\bibitem{ding2024joint}
Lei Ding, Jing Zhang, Haitao Guo, Kai Zhang, Bing Liu, and Lorenzo Bruzzone.
\newblock Joint spatio-temporal modeling for semantic change detection in
  remote sensing images.
\newblock {\em IEEE Transactions on Geoscience and Remote Sensing}, 2024.

\bibitem{dong2021deepatrophy}
Mengjin Dong, Long Xie, Sandhitsu~R Das, Jiancong Wang, Laura~EM Wisse, Robin
  DeFlores, David~A Wolk, Paul~A Yushkevich, Alzheimer's Disease~Neuroimaging
  Initiative, et~al.
\newblock Deepatrophy: Teaching a neural network to detect progressive changes
  in longitudinal mri of the hippocampal region in alzheimer's disease.
\newblock {\em Neuroimage}, 243:118514, 2021.

\bibitem{duan2019centernet}
Kaiwen Duan, Song Bai, Lingxi Xie, Honggang Qi, Qingming Huang, and Qi Tian.
\newblock Centernet: Keypoint triplets for object detection.
\newblock In {\em Proceedings of the IEEE/CVF international conference on
  computer vision}, pages 6569--6578, 2019.

\bibitem{forbes2019neural}
Maxwell Forbes, Christine Kaeser-Chen, Piyush Sharma, and Serge Belongie.
\newblock Neural naturalist: generating fine-grained image comparisons.
\newblock {\em arXiv preprint arXiv:1909.04101}, 2019.

\bibitem{gong2017superpixel}
Maoguo Gong, Tao Zhan, Puzhao Zhang, and Qiguang Miao.
\newblock Superpixel-based difference representation learning for change
  detection in multispectral remote sensing images.
\newblock {\em IEEE Transactions on Geoscience and Remote sensing},
  55(5):2658--2673, 2017.

\bibitem{guo2022clip4idc}
Zixin Guo, Tzu-Jui~Julius Wang, and Jorma Laaksonen.
\newblock Clip4idc: Clip for image difference captioning.
\newblock {\em arXiv preprint arXiv:2206.00629}, 2022.

\bibitem{hosseinzadeh2021image}
Mehrdad Hosseinzadeh and Yang Wang.
\newblock Image change captioning by learning from an auxiliary task.
\newblock In {\em Proceedings of the IEEE/CVF Conference on Computer Vision and
  Pattern Recognition}, pages 2725--2734, 2021.

\bibitem{huang2021image}
Qingbao Huang, Yu Liang, Jielong Wei, Yi Cai, Hanyu Liang, Ho-fung Leung, and
  Qing Li.
\newblock Image difference captioning with instance-level fine-grained feature
  representation.
\newblock {\em IEEE transactions on multimedia}, 24:2004--2017, 2021.

\bibitem{jhamtani2018}
Harsh Jhamtani and Taylor Berg-Kirkpatrick.
\newblock Learning to describe differences between pairs of similar images.
\newblock {\em arXiv preprint arXiv:1808.10584}, 2018.

\bibitem{jiao2024img}
Qirui Jiao, Daoyuan Chen, Yilun Huang, Yaliang Li, and Ying Shen.
\newblock Img-diff: Contrastive data synthesis for multimodal large language
  models.
\newblock {\em arXiv preprint arXiv:2408.04594}, 2024.

\bibitem{kim2021agnostic}
Hoeseong Kim, Jongseok Kim, Hyungseok Lee, Hyunsung Park, and Gunhee Kim.
\newblock Agnostic change captioning with cycle consistency.
\newblock In {\em Proceedings of the IEEE/CVF International Conference on
  Computer Vision}, pages 2095--2104, 2021.

\bibitem{kingma2014adam}
Diederik~P Kingma and Jimmy Ba.
\newblock Adam: A method for stochastic optimization.
\newblock {\em arXiv preprint arXiv:1412.6980}, 2014.

\bibitem{kuhn1955hungarian}
Harold~W Kuhn.
\newblock The hungarian method for the assignment problem.
\newblock {\em Naval research logistics quarterly}, 2(1-2):83--97, 1955.

\bibitem{openimages}
Alina Kuznetsova, Hassan Rom, Neil Alldrin, Jasper R.~R. Uijlings, Ivan Krasin,
  Jordi Pont{-}Tuset, Shahab Kamali, Stefan Popov, Matteo Malloci, Tom Duerig,
  and Vittorio Ferrari.
\newblock The open images dataset {V4:} unified image classification, object
  detection, and visual relationship detection at scale.
\newblock {\em CoRR}, abs/1811.00982, 2018.

\bibitem{li2024stade}
Zhi Li, Siying Cao, Jiakun Deng, Fengyi Wu, Ruilan Wang, Junhai Luo, and
  Zhenming Peng.
\newblock Stade-cdnet: Spatial--temporal attention with difference
  enhancement-based network for remote sensing image change detection.
\newblock {\em IEEE Transactions on Geoscience and Remote Sensing}, 62:1--17,
  2024.

\bibitem{padillaCITE2020}
R. {Padilla}, S.~L. {Netto}, and E.~A.~B. {da Silva}.
\newblock A survey on performance metrics for object-detection algorithms.
\newblock In {\em 2020 International Conference on Systems, Signals and Image
  Processing (IWSSIP)}, pages 237--242, 2020.

\bibitem{park2019robust}
Dong~Huk Park, Trevor Darrell, and Anna Rohrbach.
\newblock Robust change captioning.
\newblock In {\em Proceedings of the IEEE/CVF International Conference on
  Computer Vision}, pages 4624--4633, 2019.

\bibitem{patriarche2004review}
Julia Patriarche and Bradley Erickson.
\newblock A review of the automated detection of change in serial imaging
  studies of the brain.
\newblock {\em Journal of digital imaging}, 17:158--174, 2004.

\bibitem{phan2022deepface}
Hai Phan and Anh Nguyen.
\newblock Deepface-emd: Re-ranking using patch-wise earth mover's distance
  improves out-of-distribution face identification.
\newblock In {\em Proceedings of the IEEE/CVF Conference on Computer Vision and
  Pattern Recognition}, pages 20259--20269, 2022.

\bibitem{roy2018concurrent}
Abhijit~Guha Roy, Nassir Navab, and Christian Wachinger.
\newblock Concurrent spatial and channel ‘squeeze \& excitation’in fully
  convolutional networks.
\newblock In {\em Medical Image Computing and Computer Assisted
  Intervention--MICCAI 2018: 21st International Conference, Granada, Spain,
  September 16-20, 2018, Proceedings, Part I}, pages 421--429. Springer, 2018.

\bibitem{sachdeva2023change}
Ragav Sachdeva and Andrew Zisserman.
\newblock The change you want to see.
\newblock In {\em Proceedings of the IEEE/CVF Winter Conference on Applications
  of Computer Vision}, pages 3993--4002, 2023.

\bibitem{sachdeva2023change_3d}
Ragav Sachdeva and Andrew Zisserman.
\newblock The change you want to see (now in 3d).
\newblock In {\em Proceedings of the IEEE/CVF International Conference on
  Computer Vision}, pages 2060--2069, 2023.

\bibitem{8608001}
Sudipan Saha, Francesca Bovolo, and Lorenzo Bruzzone.
\newblock Unsupervised deep change vector analysis for multiple-change
  detection in vhr images.
\newblock {\em IEEE Transactions on Geoscience and Remote Sensing},
  57(6):3677--3693, 2019.

\bibitem{Sakurada2015ChangeDF}
Ken Sakurada and Takayuki Okatani.
\newblock Change detection from a street image pair using cnn features and
  superpixel segmentation.
\newblock In {\em British Machine Vision Conference}, 2015.

\bibitem{sarlin2020superglue}
Paul-Edouard Sarlin, Daniel DeTone, Tomasz Malisiewicz, and Andrew Rabinovich.
\newblock Superglue: Learning feature matching with graph neural networks.
\newblock In {\em Proceedings of the IEEE/CVF conference on computer vision and
  pattern recognition}, pages 4938--4947, 2020.

\bibitem{shi2020finding}
Xiangxi Shi, Xu Yang, Jiuxiang Gu, Shafiq Joty, and Jianfei Cai.
\newblock Finding it at another side: A viewpoint-adapted matching encoder for
  change captioning.
\newblock In {\em Computer Vision--ECCV 2020: 16th European Conference,
  Glasgow, UK, August 23--28, 2020, Proceedings, Part XIV 16}, pages 574--590.
  Springer, 2020.

\bibitem{stent2015detecting}
Simon Stent, Riccardo Gherardi, Bj{\"o}rn Stenger, and Roberto Cipolla.
\newblock Detecting change for multi-view, long-term surface inspection.
\newblock In {\em BMVC}, pages 127--1, 2015.

\bibitem{sun2024stvchrono}
Yanjun Sun, Yue Qiu, Mariia Khan, Fumiya Matsuzawa, and Kenji Iwata.
\newblock The stvchrono dataset: Towards continuous change recognition in time.
\newblock In {\em Proceedings of the IEEE/CVF Conference on Computer Vision and
  Pattern Recognition}, pages 14111--14120, 2024.

\bibitem{suvorov2021resolution}
Roman Suvorov, Elizaveta Logacheva, Anton Mashikhin, Anastasia Remizova,
  Arsenii Ashukha, Aleksei Silvestrov, Naejin Kong, Harshith Goka, Kiwoong
  Park, and Victor Lempitsky.
\newblock Resolution-robust large mask inpainting with fourier convolutions.
\newblock In {\em Proceedings of the IEEE conference on computer vision and
  pattern recognition}, 2021.

\bibitem{tan2019image-editing-request}
Hao Tan, Franck Dernoncourt, Zhe Lin, Trung Bui, and Mohit Bansal.
\newblock Expressing visual relationships via language.
\newblock In Anna Korhonen, David Traum, and Llu{\'\i}s M{\`a}rquez, editors,
  {\em Proceedings of the 57th Annual Meeting of the Association for
  Computational Linguistics}, pages 1873--1883, Florence, Italy, July 2019.
  Association for Computational Linguistics.

\bibitem{wang2021adaptive}
Congcong Wang, Wenbin Sun, Deqin Fan, Xiaoding Liu, and Zhi Zhang.
\newblock Adaptive feature weighted fusion nested u-net with discrete wavelet
  transform for change detection of high-resolution remote sensing images.
\newblock {\em Remote Sensing}, 13(24):4971, 2021.

\bibitem{wu2018spot}
Junhui Wu, Yun Ye, Yu Chen, and Zhi Weng.
\newblock Spot the difference by object detection.
\newblock {\em arXiv preprint arXiv:1801.01051}, 2018.

\bibitem{xu2021change}
Quanfu Xu, Keming Chen, Guangyao Zhou, and Xian Sun.
\newblock Change capsule network for optical remote sensing image change
  detection.
\newblock {\em Remote Sensing}, 13(14):2646, 2021.

\bibitem{yang2021deep}
Le Yang, Yiming Chen, Shiji Song, Fan Li, and Gao Huang.
\newblock Deep siamese networks based change detection with remote sensing
  images.
\newblock {\em Remote Sensing}, 13(17):3394, 2021.

\bibitem{yao2022image}
Linli Yao, Weiying Wang, and Qin Jin.
\newblock Image difference captioning with pre-training and contrastive
  learning.
\newblock In {\em Proceedings of the AAAI Conference on Artificial
  Intelligence}, volume~36, pages 3108--3116, 2022.

\bibitem{zheng2024segment}
Zhuo Zheng, Yanfei Zhong, Liangpei Zhang, and Stefano Ermon.
\newblock Segment any change.
\newblock {\em arXiv preprint arXiv:2402.01188}, 2024.

\end{thebibliography}

}

\newcommand{\beginsupplementary}{%
    \setcounter{table}{0}
    \renewcommand{\thetable}{A\arabic{table}}%
    
    \setcounter{figure}{0}
    \renewcommand{\thefigure}{A\arabic{figure}}%
    
    \setcounter{section}{0}
    \renewcommand{\thesection}{A\arabic{section}}
    \renewcommand{\thesubsection}{\thesection.\arabic{subsection}}
}

\beginsupplementary%
\appendix

\newcommand{\toptitlebar}{
    \hrule height 4pt
    \vskip 0.25in
    \vskip -\parskip%
}
\newcommand{\bottomtitlebar}{
    \vskip 0.29in
    \vskip -\parskip%
    \hrule height 1pt
    \vskip 0.09in%
}

\newcommand{\suptitle}{Appendix for: Improving Zero-Shot Object-Level Change Detection \\by Incorporating Visual Correspondence}

\newcommand{\maketitlesupp}{
    \newpage
    \onecolumn
        \null
        \vskip .375in
        \begin{center}
            \toptitlebar
            {\Large \bf \suptitle\par}
            \bottomtitlebar
            \vspace*{24pt}
            {
                \large
                \lineskip=.5em
                \par
            }
            \vskip .5em
            \vspace*{12pt}
        \end{center}
}
\maketitlesupp%

\section{Upper bound accuracy of correspondence algorithm}
\label{sec:Upper bound accuracy of correspondence algorithm app}
\noindent Here, we want to estimate the correspondence component. Correspondence algorithm consists of alignment step before using the Hungarian algorithm. By using ground-truth boxes, we can evaluate the maximum accuracy of the matching algorithm. 

\subsec{Experiments} To assess the effectiveness of the post-processing method we employ ground-truth boxes directly rather than utilising the change detector's projected box output as the feature extractor's input.

\subsec{Results} The findings presented in \cref{tab:corr upper bound}  upper bound indicate that our matching method demonstrates strong performance in $F_{1}$ score when applied to both the \stdlogo(\increasenoparent{100}) and \synthlogo(\increasenoparent{99.96}) algorithms. A gap persists in the availability of the \kubriclogo(96.50) and \cocologo(91.68) datasets. The efficacy of the transformation matrix is limited in certain challenging scenarios involving \kubriclogo ~or \cocologo. The \cocologo ~dataset contains numerous artifacts, which hinder the accurate estimation of the transformation matrix. 
\begin{table}[H]
\centering
\begin{tabular}{lccccc}
\toprule
\multicolumn{6}{c}{\textbf{Change (F1 Score)}}\\
\midrule
Model & \cocologo & \stdlogo & \kubriclogo & \synthlogo & \openlogo \\
\midrule
\text{Ground-truth} Baseline   & 91.68          & \textbf{100}           & 96.50           & 99.96       &99.94\\
\bottomrule
\end{tabular}
\caption{\textbf{Correspondence Accuracy Upper Bound.} Using ground truth boxes as input for matching algorithm}
\label{tab:corr upper bound}
\end{table}

\section{Features of mean pooling provide more accurate correspondence than cropped images features}
\label{sec:crop vs avg}
\noindent The proposed approach offers flexibility in selecting methods for assigning embeddings to predicted boxes. This section evaluates two methodologies for generating embeddings. To identify the optimal method, we conduct a comparative analysis using our fine-tuned model. The effectiveness of each approach is assessed based on the matching score (F1). 

\subsec{Experiments} This section analyzes the impact of two embedding assignment methods: mean-pooling and region cropping on the correspondence score. The analysis is conducted based on the methodologies outlined in (\cref{sec:corr}).

\subsec{Results} We hypothesize that using only cropped images reduces the availability of contextual information surrounding the object, resulting in lower correspondence accuracy. The average feature method consistently outperforms the cropping method across all five datasets, with significant improvements observed in the \synthlogo~ and \openlogo~ datasets. Consequently, we have adopted the average feature technique for all subsequent experiments. Detailed results are presented in \cref{tab:average feature vs crop feature}.

\begin{table}[h]
\centering
\begin{tabular}{lcccccccc}
\toprule
Model &Average &Crop &Thres & \cocologo & \stdlogo & \kubriclogo & \synthlogo & \openlogo \\
\midrule
\text{Our + ResNet-50}    &       &\checkmark    & 0.25            & 44.10          & 56.29            & 68.10           & 67.73         & 62.25 \\
\text{Our + ResNet-50}    &\checkmark     &    & 0.25            & 46.19          & 56.94            & 69.52           & 85.33         & 69.53 \\
\text{}    &     &    &          & \increasenoparent{2.09}          & \increasenoparent{0.65}           & \increasenoparent{1.42}           & \increasenoparent{17.60}       & \increasenoparent{7.28} \\
\bottomrule
\end{tabular}
\caption{Features obtained using the \textbf{average} method achieve higher F1 scores compared to those derived from the cropping method. This approach consistently produces reliable results across all datasets, with particularly notable performance on the \synthlogo~ and \openlogo~ datasets.}
\label{tab:average feature vs crop feature}
\end{table}

\section{Training hyperparameters}
\label{sec:training_hyper_app}
\subsec{Results} We follow the training hyperparameters in (\cref{sec:training params}). We investigate the impact of training parameters, including the number of epochs and learning rate, on model performance. Training for 500 epochs led to overfitting, reducing zero-shot accuracy on the \stdlogo, \kubriclogo, \synthlogo, and \openlogo ~datasets (\cref{tab:200 vs 500 epochs}). Increasing the learning rate from 0.0001 to 0.0005 further degraded accuracy (\cref{tab:influence of learning rate}). Additionally, using a deeper decoder did not improve accuracy (\cref{tab:3 vs 5 decoder layers app}).
\begin{table}[H]
\centering
\begin{tabular}{lccccccccc}
\toprule
\multicolumn{10}{c}{\textbf{Change (mAP Score)}}\\
\midrule
Model & LR & CenterNet & DETR &Contrastive & \cocologo & \stdlogo & \kubriclogo & \synthlogo & \openlogo \\
\midrule
\text{Our} &0.0005   &\checkmark   &\checkmark  &\checkmark  &57.87          &47.95           &68.20           &88.54        &60.33\\
\text{Our} &0.0001   &\checkmark   &\checkmark  &\checkmark  &\textbf{71.77}          &\textbf{57.00}           &\textbf{81.07}           &\textbf{90.02}       &\textbf{78.84}\\
\bottomrule
\end{tabular}
\caption{\textbf{Training with different learning rate (LR).} Using different learning rate in training}
\label{tab:influence of learning rate}
\end{table}

\begin{table}[H]
\centering
\begin{tabular}{lccccccccc}
\toprule
\multicolumn{10}{c}{\textbf{Change (mAP Score)}}\\
\midrule
Model & Epochs & CenterNet & DETR &Contrastive & \cocologo & \stdlogo & \kubriclogo & \synthlogo & \openlogo\\
\midrule
\text{Our} &500   &\checkmark   &\checkmark  &\checkmark  &\textbf{72.15}          &54.17           &78.64           &89.01            & 78.65\\
\text{Our} &200   &\checkmark   &\checkmark  &\checkmark  &71.77          &\textbf{57.00}           &\textbf{81.07}           &\textbf{90.02}         &\textbf{78.84}\\
\bottomrule
\end{tabular}
\caption{\textbf{Training with more epochs.} Training the model for 500 epochs decreases accuracy in zero-shot testing on the \stdlogo, \kubriclogo, \synthlogo, and \openlogo ~datasets.}
\label{tab:200 vs 500 epochs}
\end{table}

\clearpage
\section{Training with a deeper decoder does not enhance model accuracy}
\label{sec:deeper decoder}
\noindent In order to find the best change detection architecture, we added more layers to the decoder in this section. 

\subsec{Experiment} We used [256, 128, 64] channels for each decoder layer in the prior configuration. We add two further layers with 32 and 8 channels, respectively, in this configuration.  

\subsec{Results} The outcomes of employing deeper decoder layers are displayed in \cref{tab:3 vs 5 decoder layers app}. The findings demonstrate that the final accuracy decreases with the number of decoder layers. 

\begin{table}[H]
\centering
\begin{tabular}{lccccccccc}
\toprule
\multicolumn{10}{c}{\textbf{Change (mAP Score)}}\\
\midrule
Model & Epochs & CenterNet & DETR &Contrastive & \cocologo & \stdlogo & \kubriclogo & \synthlogo & \openlogo \\
\midrule
\text{Our} &200   &\checkmark   &\checkmark  &\checkmark  &51.76          &49.70           &63.95           &87.57            & 54.39\\
\text{Our} &200   &\checkmark   &\checkmark  &\checkmark  &\textbf{71.77}          &\textbf{57.00}           &\textbf{81.07}           &\textbf{90.02}          &\textbf{78.84}\\
\bottomrule
\end{tabular}
\caption{\textbf{Training with a deeper decoder} does not enhance model accuracy}
\label{tab:3 vs 5 decoder layers app}
\end{table}

\section{The alignment stage plays a crucial role in the success of the matching algorithm} 
\label{sec:alignment app}
\begin{figure*}
     \centering
     \resizebox{\textwidth}{!}{
     \fontsize{6pt}{6pt}\selectfont
     \begin{tabular}{c@{}c@{}c@{}c@{}c@{}c}
     &(a) \cocologo &(b) \stdlogo &(c) \kubriclogo &(d) \synthlogo &(e) \openlogo \\ 
     \vspace{-.15cm}
     \rotatebox{90}{\hspace{0.4cm}w/o Align} 
     &\includegraphics[width=0.23\linewidth]{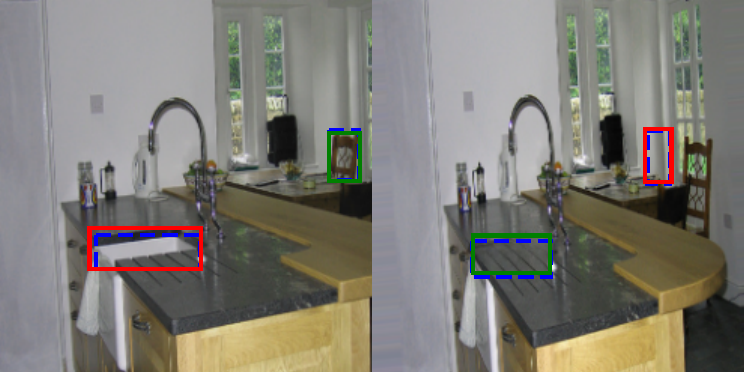} &\includegraphics[width=0.23\linewidth]{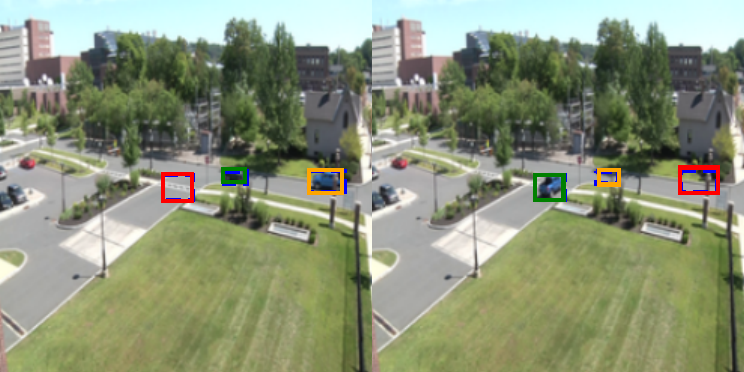}
     &\includegraphics[width=0.23\linewidth]{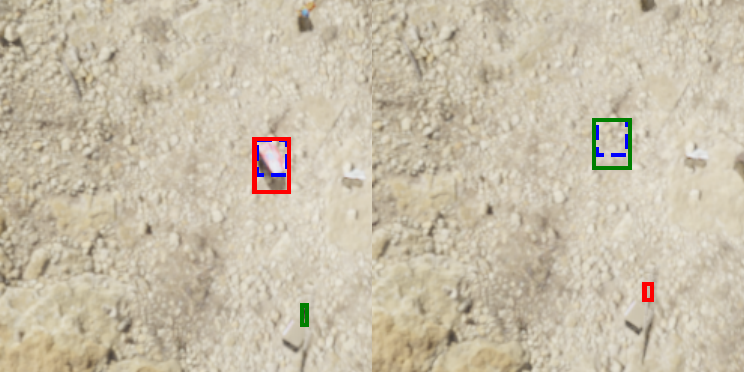}
     &\includegraphics[width=0.23\linewidth]{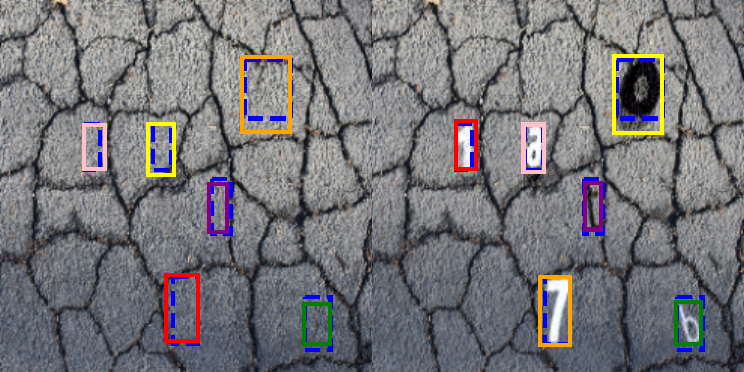}
     &\includegraphics[width=0.23\linewidth]{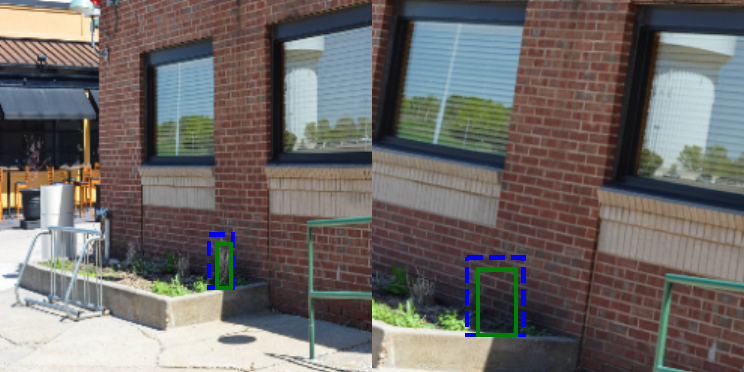}\\
     \rotatebox{90}{\hspace{0.35cm}w/ Align} 
     &\includegraphics[width=0.23\linewidth]{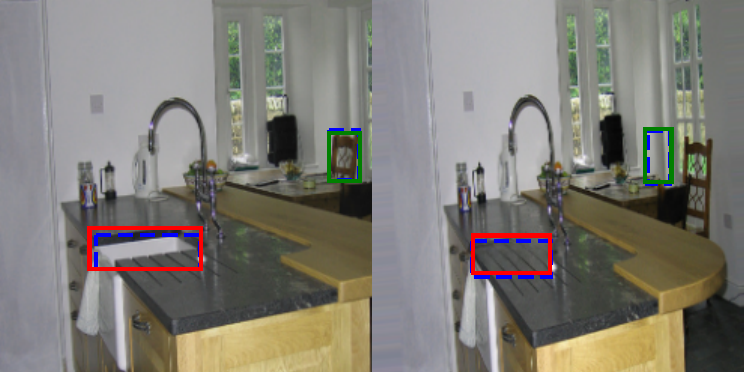} &\includegraphics[width=0.23\linewidth]{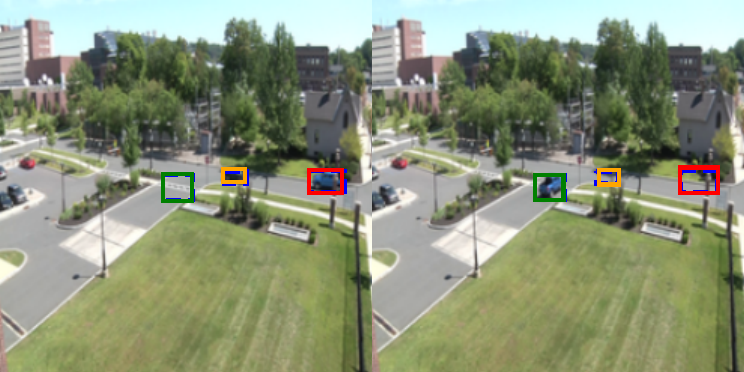} 
     &\includegraphics[width=0.23\linewidth]{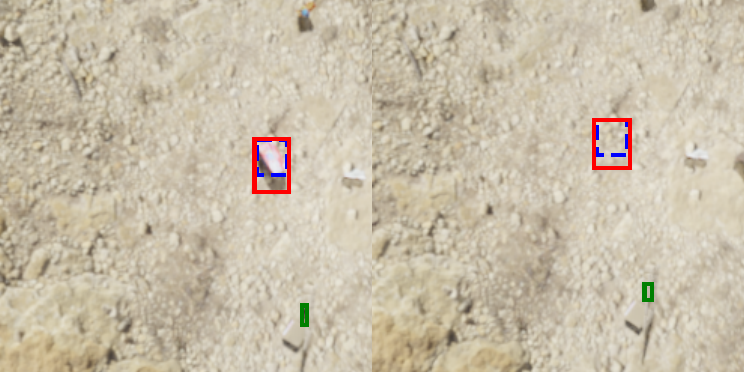}
     &\includegraphics[width=0.23\linewidth]{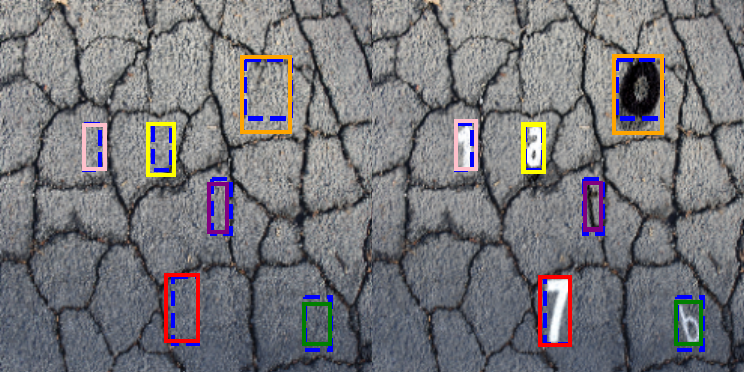}
     &\includegraphics[width=0.23\linewidth]{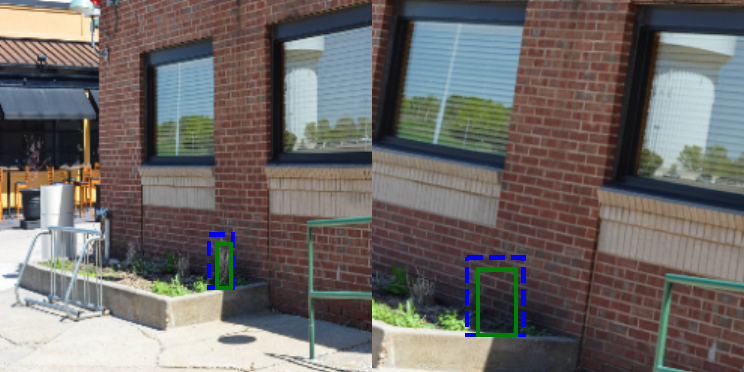}\\
     \end{tabular}}
     \caption{With the significant improvement in the \stdlogo and \synthlogo datasets, the alignment stage is a crucial component in increasing correspondence accuracy. The second row's findings demonstrate how the alignment step aids in correcting every case's incorrect matching in the first row. You may view the improvement's specifics in the \cref{tab:add align results}.
     }
     \label{fig:w/ and w/o alignment}
\end{figure*}

\noindent The qualitative results \cref{fig:w/ and w/o alignment} we present in this section demonstrate how well our alignment stage worked to enhance the matched pairs of modifications displayed in the \cref{tab:add align results}.

\section{Correspondence} 
\label{sec:Corr app}
\begin{figure*}[t]
     \centering
     \resizebox{\textwidth}{!}{
     \fontsize{6pt}{6pt}\selectfont
     \begin{tabular}{c@{}c@{}c@{}c@{}c@{}c}
     &(a) \cocologo &(b) \stdlogo &(c) \kubriclogo &(d) \synthlogo &(e) \openlogo \\ 
     \vspace{-.07cm}
     \rotatebox{90}{\hspace{0.45cm} CYWS} 
     &\includegraphics[width=0.23\linewidth]{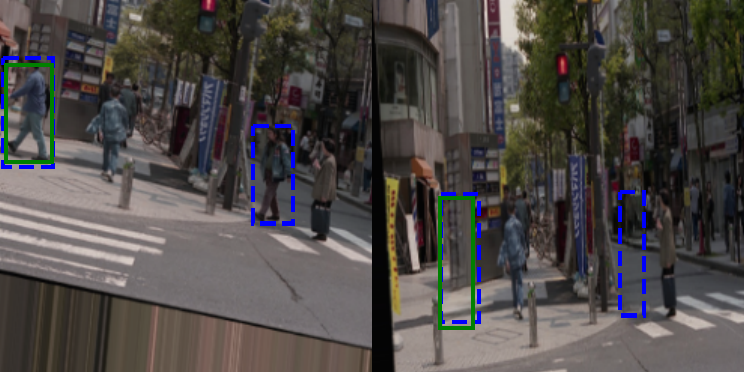} &\includegraphics[width=0.23\linewidth]{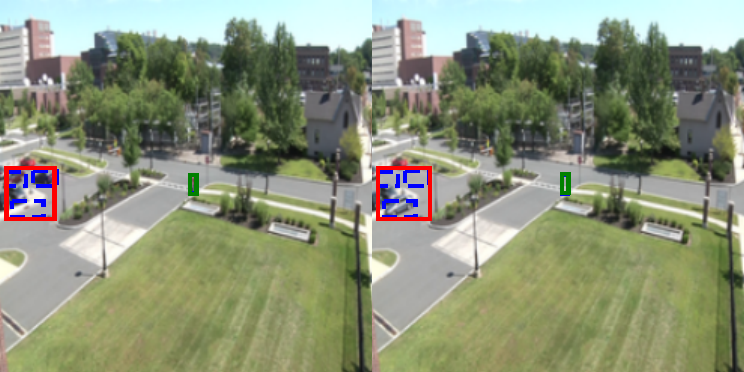}
     &\includegraphics[width=0.23\linewidth]{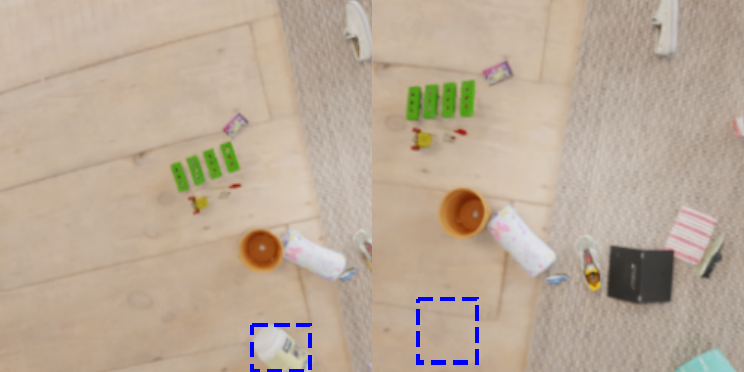}
     &\includegraphics[width=0.23\linewidth]{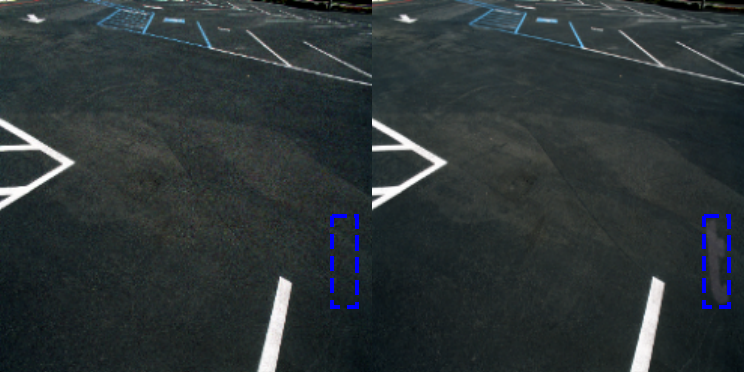}
     &\includegraphics[width=0.23\linewidth]{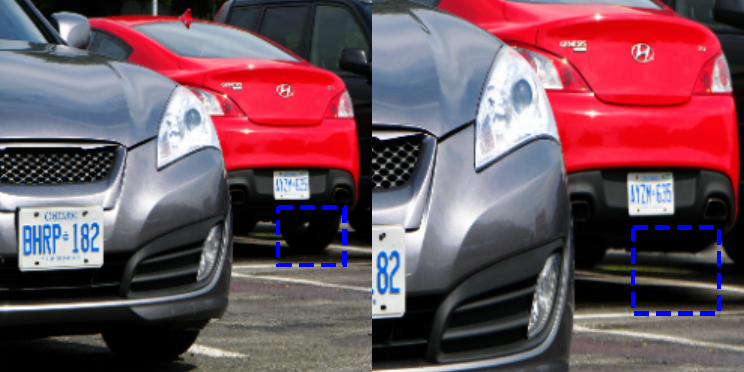}\\
     \rotatebox{90}{\hspace{0.5cm} Our} 
     &\includegraphics[width=0.23\linewidth]{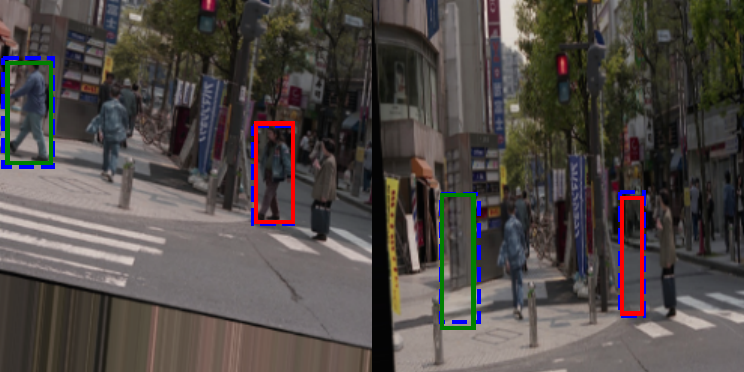}
     &\includegraphics[width=0.23\linewidth]{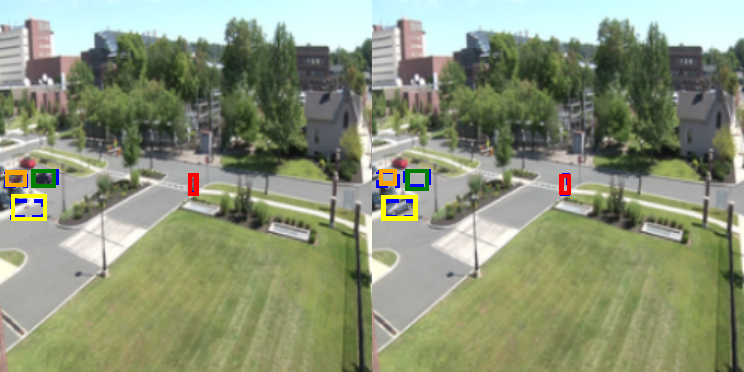}
     &\includegraphics[width=0.23\linewidth]{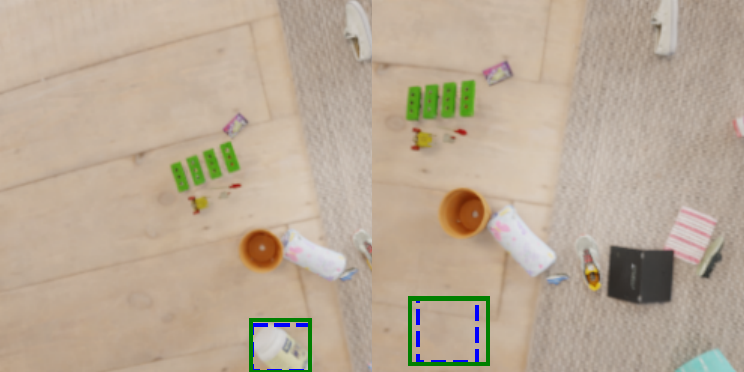}
     &\includegraphics[width=0.23\linewidth]{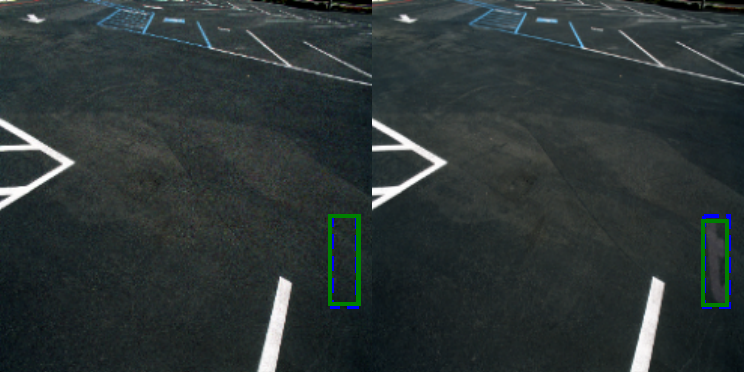}
     &\includegraphics[width=0.23\linewidth]{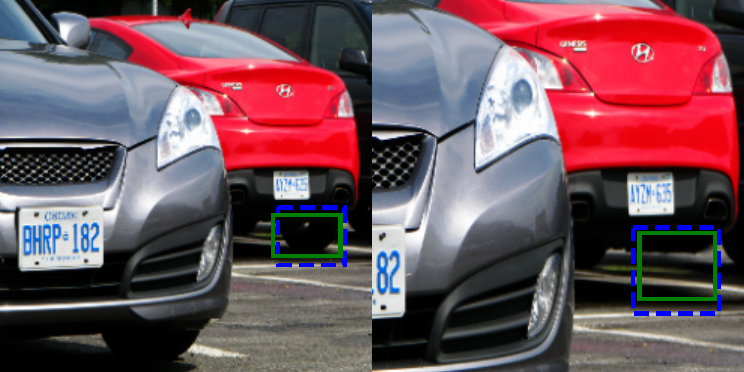}\\
     \end{tabular}}
     \caption{CYWS model, as seen in (a), (b), (c), (d) and (e), is unable to identify every difference between two images. 
     Conversely, our model is able to identify every change in the two images. CYWS model can only identify one change for the entire region in the \stdlogo ~example, where three changes appear at nearly the same location. Our model, on the other hand, can identify each of the three changes independently. We hypothesise that the model learns the number of changes implicitly based on information gleaned from the contrastive matching loss. Check \cref{tab:corr f1} for quantitative results. 
     }
     \label{fig:correspondence our model and CYWS}
\end{figure*}

We present qualitative results in this part that contrast our model with CYWS model in terms of matching qualitative. According to the qualitative results, our model outperforms CYWS model in the matching score, as indicated by the \cref{tab:corr f1}. See qualitative results in \cref{fig:correspondence our model and CYWS}

\section{Reduce false positive predicted box in no-change case}
\begin{figure*}
     \centering
     \fontsize{6pt}{6pt}\selectfont
     \begin{tabular}{c@{}c@{}c@{}c@{}c}
     &(a) \cocologo ~No-Change &(b) \stdlogo ~No-Change &(c) \kubriclogo ~No-Change &(d) \synthlogo ~No-Change \\ 
     \vspace{-.07cm}
     \rotatebox{90}{\hspace{0.45cm} CYWS} 
     &\includegraphics[width=0.23\linewidth]{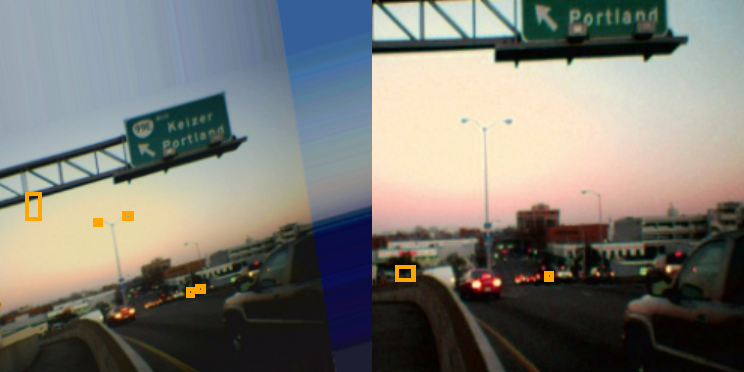} &\includegraphics[width=0.23\linewidth]{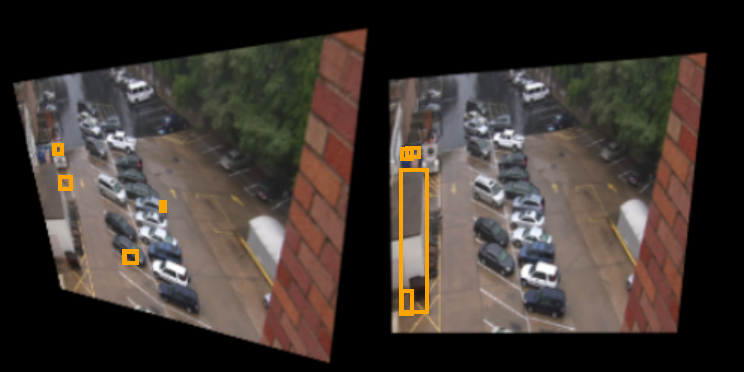}
     &\includegraphics[width=0.23\linewidth]{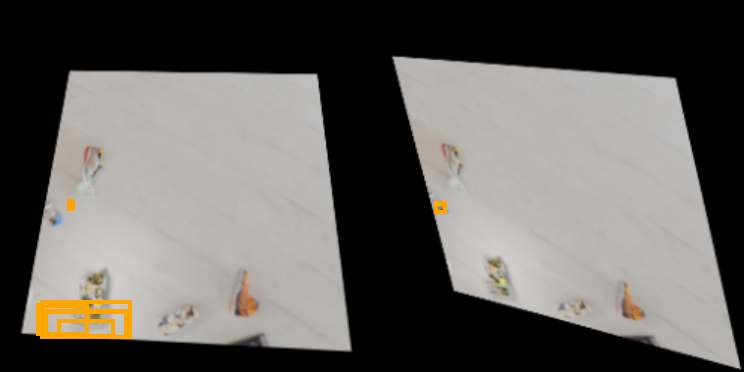}
     &\includegraphics[width=0.23\linewidth]{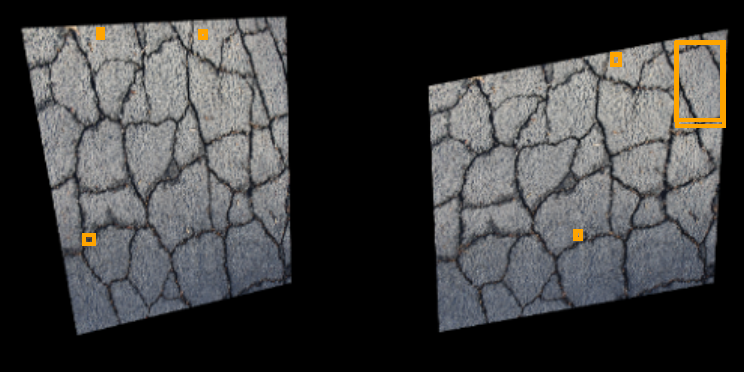} \\
     \rotatebox{90}{\hspace{0.2cm} CYWS + PP} 
     &\includegraphics[width=0.23\linewidth]{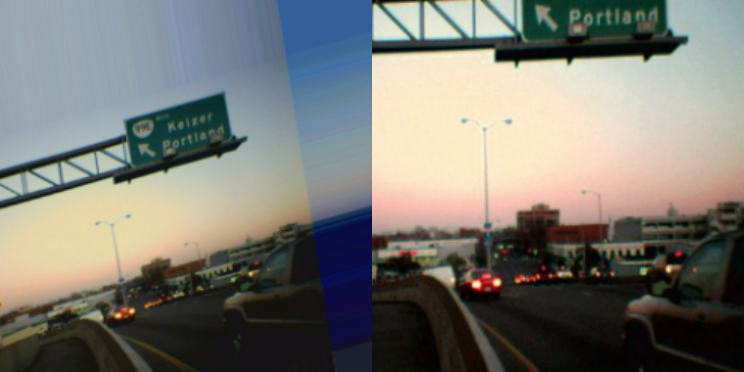}
     &\includegraphics[width=0.23\linewidth]{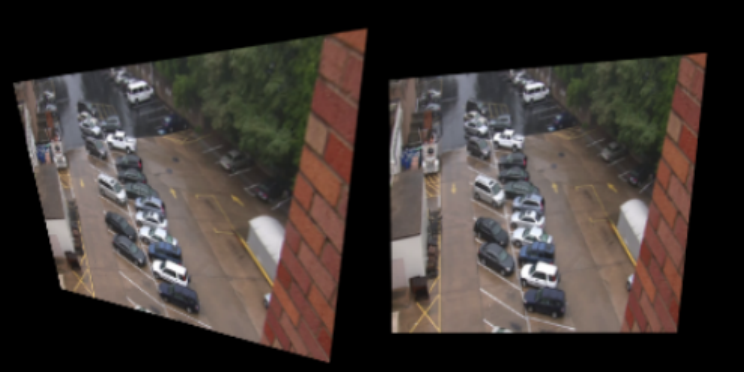}
     &\includegraphics[width=0.23\linewidth]{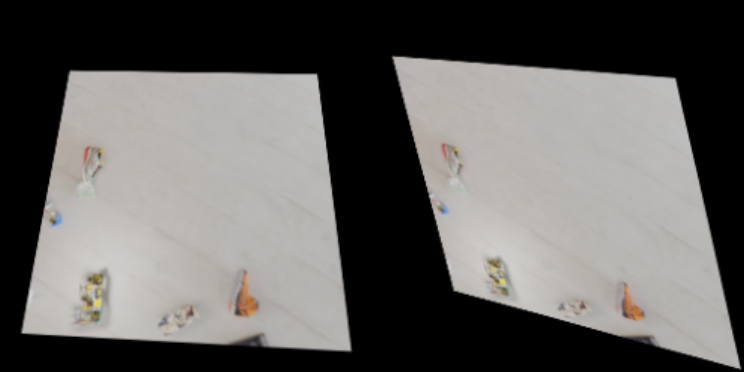}
     &\includegraphics[width=0.23\linewidth]{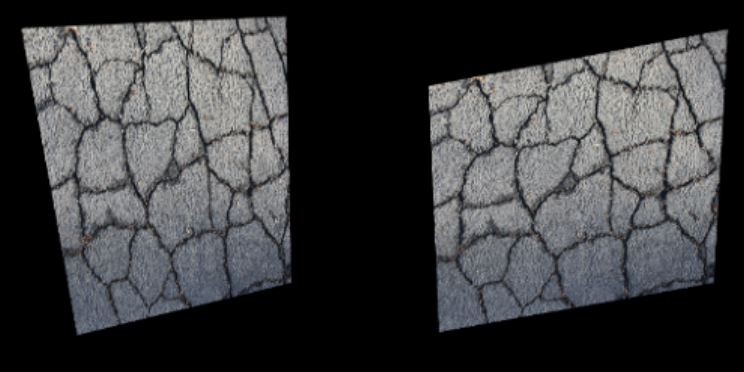}\\
     \end{tabular}
     \caption{In no-change scenarios, our post-processing approach reduces false positive predicted boxes.
     }
     \label{fig:false positive predicted boxes reduction}
\end{figure*}

The output from CYWS model in the default situations is shown in the first row of \cref{fig:false positive predicted boxes reduction}. The outcomes of our post-processing procedure are shown in the row that follows.

\section{Additional qualitative results}
\begin{figure*}[t]
     \centering
     \resizebox{\textwidth}{!}{
     \fontsize{6pt}{6pt}\selectfont
     \begin{tabular}{c@{}c@{}c@{}c@{}c@{}c}
     &(a) \cocologo &(b) \stdlogo &(c) \kubriclogo &(d) \synthlogo &(e) \openlogo \\ 
     \vspace{-.07cm}
     \rotatebox{90}{\hspace{0.4cm} CYWS} 
     &\includegraphics[width=0.23\linewidth]{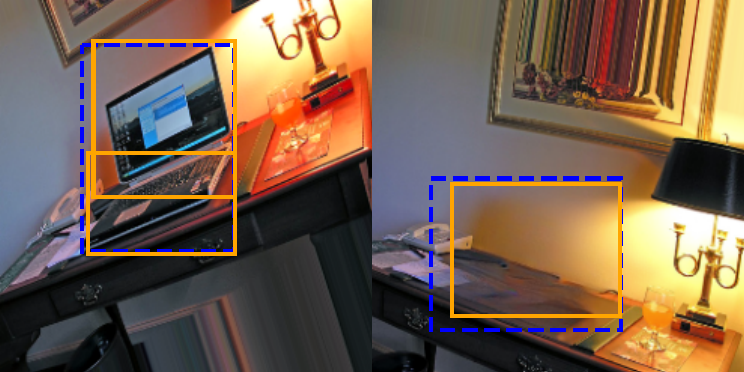} &\includegraphics[width=0.23\linewidth]{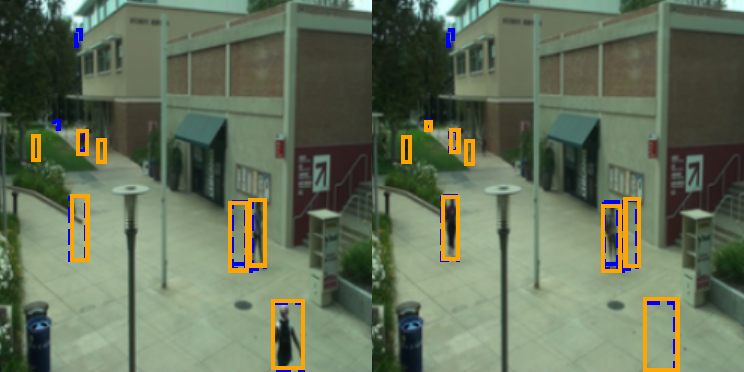}
     &\includegraphics[width=0.23\linewidth]{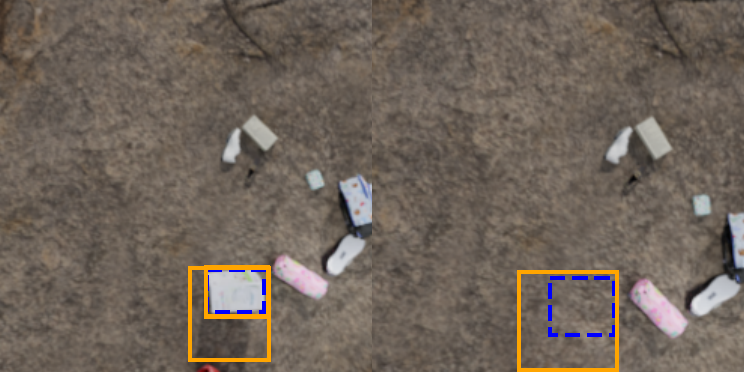}
     &\includegraphics[width=0.23\linewidth]{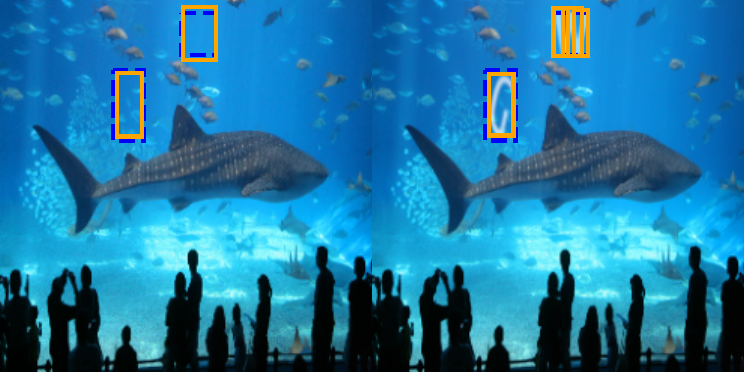}
     &\includegraphics[width=0.23\linewidth]{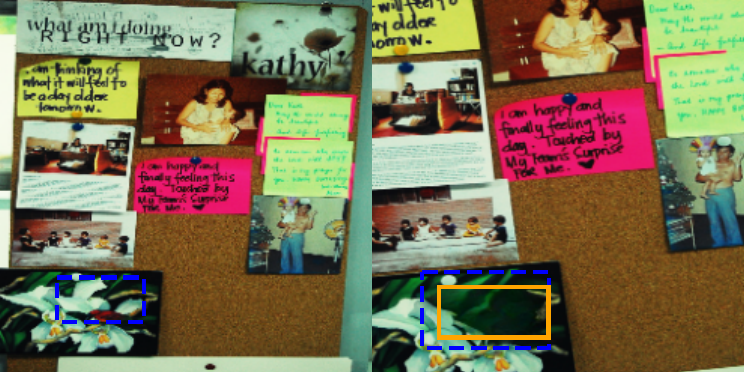}\\
     \vspace{-.059cm}
     \rotatebox{90}{\hspace{0.3cm} Our + PP} 
     &\includegraphics[width=0.23\linewidth]{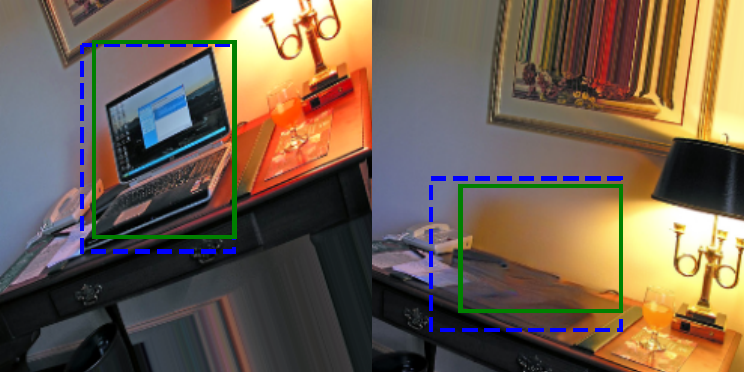}
     &\includegraphics[width=0.23\linewidth]{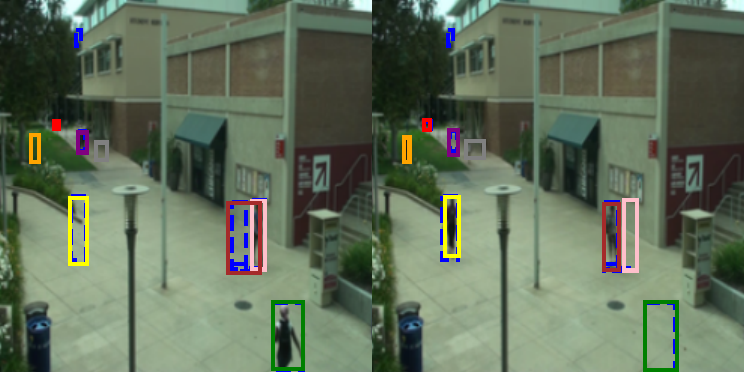}
     &\includegraphics[width=0.23\linewidth]{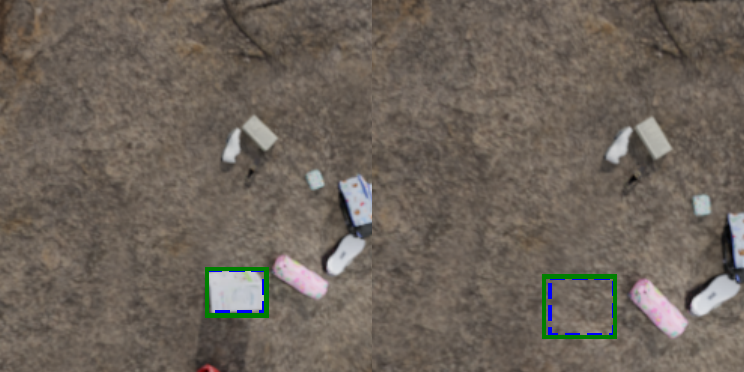}
     &\includegraphics[width=0.23\linewidth]{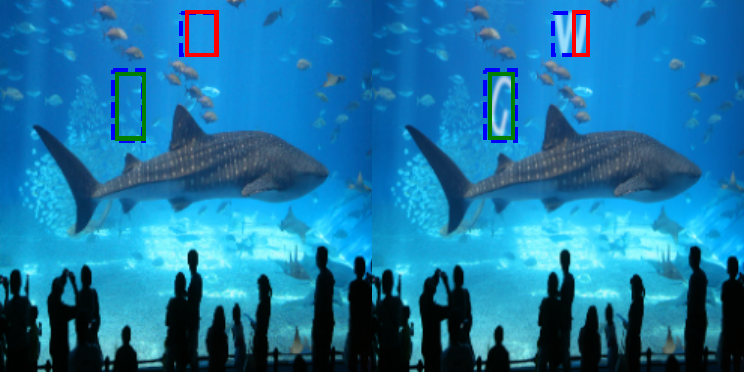}
     &\includegraphics[width=0.23\linewidth]{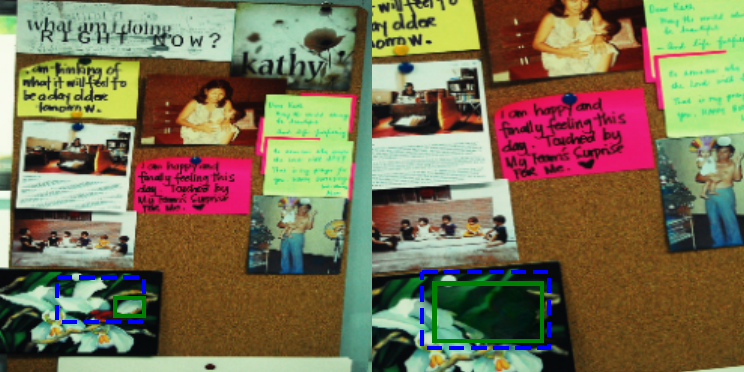}\\

     \vspace{-.07cm}
     \rotatebox{90}{\hspace{0.4cm} CYWS} 
     &\includegraphics[width=0.23\linewidth]{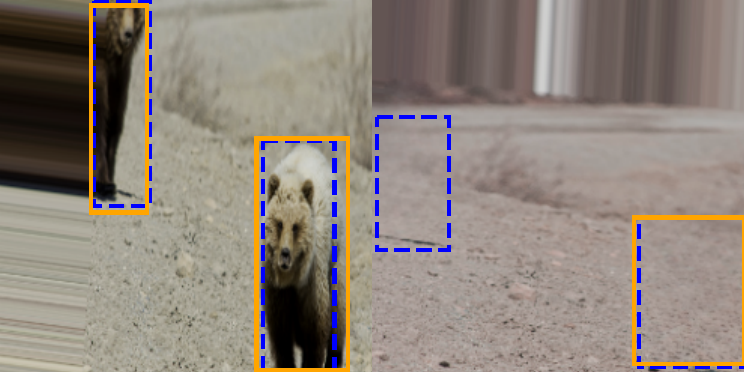} &\includegraphics[width=0.23\linewidth]{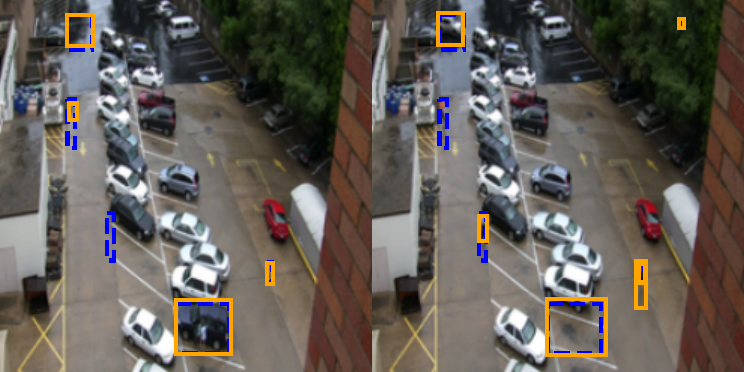}
     &\includegraphics[width=0.23\linewidth]{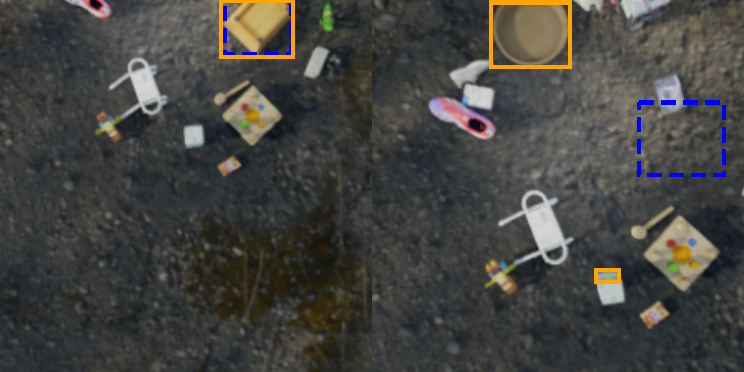}
     &\includegraphics[width=0.23\linewidth]{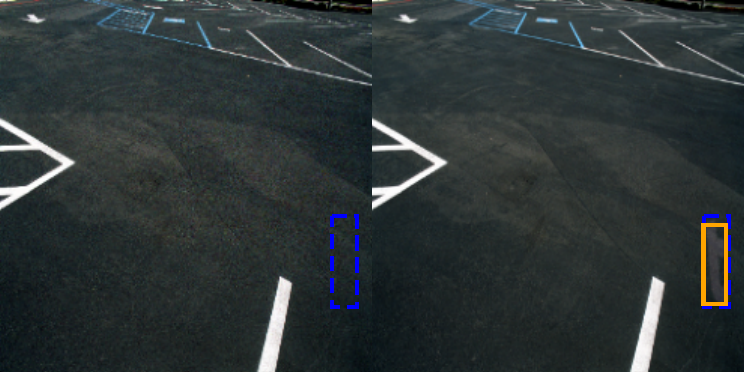} 
     &\includegraphics[width=0.23\linewidth]{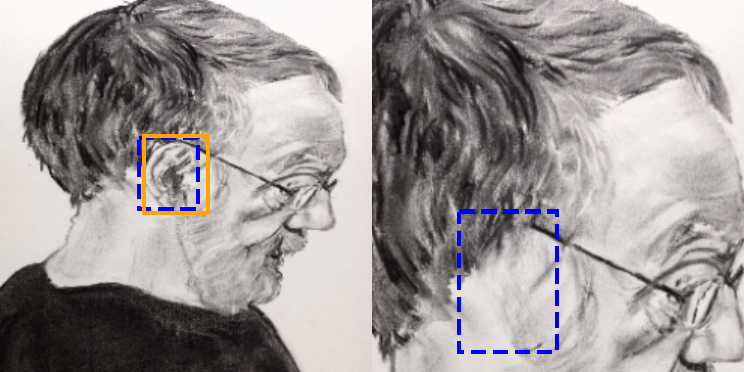} \\
     \vspace{-.059cm}
     \rotatebox{90}{\hspace{0.3cm} Our + PP} 
     &\includegraphics[width=0.23\linewidth]{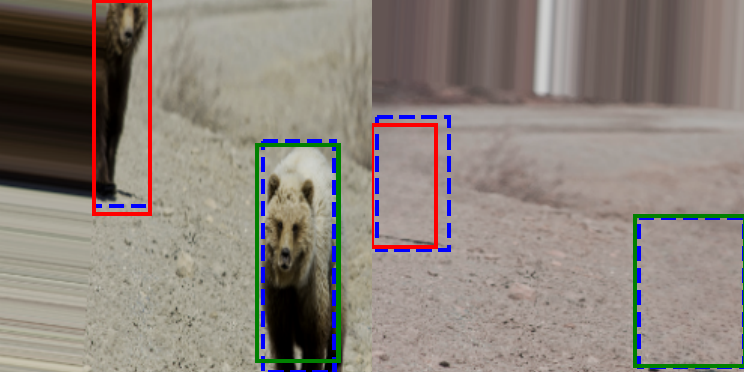}
     &\includegraphics[width=0.23\linewidth]{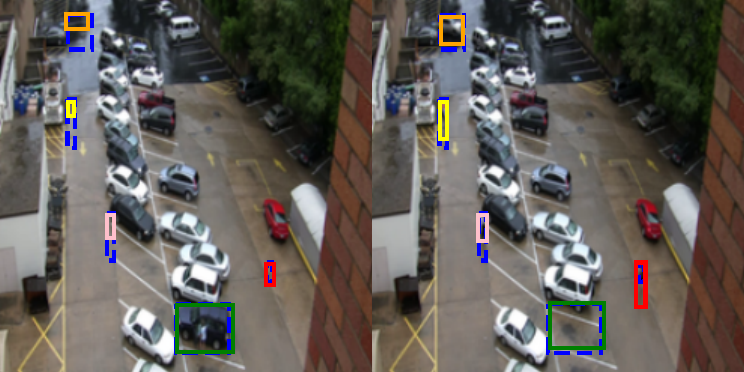}
     &\includegraphics[width=0.23\linewidth]{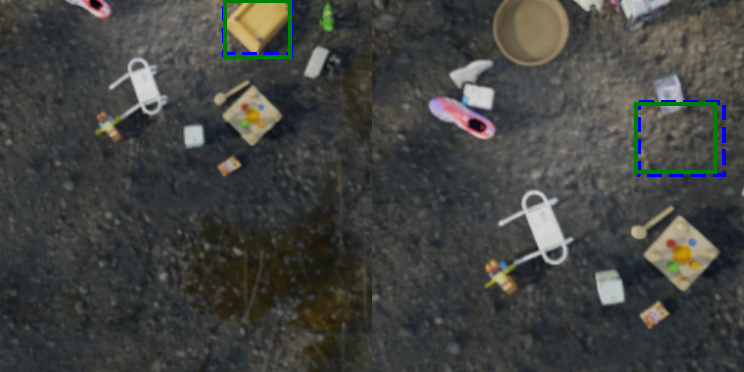}
     &\includegraphics[width=0.23\linewidth]{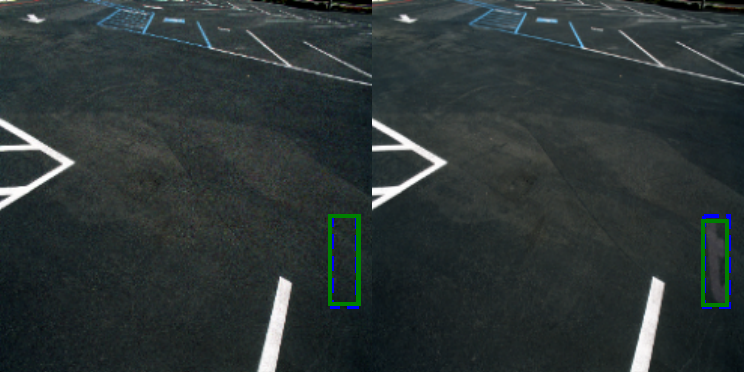}
     &\includegraphics[width=0.23\linewidth]{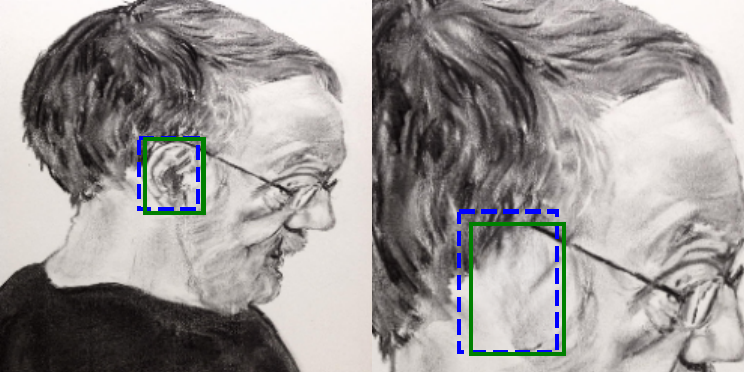}\\

     \vspace{-.07cm}
     \rotatebox{90}{\hspace{0.4cm} CYWS} 
     &\includegraphics[width=0.23\linewidth]{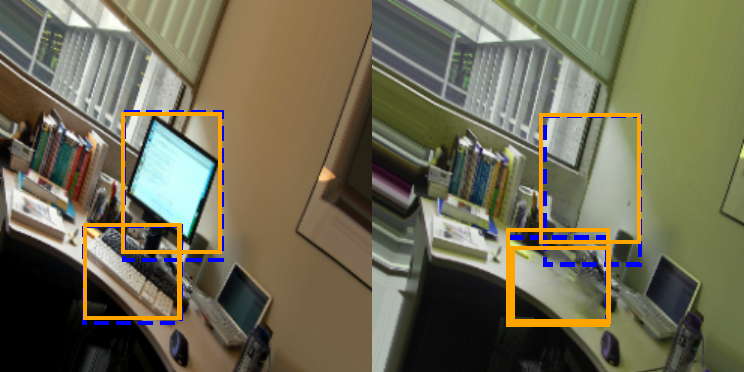} &\includegraphics[width=0.23\linewidth]{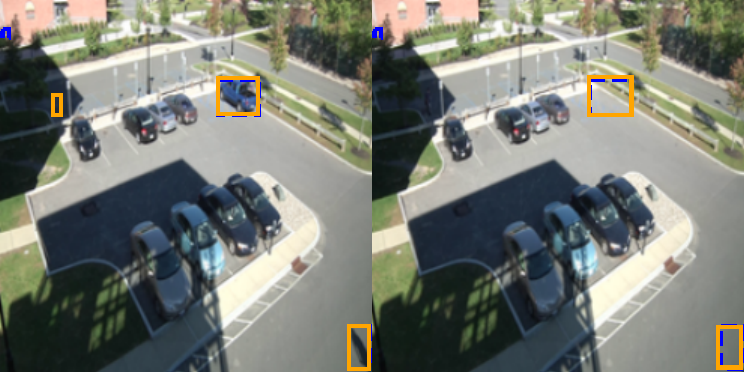}
     &\includegraphics[width=0.23\linewidth]{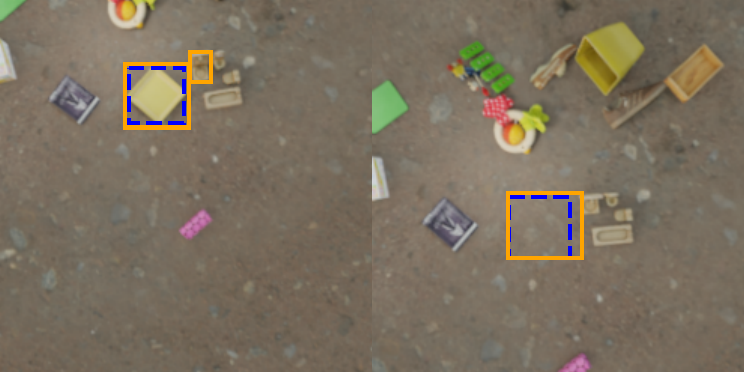}
     &\includegraphics[width=0.23\linewidth]{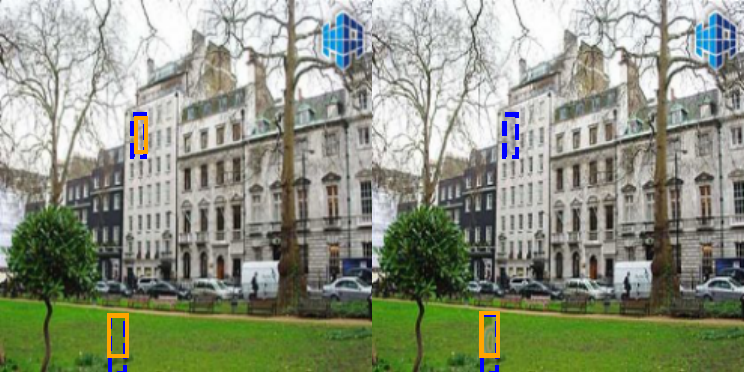} 
     &\includegraphics[width=0.23\linewidth]{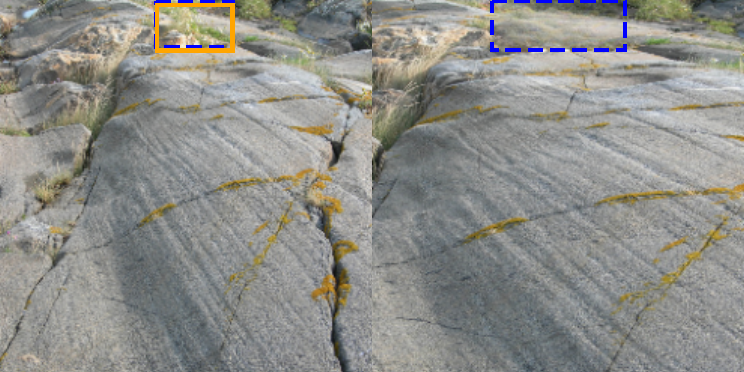}\\
     \vspace{-.059cm}
     \rotatebox{90}{\hspace{0.3cm} Our + PP} 
     &\includegraphics[width=0.23\linewidth]{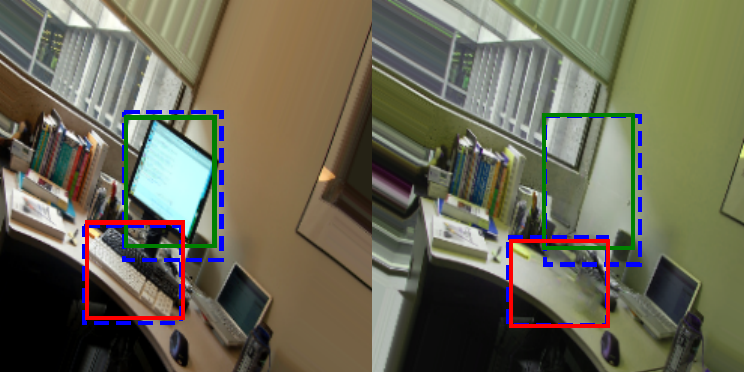}
     &\includegraphics[width=0.23\linewidth]{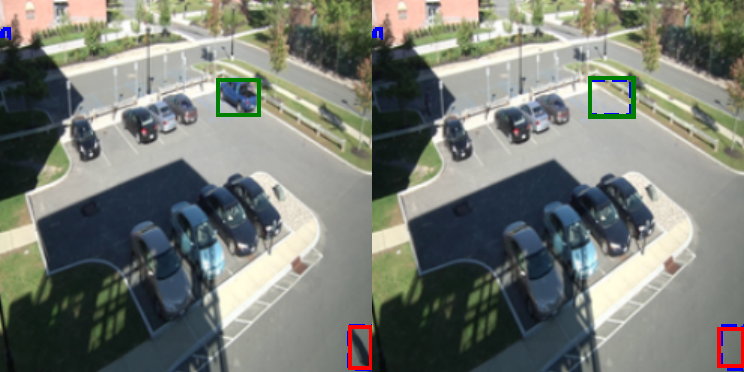}
     &\includegraphics[width=0.23\linewidth]{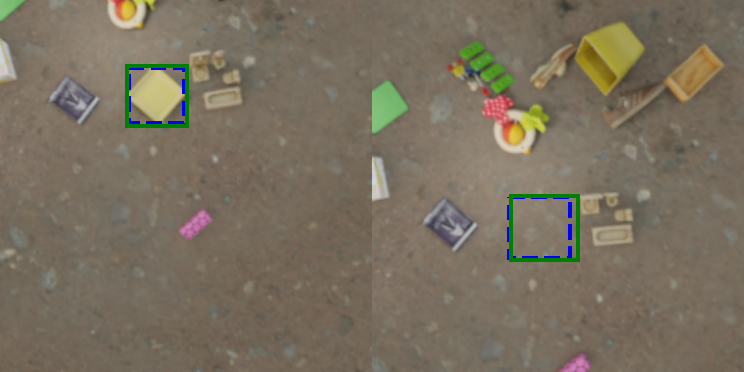}
     &\includegraphics[width=0.23\linewidth]{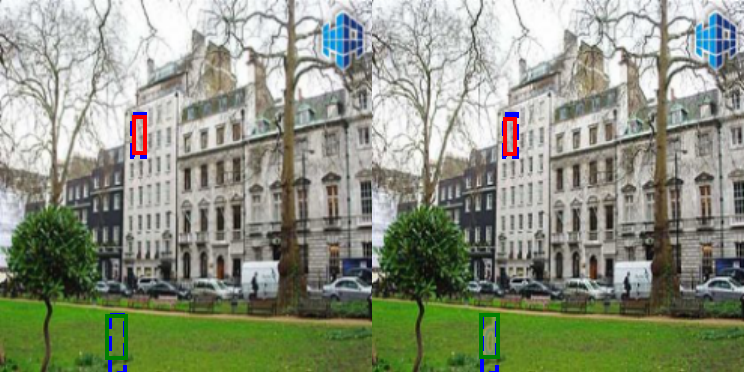}
     &\includegraphics[width=0.23\linewidth]{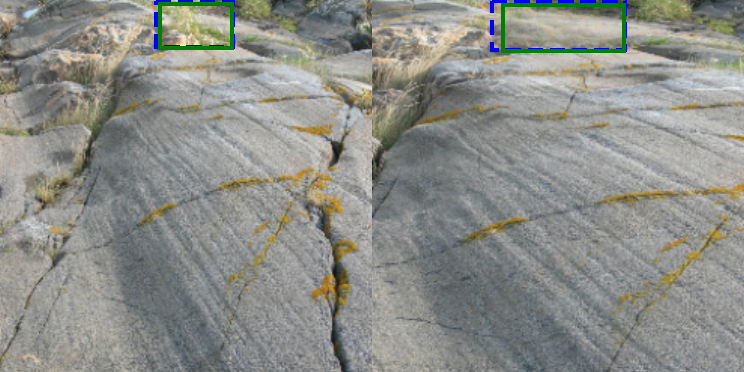}\\

     \vspace{-.07cm}
     \rotatebox{90}{\hspace{0.4cm} CYWS} 
     &\includegraphics[width=0.23\linewidth]{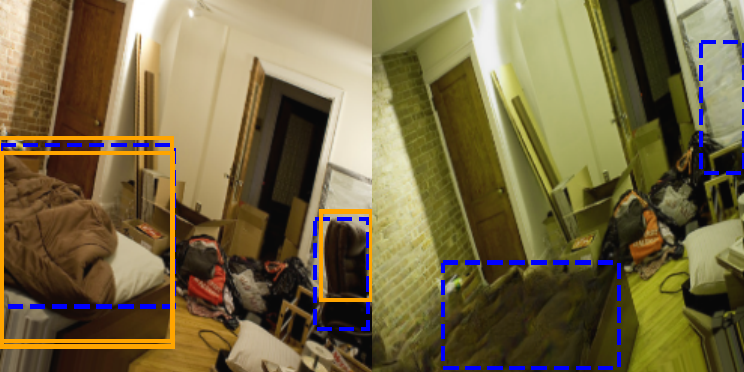} &\includegraphics[width=0.23\linewidth]{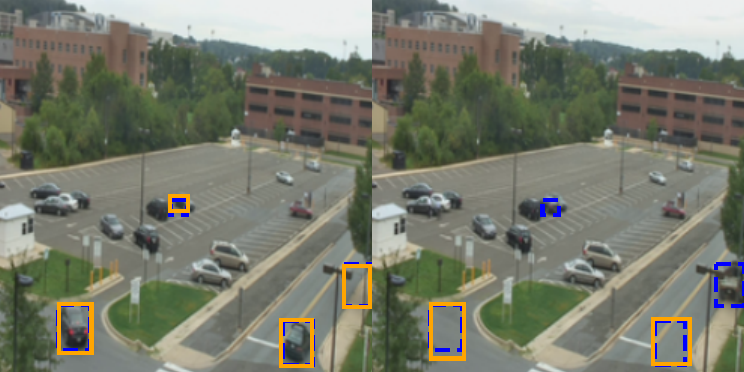}
     &\includegraphics[width=0.23\linewidth]{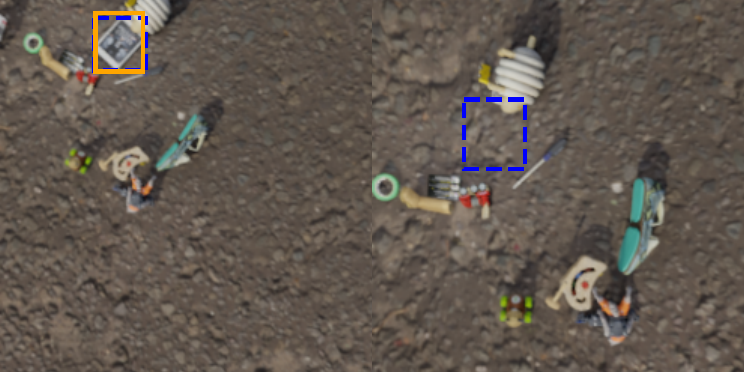}
     &\includegraphics[width=0.23\linewidth]{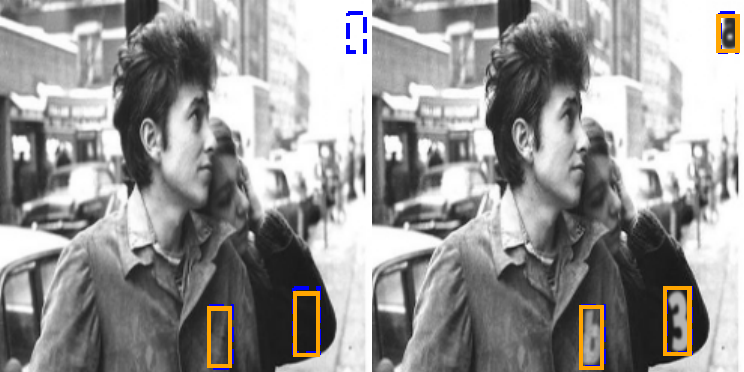}
     &\includegraphics[width=0.23\linewidth]{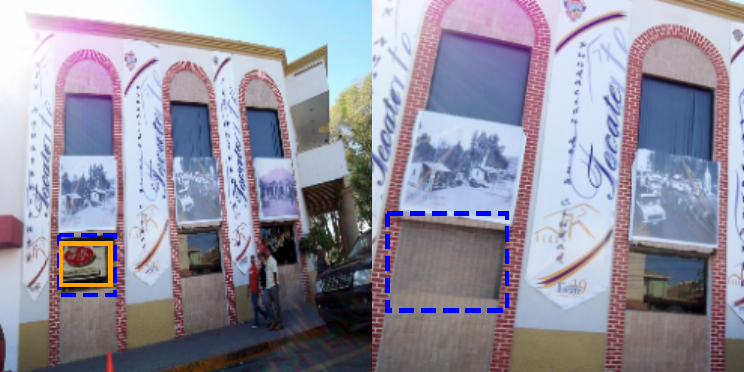}\\
     \vspace{-.059cm}
     \rotatebox{90}{\hspace{0.3cm} Our + PP} 
     &\includegraphics[width=0.23\linewidth]{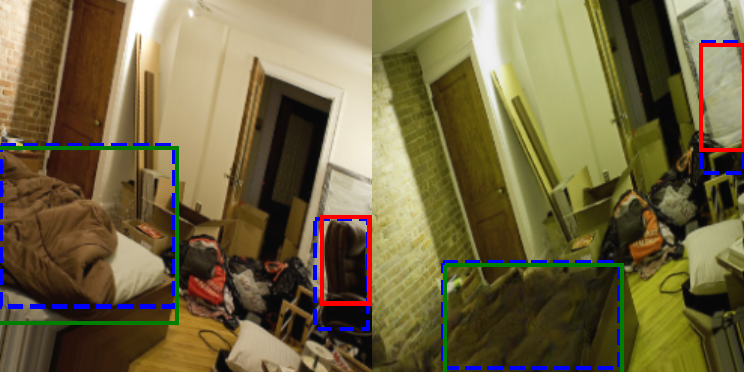}
     &\includegraphics[width=0.23\linewidth]{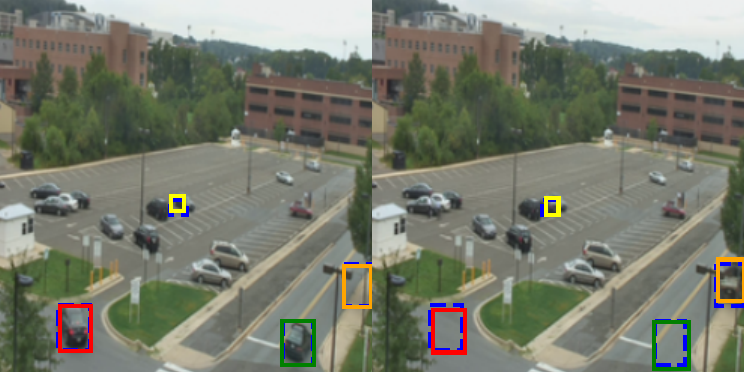}
     &\includegraphics[width=0.23\linewidth]{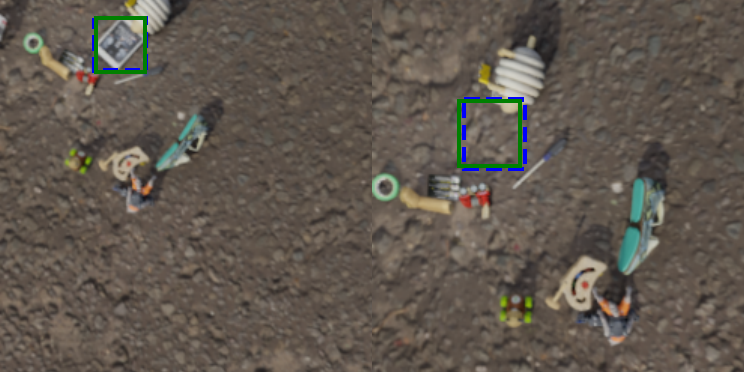}
     &\includegraphics[width=0.23\linewidth]{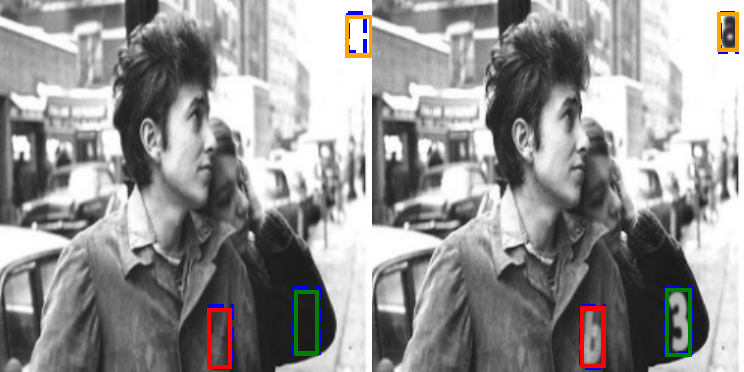}
     &\includegraphics[width=0.23\linewidth]{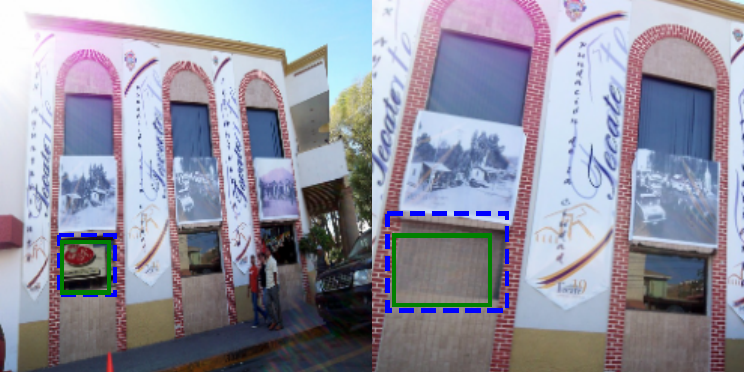}\\

     \vspace{-.07cm}
     \rotatebox{90}{\hspace{0.4cm} CYWS} 
     &\includegraphics[width=0.23\linewidth]{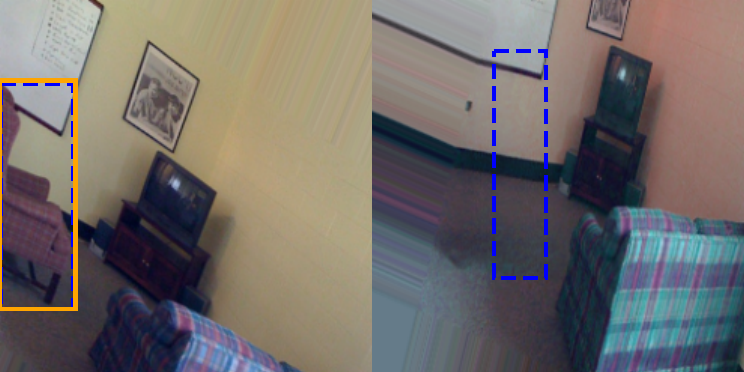} &\includegraphics[width=0.23\linewidth]{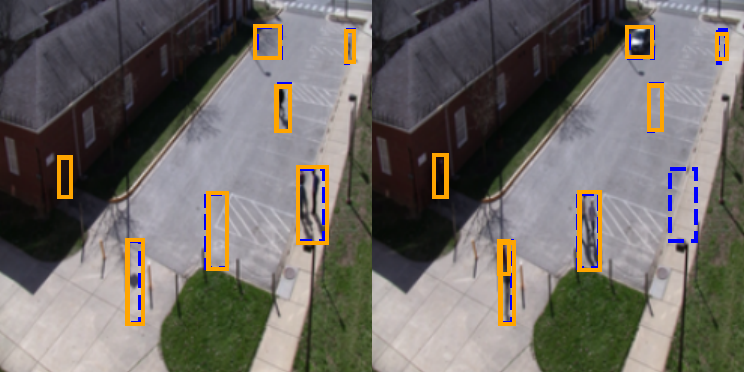}
     &\includegraphics[width=0.23\linewidth]{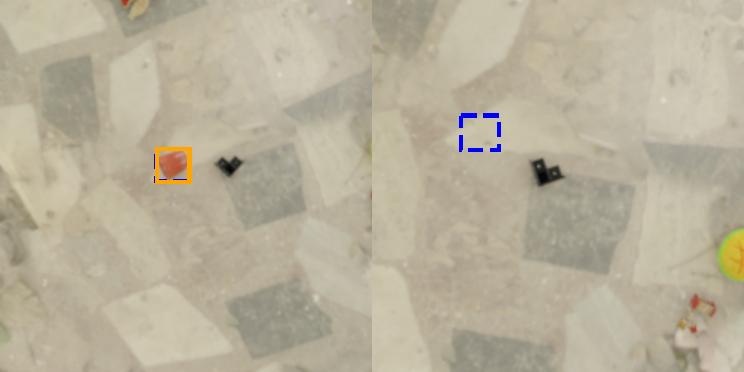}
     &\includegraphics[width=0.23\linewidth]{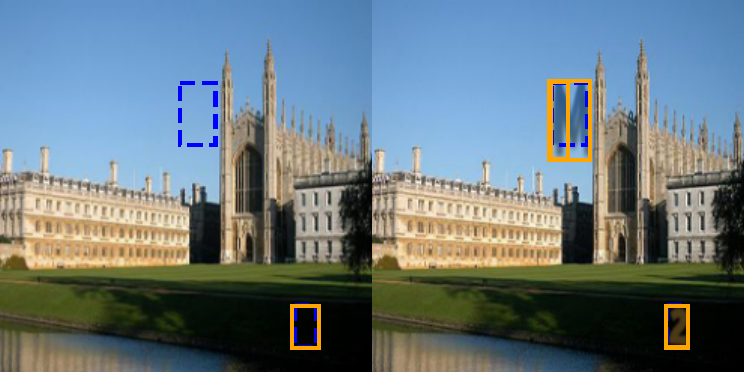} 
     &\includegraphics[width=0.23\linewidth]{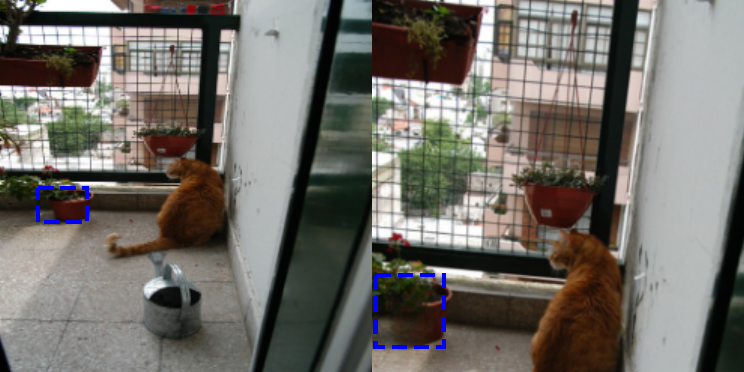} \\
     \vspace{-.059cm}
     \rotatebox{90}{\hspace{0.3cm} Our + PP} 
     &\includegraphics[width=0.23\linewidth]{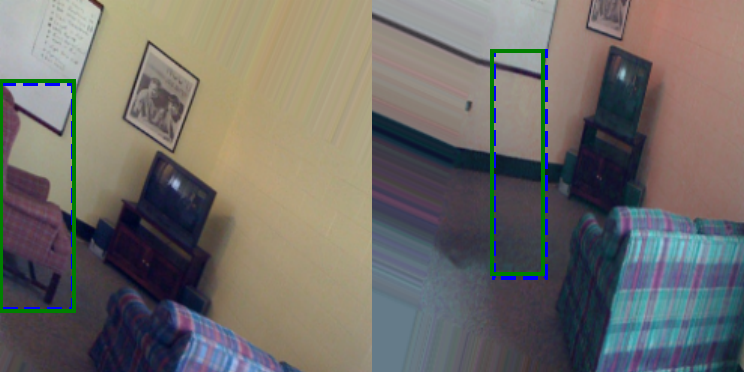}
     &\includegraphics[width=0.23\linewidth]{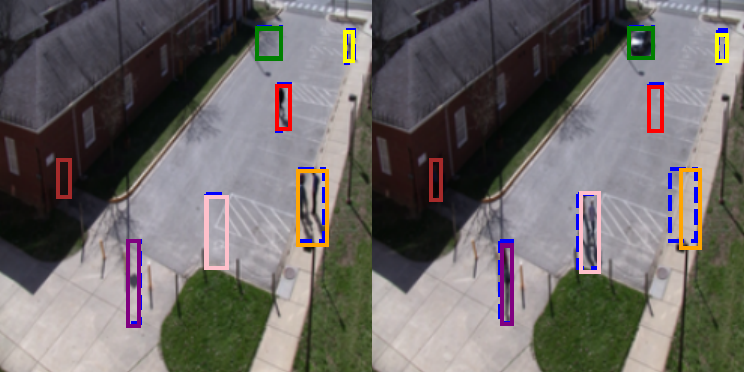}
     &\includegraphics[width=0.23\linewidth]{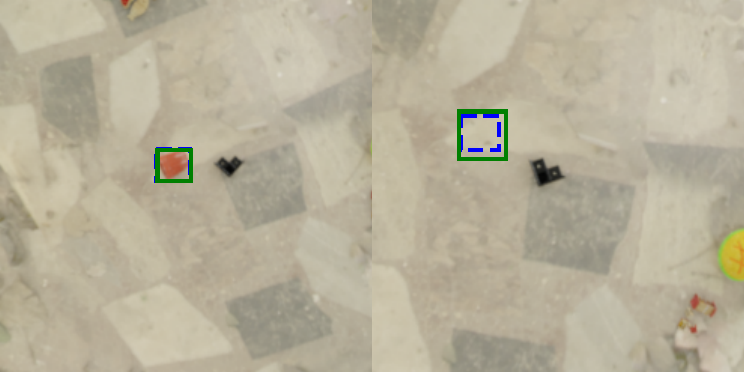}
     &\includegraphics[width=0.23\linewidth]{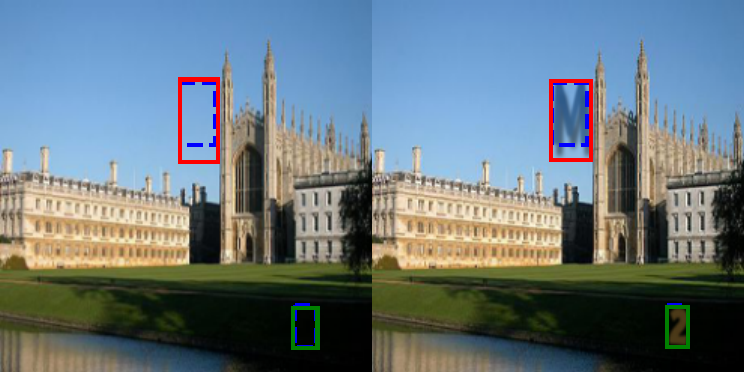}
     &\includegraphics[width=0.23\linewidth]{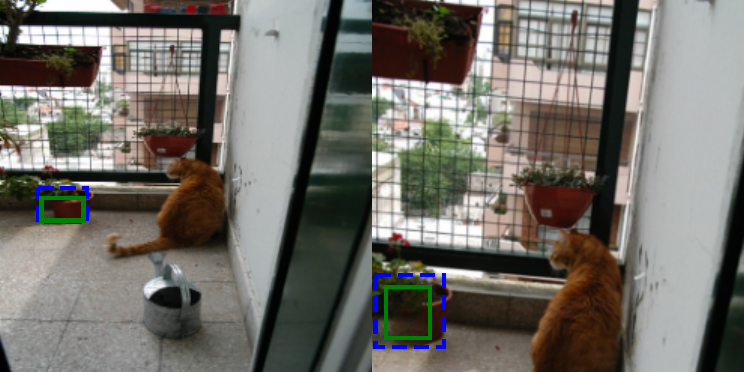} \\

     \end{tabular}}
     \caption{When comparing our model's change detection output to that of CYWS model, it is evident that our contrastive matching loss enhances the model's accuracy. Additionally, our post-processing technique can apply in many situations with multiple modifications
     }
     \label{fig:more qualitative results betwen cyws and our model}
\end{figure*}

In this part, we present further qualitative comparison findings between our fine-tuned model and CYWS\cite{sachdeva2023change} model following the use of a detection threshold of 0.25 and a post-processing technique. CYWS findings are shown in the first row, while the results of our model are shown in the second row. For qualitative results, see \cref{fig:more qualitative results betwen cyws and our model}.

\begin{figure*}[t]
     \centering
     \resizebox{\textwidth}{!}{
     \fontsize{6pt}{6pt}\selectfont
     \begin{tabular}{c@{}c@{}c@{}c@{}c@{}c}
     &(a) \cocologo &(b) \stdlogo &(c) \kubriclogo &(d) \synthlogo  &(e) \openlogo \\ 
     \vspace{-.07cm}
     \rotatebox{90}{\hspace{0.1cm} After Detection} 
     &\includegraphics[width=0.22\linewidth]{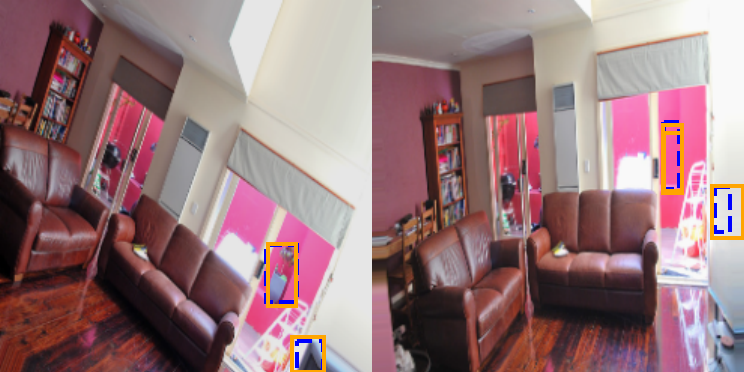} 
     &\includegraphics[width=0.22\linewidth]{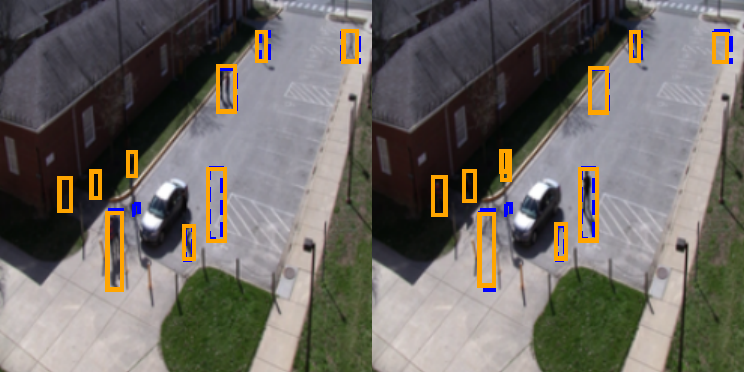}
     &\includegraphics[width=0.22\linewidth]{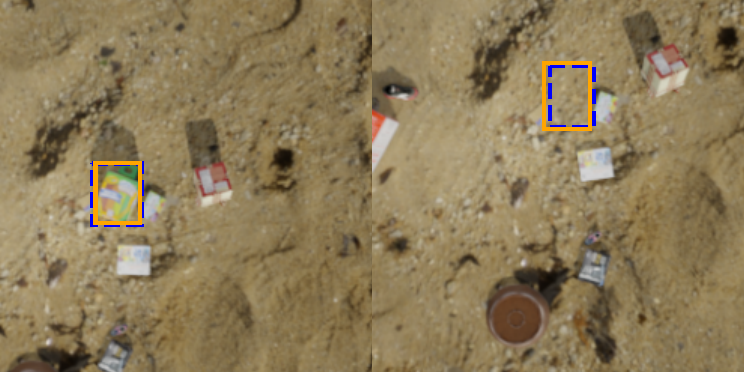}
     &\includegraphics[width=0.22\linewidth]{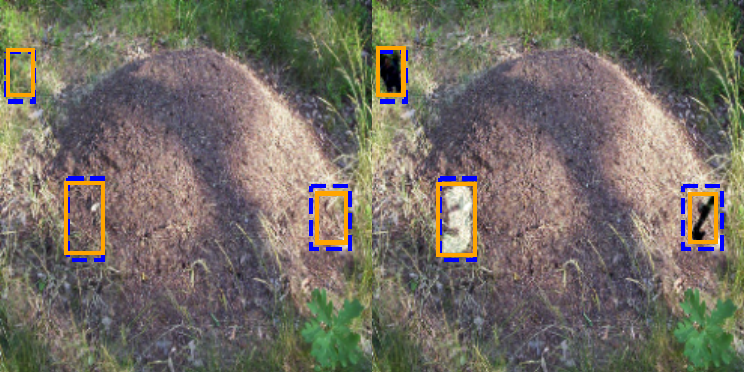} 
     &\includegraphics[width=0.22\linewidth]{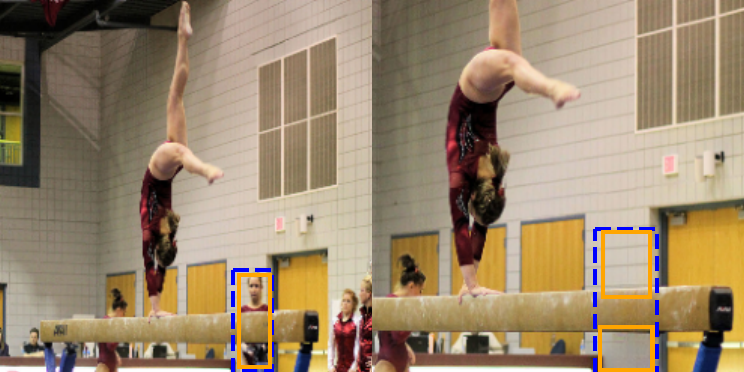}\\
     \vspace{-.059cm}
     \rotatebox{90}{\hspace{0.3cm} W/o Align} 
     &\includegraphics[width=0.22\linewidth]{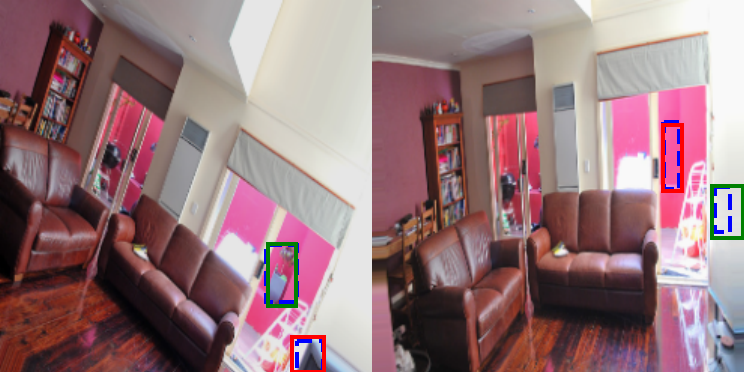}
     &\includegraphics[width=0.22\linewidth]{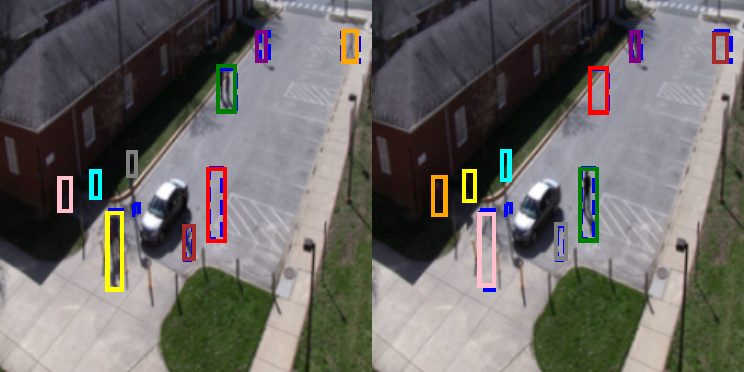}
     &\includegraphics[width=0.22\linewidth]{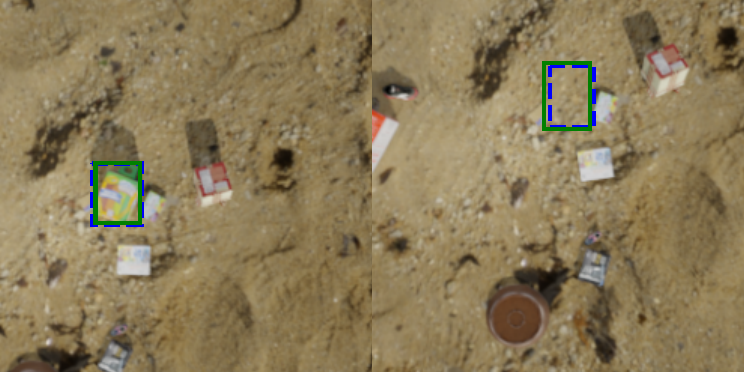}
     &\includegraphics[width=0.22\linewidth]{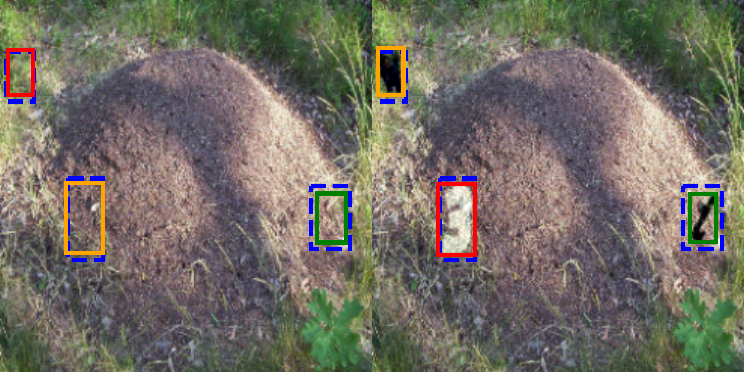}
     &\includegraphics[width=0.22\linewidth]{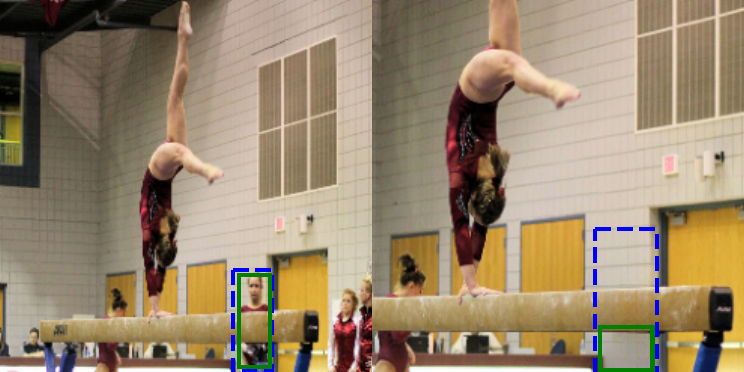}\\
      \vspace{-.059cm}
     \rotatebox{90}{\hspace{0.1cm} Post-processing} 
     &\includegraphics[width=0.22\linewidth]{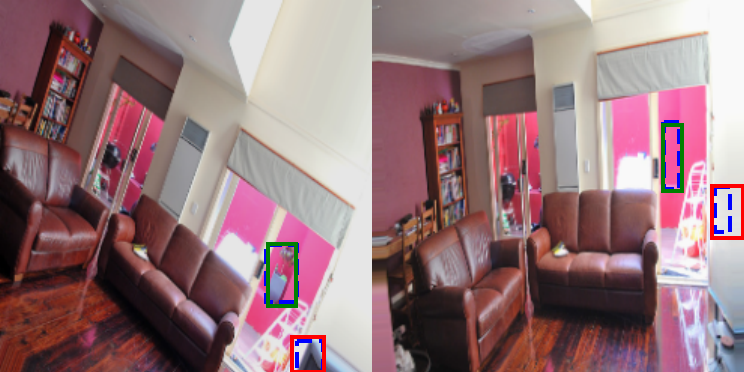}
     &\includegraphics[width=0.22\linewidth]{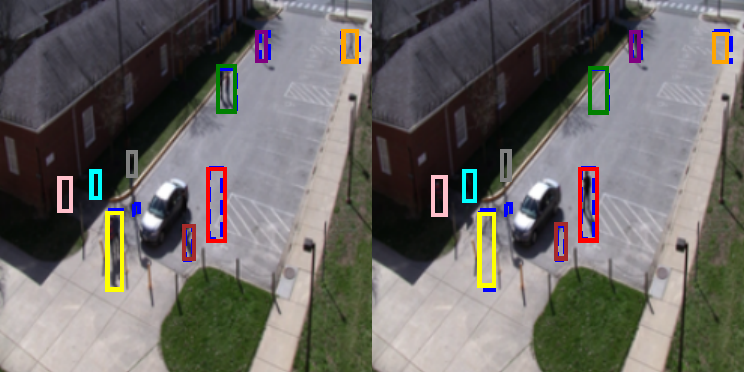}
     &\includegraphics[width=0.22\linewidth]{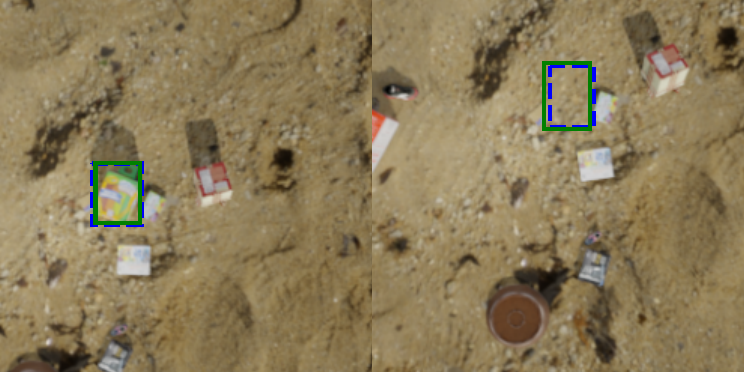}
     &\includegraphics[width=0.22\linewidth]{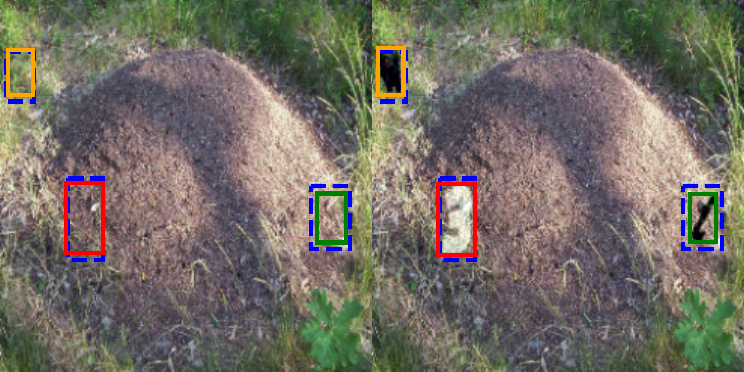}
     &\includegraphics[width=0.22\linewidth]{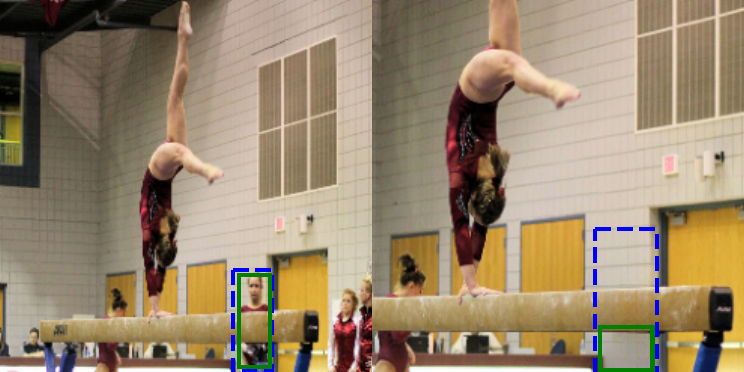}\\

     \vspace{-.07cm}
     \rotatebox{90}{\hspace{0.1cm} After Detection} 
     &\includegraphics[width=0.22\linewidth]{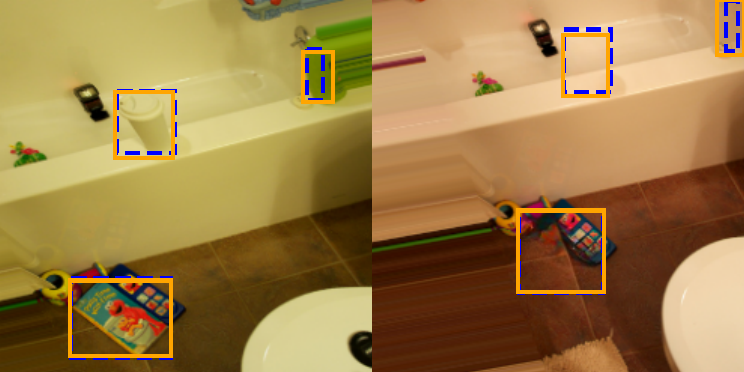} 
     &\includegraphics[width=0.22\linewidth]{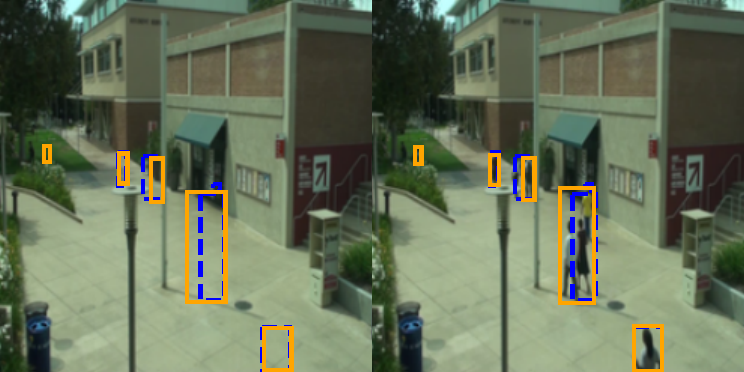}
     &\includegraphics[width=0.22\linewidth]{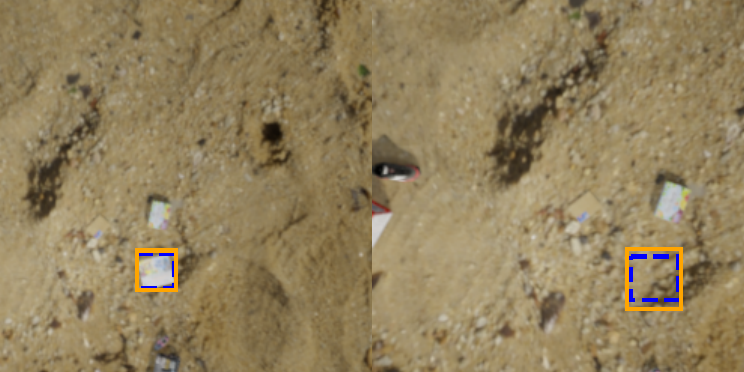}
     &\includegraphics[width=0.22\linewidth]{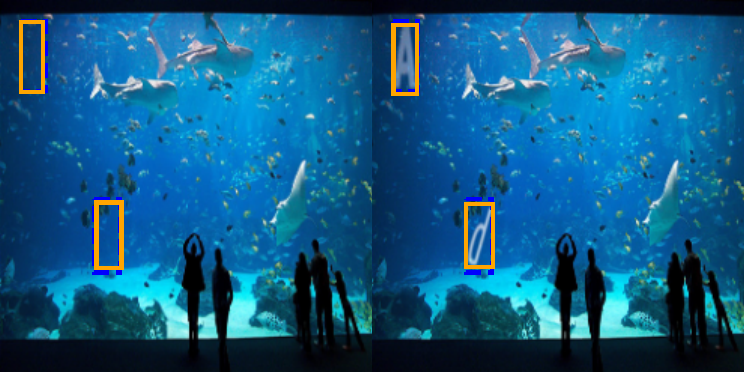} 
     &\includegraphics[width=0.22\linewidth]{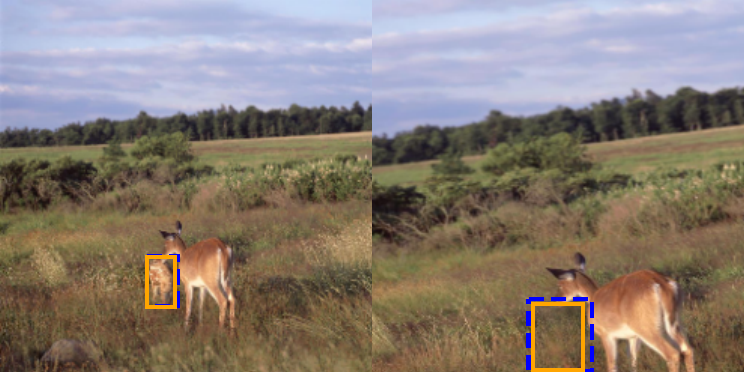}\\
     \vspace{-.059cm}
     \rotatebox{90}{\hspace{0.3cm} W/o Align} 
     &\includegraphics[width=0.22\linewidth]{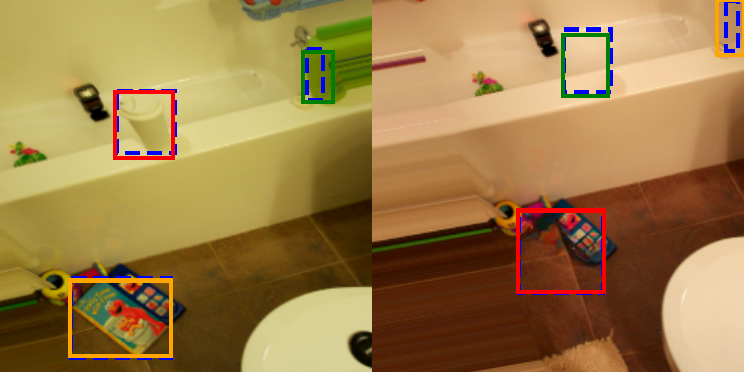}
     &\includegraphics[width=0.22\linewidth]{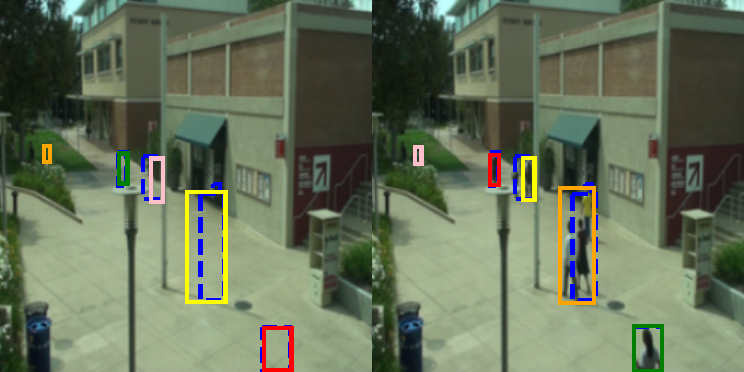}
     &\includegraphics[width=0.22\linewidth]{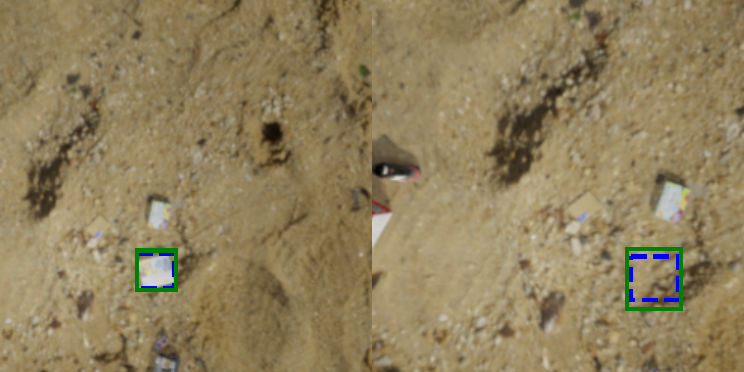}
     &\includegraphics[width=0.22\linewidth]{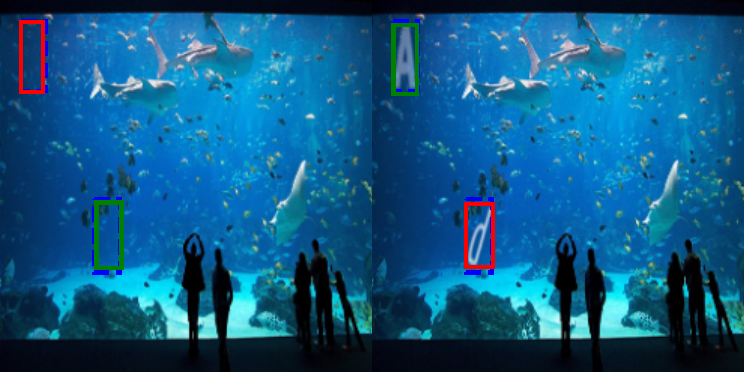}
     &\includegraphics[width=0.22\linewidth]{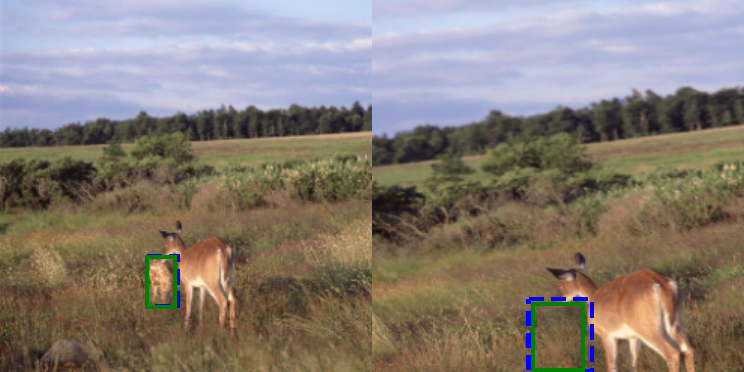}\\
      \vspace{-.059cm}
     \rotatebox{90}{\hspace{0.1cm} Post-processing} 
     &\includegraphics[width=0.22\linewidth]{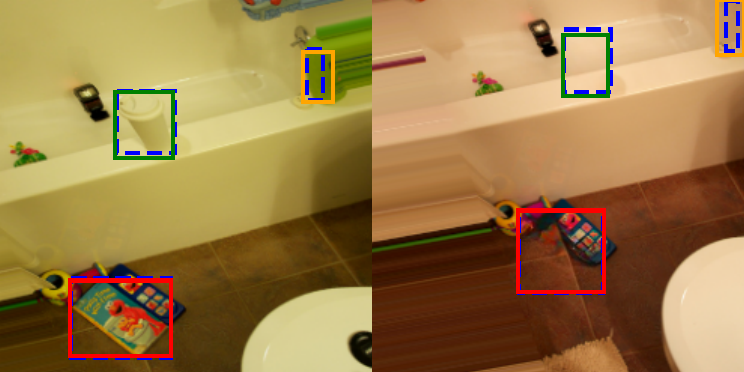}
     &\includegraphics[width=0.22\linewidth]{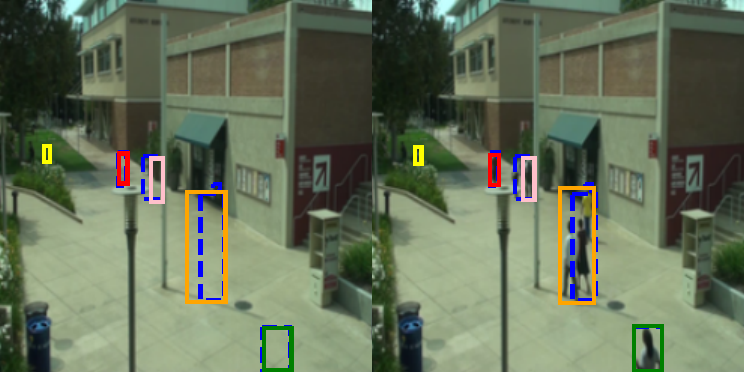}
     &\includegraphics[width=0.22\linewidth]{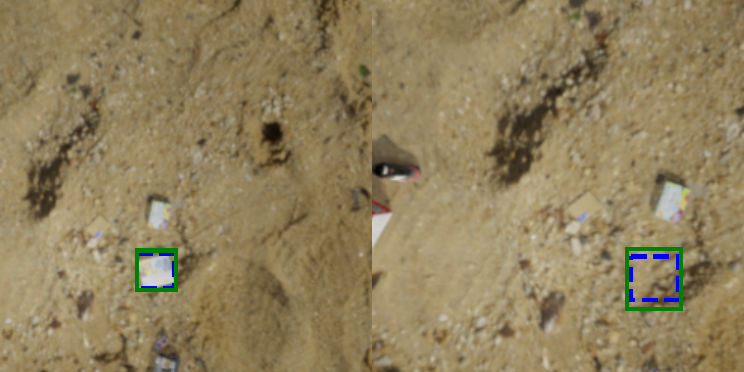}
     &\includegraphics[width=0.22\linewidth]{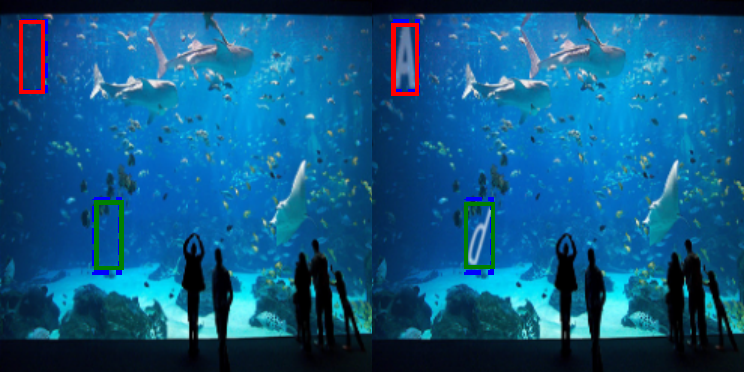}
     &\includegraphics[width=0.22\linewidth]{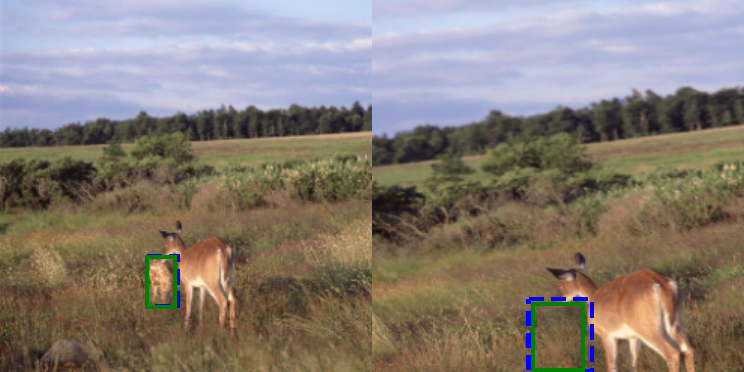}\\

     \vspace{-.07cm}
     \rotatebox{90}{\hspace{0.1cm} After Detection} 
     &\includegraphics[width=0.22\linewidth]{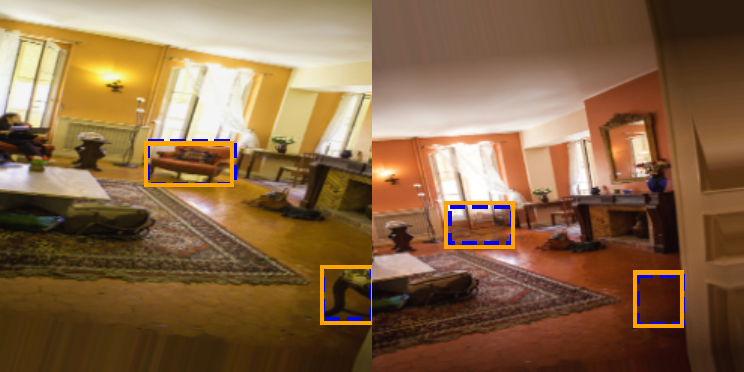} 
     &\includegraphics[width=0.22\linewidth]{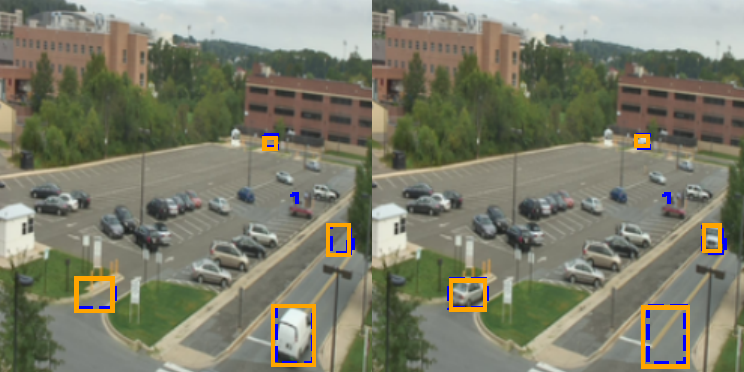}
     &\includegraphics[width=0.22\linewidth]{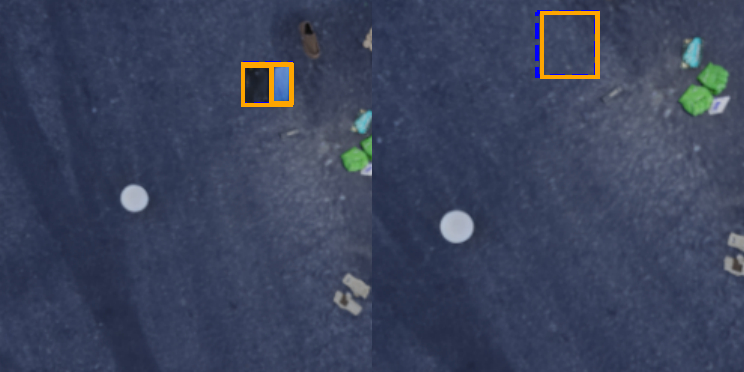}
     &\includegraphics[width=0.22\linewidth]{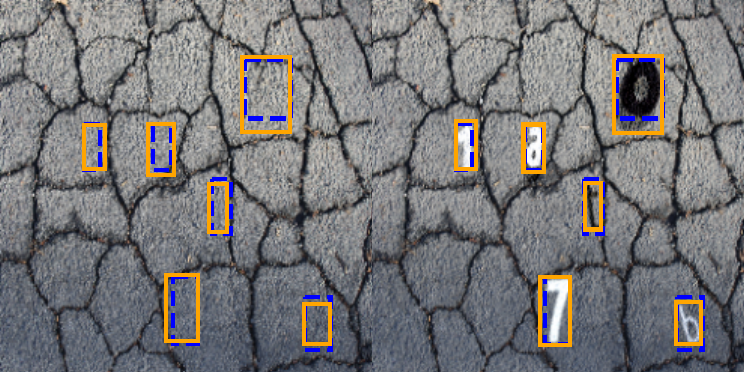}
     &\includegraphics[width=0.22\linewidth]{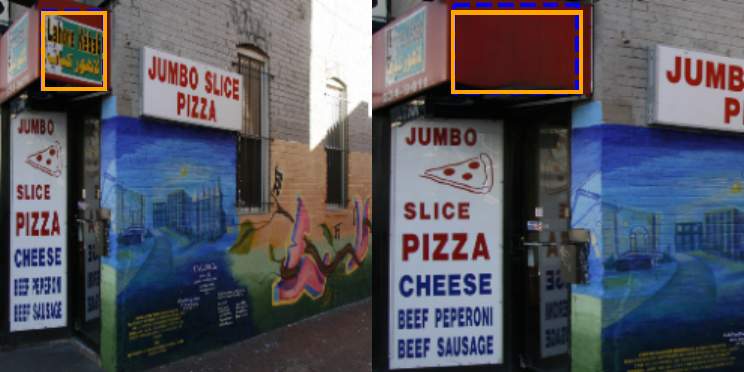}\\
     \vspace{-.059cm}
     \rotatebox{90}{\hspace{0.3cm} W/o Align} 
     &\includegraphics[width=0.22\linewidth]{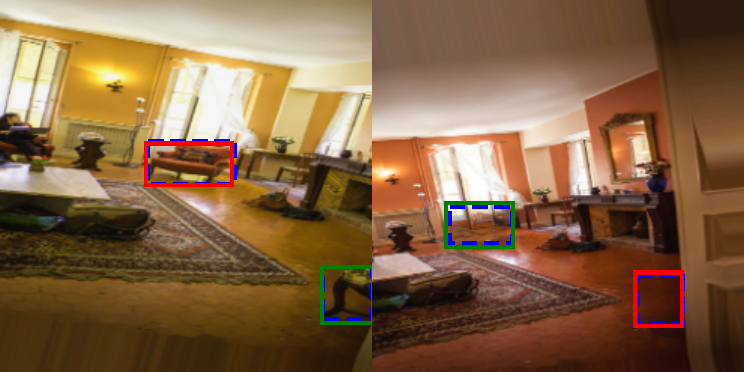}
     &\includegraphics[width=0.22\linewidth]{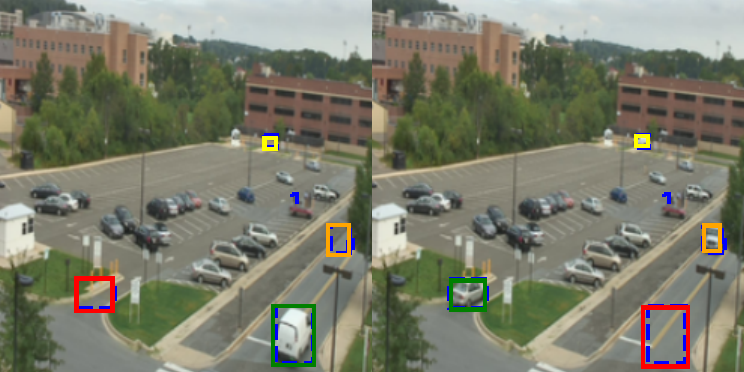}
     &\includegraphics[width=0.22\linewidth]{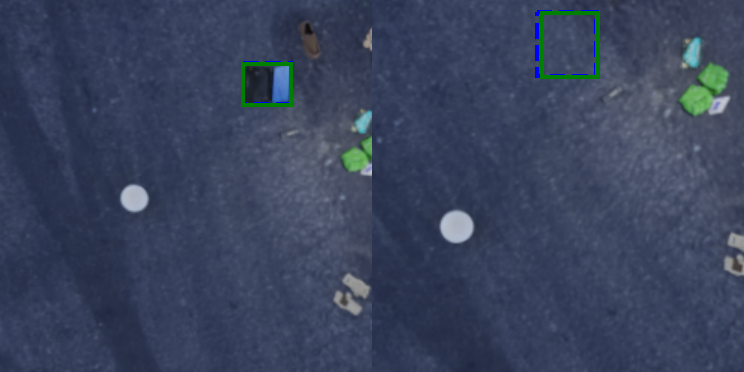}
     &\includegraphics[width=0.22\linewidth]{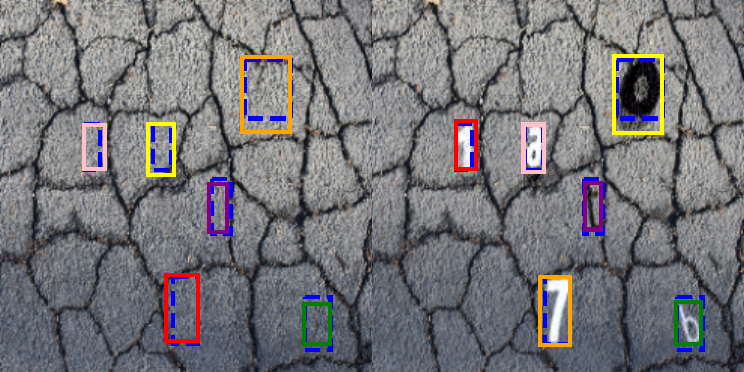}
     &\includegraphics[width=0.22\linewidth]{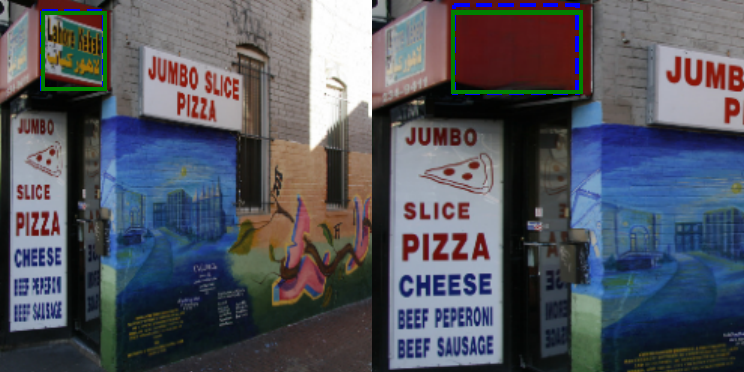}\\
      \vspace{-.059cm}
     \rotatebox{90}{\hspace{0.1cm} Post-processing} 
     &\includegraphics[width=0.22\linewidth]{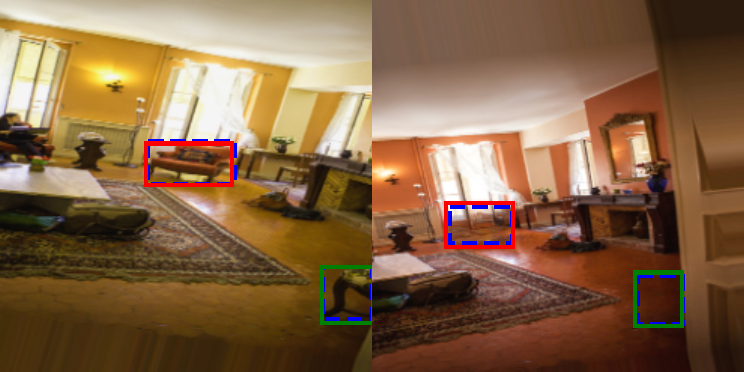}
     &\includegraphics[width=0.22\linewidth]{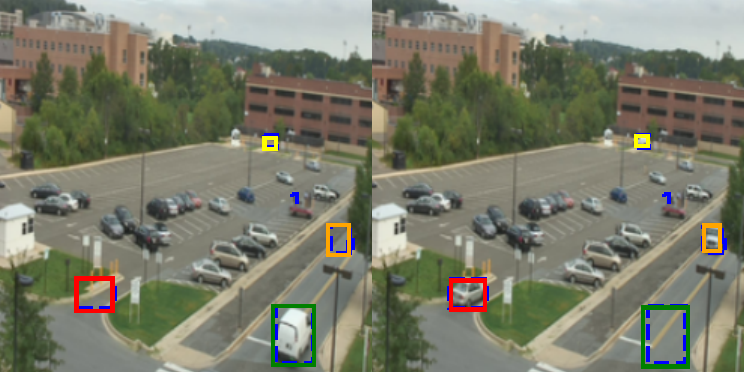}
     &\includegraphics[width=0.22\linewidth]{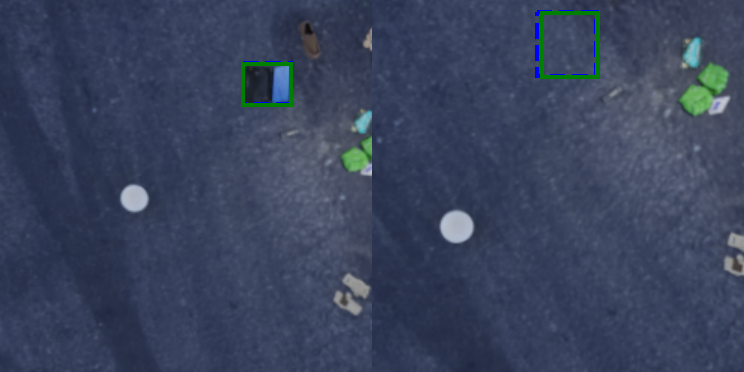}
     &\includegraphics[width=0.22\linewidth]{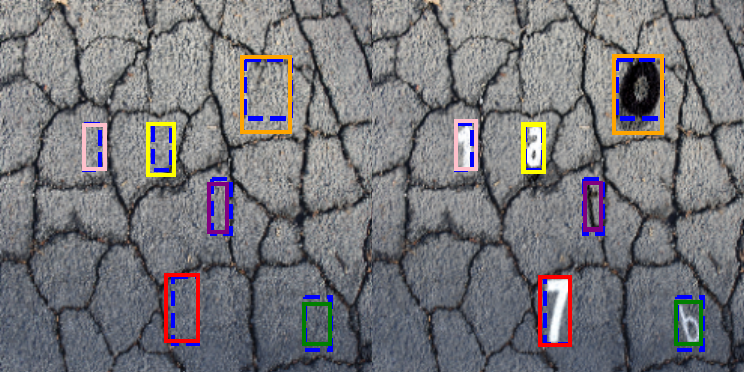}
     &\includegraphics[width=0.22\linewidth]{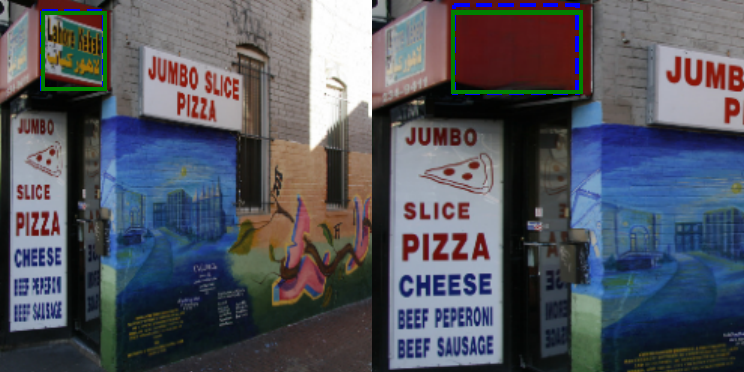}\\
     
     \vspace{-.07cm}
     \rotatebox{90}{\hspace{0.1cm} After Detection} 
     &\includegraphics[width=0.22\linewidth]{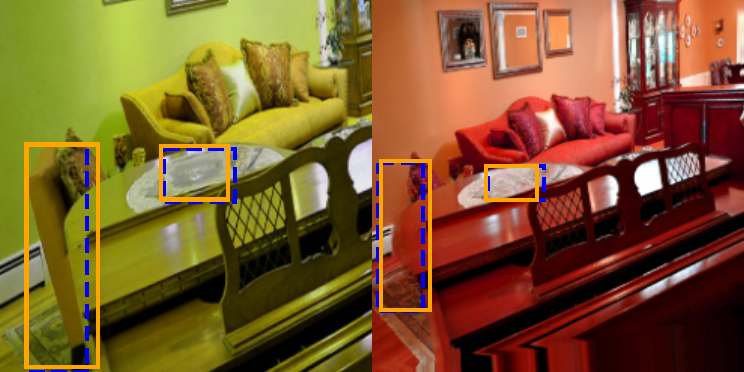} 
     &\includegraphics[width=0.22\linewidth]{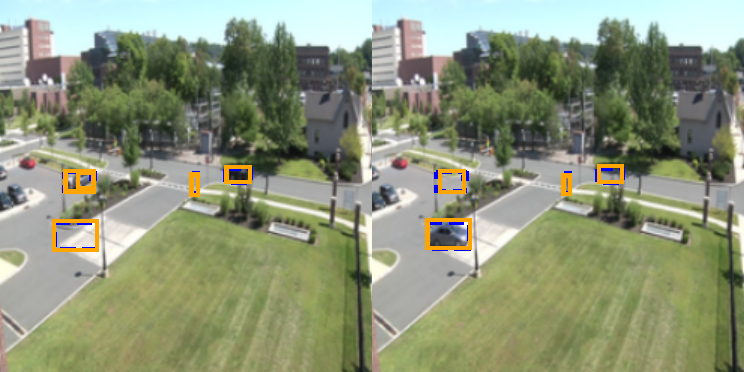}
     &\includegraphics[width=0.22\linewidth]{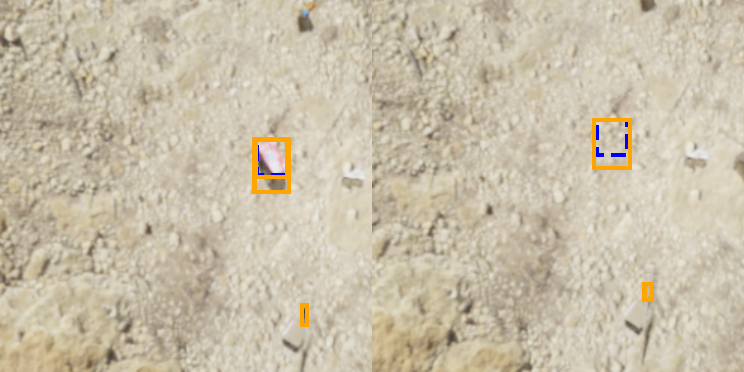}
     &\includegraphics[width=0.22\linewidth]{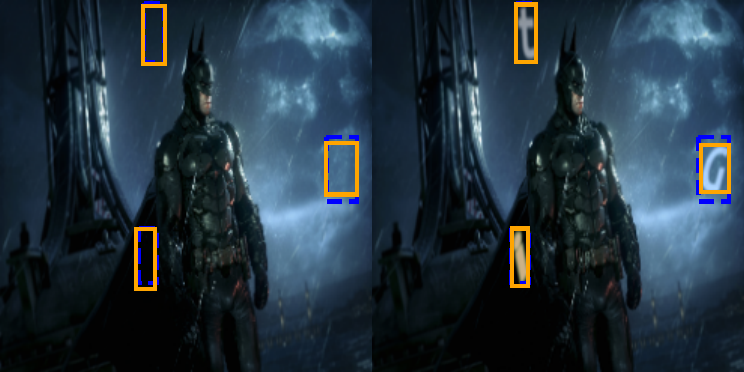} 
     &\includegraphics[width=0.22\linewidth]{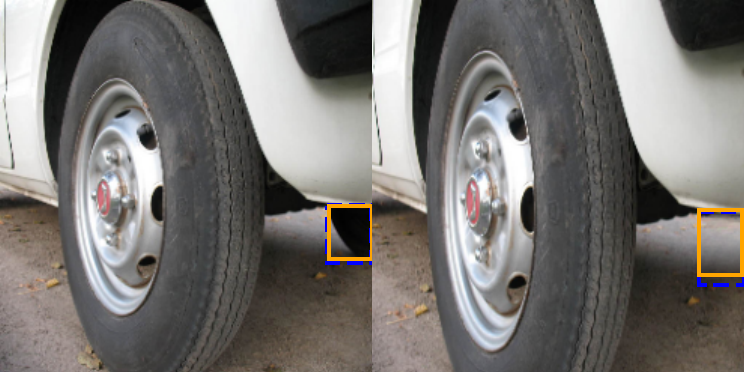}\\
     \vspace{-.059cm}
     \rotatebox{90}{\hspace{0.3cm} W/o Align} 
     &\includegraphics[width=0.22\linewidth]{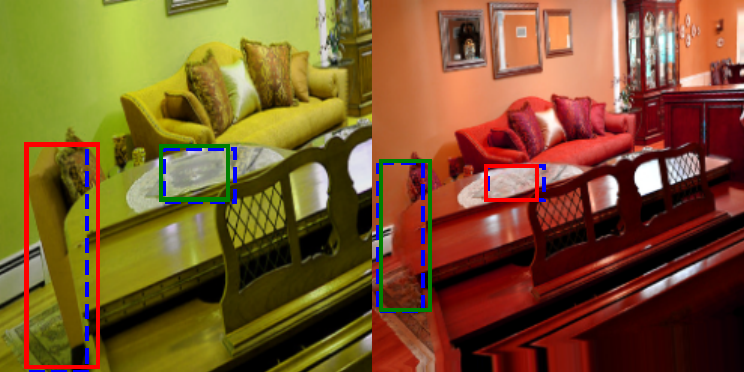}
     &\includegraphics[width=0.22\linewidth]{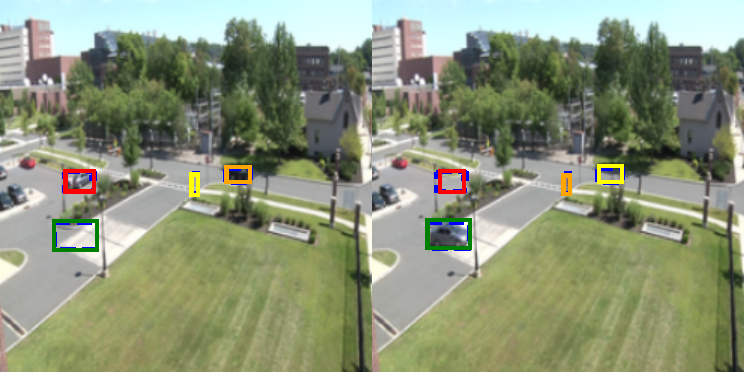}
     &\includegraphics[width=0.22\linewidth]{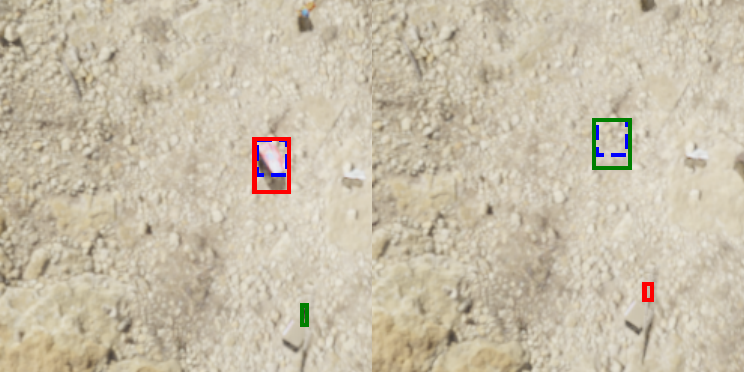}
     &\includegraphics[width=0.22\linewidth]{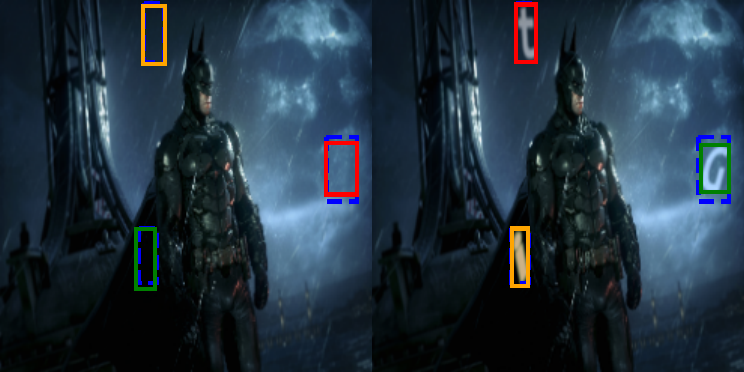}
      &\includegraphics[width=0.22\linewidth]{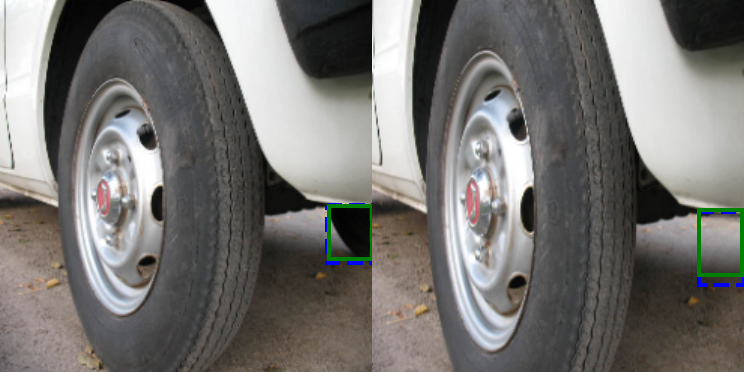}\\
      \vspace{-.059cm}
     \rotatebox{90}{\hspace{0.1cm} Post-processing} 
     &\includegraphics[width=0.22\linewidth]{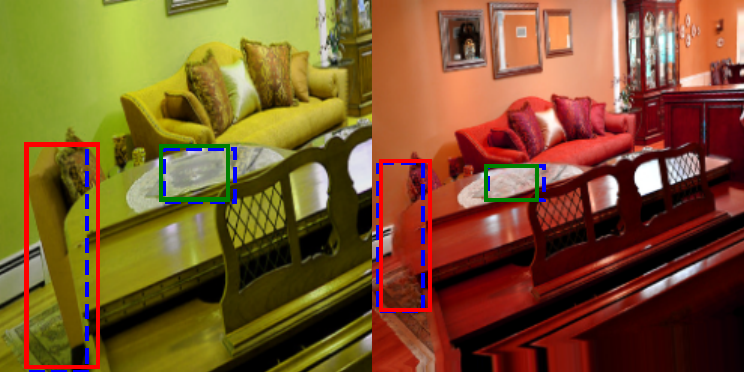}
     &\includegraphics[width=0.22\linewidth]{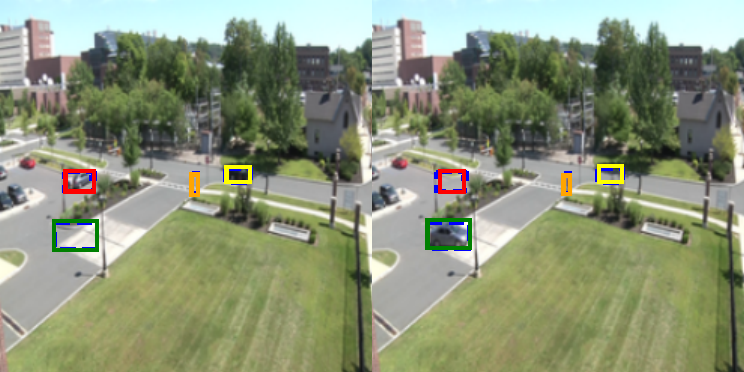}
     &\includegraphics[width=0.22\linewidth]{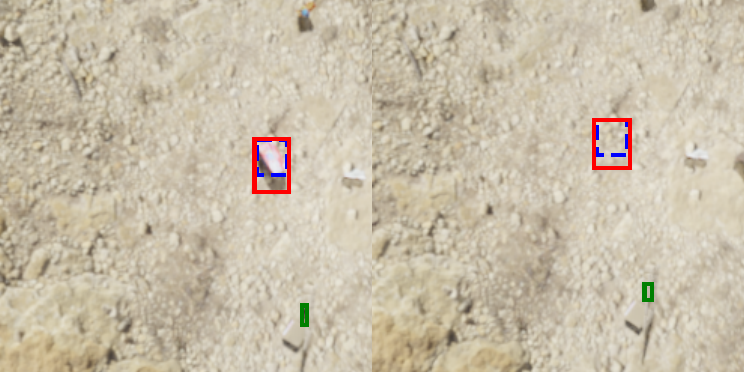}
     &\includegraphics[width=0.22\linewidth]{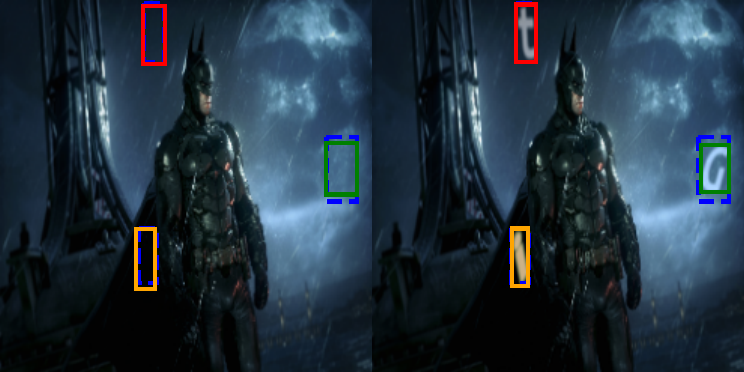}
     &\includegraphics[width=0.22\linewidth]{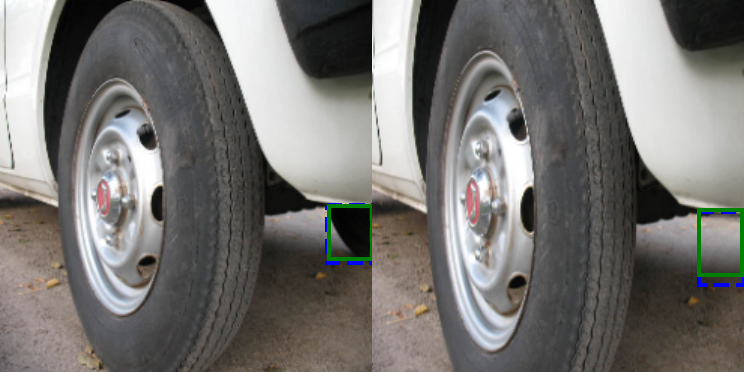}\\

     \end{tabular}}
     \caption{Contrasting the results following change detection and using our post-processing both with and without the alignment step. Evaluation of the findings in \cocologo, \stdlogo, \kubriclogo, \synthlogo, and \openlogo ~demonstrates the significance of the alignment stage
     }
     \label{fig:compare between after detection, without alignment, and post processing}
\end{figure*}

\section{Number of predicted box after applying detection threshold}
For both the ground-truth and our refined model with different thresholds, CYWS, we display the average number of boxes per image. You can view the detail in the \cref{tab:average number of predicted box in change case}

\begin{table}[ht]
\centering
\begin{tabular}{lccccccccc}
    \toprule
    \multicolumn{7}{c}{\textbf{Change}}\\
    \midrule
    \multicolumn{7}{c}{\textbf{Avg Predicted Box Per Image}} \\
    \midrule
    Model &Thres & \cocologo & \stdlogo & \kubriclogo & \synthlogo & \openlogo \\
    \midrule
    \text{Ground-Truth}        &\na            & \textbf{1.93}         & \textbf{5.85}           & \textbf{1.80}          & \textbf{1.10}    &\textbf{1.0}\\
    \midrule
    \textsc{CYWS}              &\na            & 100          & 100            & 100           & 100          & 100\\
    \text{Our}                 &\na            & 100          & 100            & 100           & 100          & 100\\
    \midrule
    \textsc{CYWS}              & 0.1           & 3.63         & 6.55           & 2.27          & 2.55        & 3.54\\
    \text{Our}                 & 0.1           & 3.21         & 7.23           & 2.37          & 2.25        & 2.64\\
    \midrule
    \textsc{CYWS}              & 0.2           & 1.75         & 4.38           & 1.95          & 1.23        & 1.17\\
    \text{Our}                 & 0.2           & 1.85         & 4.90           & 1.96          & 1.19        & 1.14\\
    \midrule
    \textsc{CYWS}              & 0.3           & 0.98         & 2.81           & 1.75          & 0.80        & 0.54\\
    \text{Our}                 & 0.3           & 1.20         & 3.13           & 1.79          & 0.84        & 0.67\\      
    \midrule
    \textsc{CYWS}              & 0.4           & 0.55         & 1.38           & 1.50          & 0.49        & 0.26\\
    \text{Our}                 & 0.4           & 0.70         & 1.48           & 1.56          & 0.63        & 0.37\\
    \midrule
    \textsc{CYWS}              & 0.5           & 0.29         & 0.59           & 1.08          & 0.42        & 0.10\\
    \text{Our}                 & 0.5           & 0.45         & 0.60           & 1.18          & 0.44        & 0.18\\
    \bottomrule
    \end{tabular}
     \caption{\textbf{Average Predicted Box Per Image for Change with Different Thresholds.} Evaluate the influence of detection threshold on the number of predicted boxes per image in change case with CYWS model and our fineturned model}
    \label{tab:average number of predicted box in change case}
\end{table}

\end{document}